\documentclass[letterpaper]{article} 
\usepackage{aaai2026}  
\usepackage{times}  
\usepackage{helvet}  
\usepackage{courier}  
\usepackage[hyphens]{url}  
\usepackage{graphicx} 
\usepackage{natbib}  
\usepackage{caption} 
\usepackage{algorithm}

\usepackage{newfloat}
\usepackage{listings}
\DeclareCaptionStyle{ruled}{labelfont=normalfont,labelsep=colon,strut=off} 
\floatstyle{ruled}
\newfloat{listing}{tb}{lst}{}
\floatname{listing}{Listing}
\usepackage{amsmath}
\usepackage{adjustbox}
\usepackage{multirow}
\usepackage{booktabs}
\usepackage{makecell}
\usepackage{algpseudocode}
\usepackage{xcolor}
\usepackage{bm}
\usepackage{anyfontsize}
\usepackage{subcaption}
\usepackage{svg}
\usepackage{amsfonts}

\newtheorem{theorem}{Theorem}
\newtheorem{corollary}{Corollary}

\def\eqref#1{(\ref{eq:#1})}
\def\eqlabel#1{\label{eq:#1}}
\def\figref#1{\ref{fig:#1}}
\def\figlabel#1{\label{fig:#1}}

\def\m#1{\ensuremath{\mathtt{#1}}}

\def\v#1{\ensuremath{\mathbf{#1}}}

\def\real{\mathbb{R}}
\def\tr{^\top}

\def\norm#1{\left\lVert#1\right\rVert}
\def\l2#1{\norm{#1}_2}

\usepackage{xcolor}

\newcommand{\hide}[1]{\iffalse #1\fi}
\title{PINGS-X: Physics-Informed Normalized Gaussian Splatting with Axes Alignment for Efficient Super-Resolution of 4D Flow MRI}

\author{
    Sun Jo\textsuperscript{\rm 1,5}\equalcontrib, 
    Seok Young Hong\textsuperscript{\rm 4}\equalcontrib, 
    Jinhyun Kim\textsuperscript{\rm 2,5}, 
    Seungmin Kang\textsuperscript{\rm 3,5}, 
    Ahjin Choi\textsuperscript{\rm 2,5}, \\
    Don-Gwan An\textsuperscript{\rm 3,5}, 
    Simon Song\textsuperscript{\rm 3,5}{\textsuperscript{\rm \dag}}, 
    Je Hyeong Hong\textsuperscript{\rm 1,2,5}\thanks{Corresponding authors.}
}
\affiliations{

    \textsuperscript{\rm 1}Department of Artificial Intelligence, Hanyang University, Seoul, Republic of Korea\\
    \textsuperscript{\rm 2}Department of Electronic Engineering, Hanyang University, Seoul, Republic of Korea\\
    \textsuperscript{\rm 3}Department of Mechanical Engineering, Hanyang University, Seoul, Republic of Korea\\
    \textsuperscript{\rm 4}School of Social Sciences and School of Physical and Mathematical Sciences, Nanyang Technological University, Singapore\\
    \textsuperscript{\rm 5}Center for Precision Medicine Platform Based on Smart Hemo-Dynamic Index, Hanyang University\\
    \{choyw5, jhkim02, tmdals1213, caj0328, adg0324, simonsong, jhh37\}@hanyang.ac.kr, 
    seokyoung.hong@ntu.edu.sg
}

\usepackage{bibentry}

\begin{document}

\maketitle

\begin{abstract}

4D flow magnetic resonance imaging (MRI) is a reliable, non-invasive approach for estimating blood flow velocities, vital for cardiovascular diagnostics. Unlike conventional MRI focused on anatomical structures, 4D flow MRI requires high spatiotemporal resolution for early detection of critical conditions such as stenosis or aneurysms. However, achieving such resolution typically results in prolonged scan times, creating a trade-off between acquisition speed and prediction accuracy.
Recent studies have leveraged physics-informed neural networks (PINNs) for super-resolution of MRI data, but their practical applicability is limited as the prohibitively slow training process must be performed for each patient.
To overcome this limitation, we propose PINGS-X, a novel framework modeling high-resolution flow velocities using axes-aligned spatiotemporal Gaussian representations.
Inspired by the effectiveness of 3D Gaussian splatting (3DGS) in novel view synthesis, PINGS-X extends this concept through several non-trivial novel innovations: (i) normalized Gaussian splatting with a formal convergence guarantee, (ii) axes-aligned Gaussians that simplify training for high-dimensional data while preserving accuracy and the convergence guarantee, and (iii) a Gaussian merging procedure to prevent degenerate solutions and boost computational efficiency.
Experimental results on computational fluid dynamics (CFD) and real 4D flow MRI datasets demonstrate that PINGS-X substantially reduces training time while achieving superior super-resolution accuracy.

\end{abstract}

\begin{links}
    \link{Code}{https://github.com/SpatialAILab/PINGS-X}
\end{links}

\section{Introduction}

4D flow MRI is a powerful, non-invasive method for quantifying blood flow velocities~\cite{markl20124d} without involving ionizing radiation.
It captures time-resolved, three-dimensional velocity fields using phase-contrast MRI, which extends conventional MRI by encoding velocity-sensitive phase shifts into the MR signal.
Crucially, the detailed hemodynamic insights from high-resolution 4D flow MRI are vital for predicting the progression of critical conditions such as stenosis and aneurysms~\cite{ferdian2023cerebrovascular}.

\begin{figure}[t]%
\centering
\small
\includegraphics[width=\columnwidth]{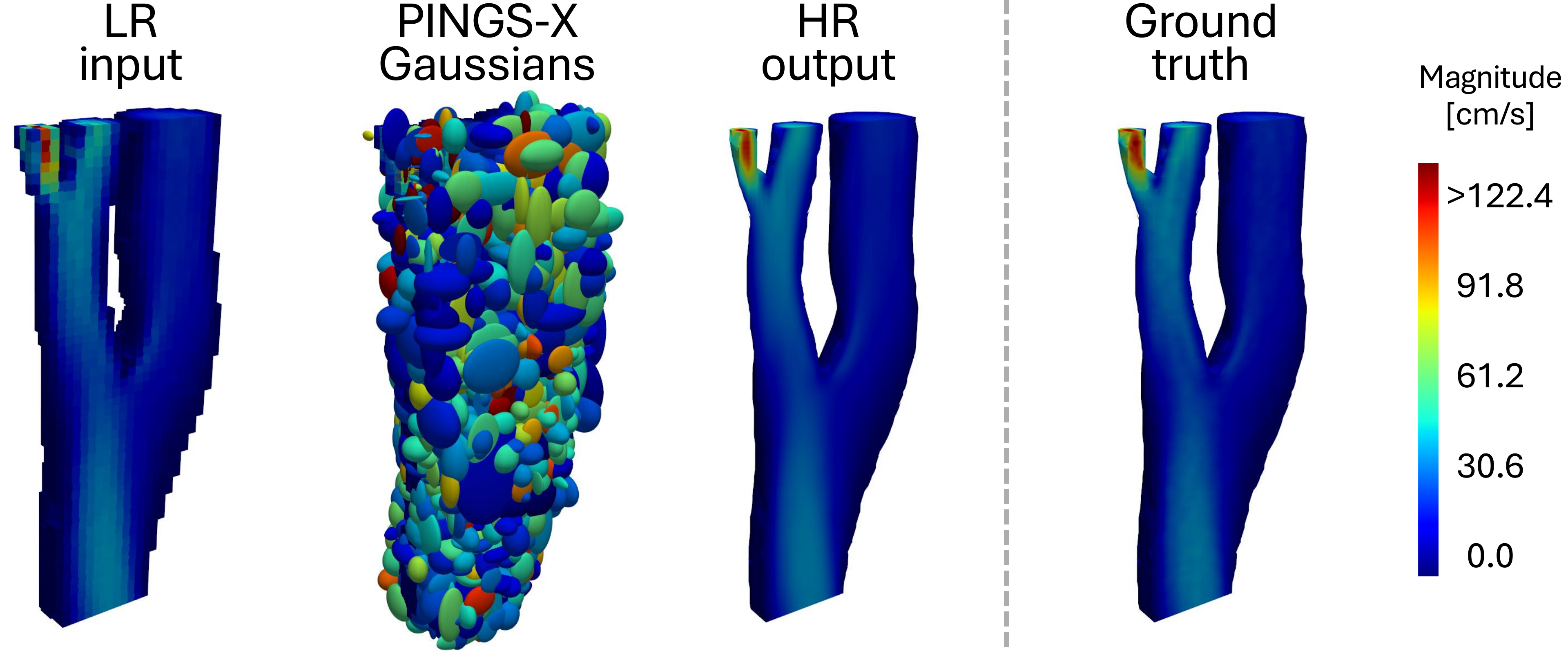}
\caption{
Super-resolution of a carotid artery velocity field. From a low-resolution (LR) input, PINGS-X optimizes a set of 4D spatiotemporal Gaussians (visualized here in 3D at a single time frame). 
This explicit representation allows us to reconstruct a high-resolution (HR) output that faithfully recovers the complex flow patterns of the ground truth.
}
\label{fig:teaser}
\end{figure}

Nevertheless, achieving high spatiotemporal resolution is challenging, as the multi-dimensional acquisition process involving four velocity encodings for each 3D volume and time frame is inherently slow~\cite{markl20124d}. While modern acceleration techniques like parallel imaging~\cite{deshmane2012parallel} and compressed sensing~\cite{lustig2007sparse, kontogiannis2022physics} are used, they have inherent limitations. Parallel imaging suffers from noise amplification at high acceleration rates, while compressed sensing relies on sparsity assumptions that fail in complex flow, often blurring structures~\cite{gottwald2020pseudo} and underestimating peak velocities~\cite{pathrose2021highly}. This creates a clinical dilemma: accept rapid, low-resolution scans that risk obscuring information, or perform long, high-fidelity scans that are impractical due to patient discomfort and motion artifacts~\cite{zaitsev15motion}. This motivates a complementary strategy of post-hoc spatial super-resolution to enhance detail from rapidly acquired data without increasing scan duration.


This potential has motivated various super-resolution strategies for 4D flow MRI~\cite{ferdian2023cerebrovascular, saitta2024implicit}. Data-driven deep learning methods can learn effective low-to-high-resolution mappings~\cite{shit2022srflow, ferdian2023cerebrovascular} but require large patient datasets and often struggle with generalization to unseen cardiovascular domains~\cite{ericsson2024generalized}. The alternative Physics-Informed Neural Network (PINN) paradigm~\cite{raissi2019physics} enforces physical laws on a per-scan basis, circumventing the need for large datasets. However, PINNs must be re-trained for each patient~\cite{callmer2025temporal}, and their reliance on slow implicit neural representations creates a critical computational bottleneck for high-dimensional 4D flow MRI~\cite{saitta2024implicit, callmer2025temporal}.


To address these challenges, we draw inspiration from novel view synthesis, where explicit representations have recently surpassed slower, implicit models. Notably, 3D Gaussian splatting (3DGS)~\cite{3dgs} achieved state-of-the-art results with training times orders of magnitude faster than neural radiance fields (NeRFs)~\cite{nerf}. This motivates our key question: \emph{Can a bespoke Gaussian-based representation overcome the computational bottlenecks of PINNs for 4D flow MRI super-resolution while retaining theoretical guarantee of convergence?}

In this paper, we address above question by introducing PINGS-X: Physics-Informed Normalized Gaussian Splatting with aXes alignment—a novel framework that adapts the principles of 3DGS to model 4D velocity fields using a set of optimized spatiotemporal Gaussians. As  direct adoption of 3DGS is infeasible in physics-informed super-resolution, we propose several key innovations as follows:
\begin{itemize}

\item \textbf{Normalized Gaussian splatting with a formal convergence guarantee}: We propose a normalized Gaussian splatting scheme which enables faithful representation of continuous flow profiles. 
We provide a formal proof of convergence with a highly favorable convergence rate.

\item \textbf{Axes-aligned Gaussians for efficient training}: 
We introduce axes-aligned Gaussians that simplify computation and improve training efficiency for high-dimensional data such as 4D flow MRI while preserving the convergence guarantee with minimal impact on accuracy.

\item \textbf{Gaussian merging for stability and scalability}:
 To prevent potential degenerate solutions induced by the normalization scheme and enhance computational scalability, we present a density control strategy that merges nearby Gaussians with similar characteristics.

\end{itemize}
Each contribution is evaluated on synthetic CFD and real 4D flow MRI datasets with spatially averaged data, creating a more realistic challenging testbed than the sampling‑based settings used in previous work~\cite{dirix2022synthesis}.

\begin{figure}[t]
\centering
    \includegraphics[width=\columnwidth]{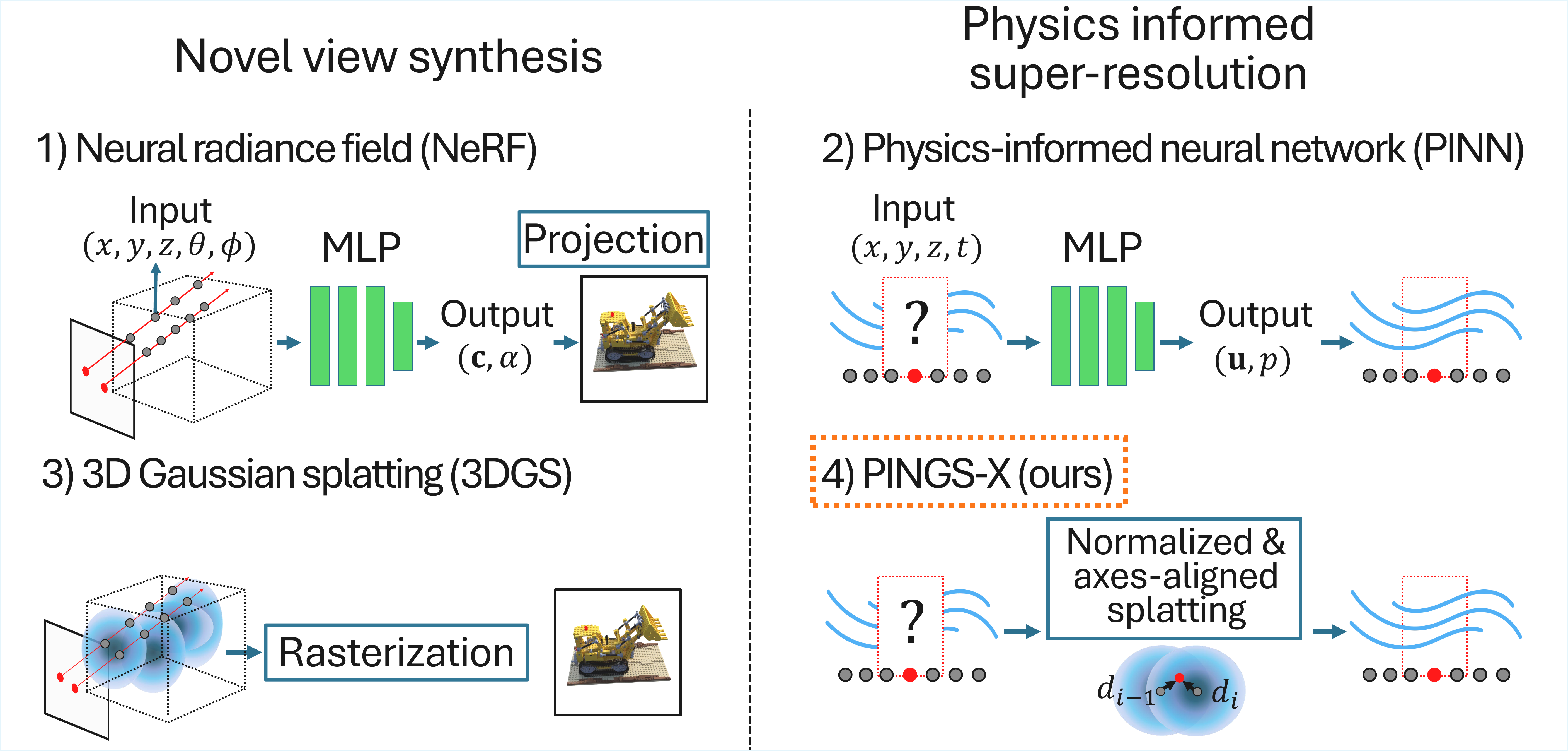}
    \small
    \caption{
        Illustration of the architectural parallel between neural rendering and physics-informed learning. Our work is motivated by the shared use of slow implicit representations (NeRFs and PINNs) and aims to transfer the efficiency of the explicit 3DGS~\cite{3dgs} framework to our domain.
        }
    \figlabel{nerf_pinn_comparison}
\end{figure}

\begin{figure*}[t]
\centering
    \includegraphics[width=1.0\textwidth]{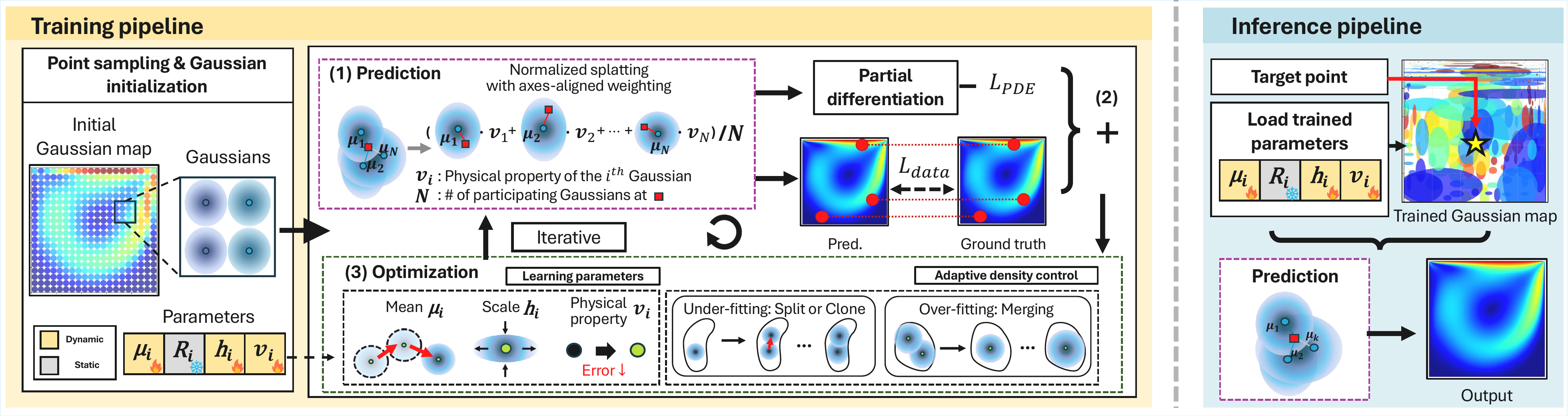}
    \caption{
        Our PINGS-X framework. For training, we initialize (axes-aligned) spatiotemporal Gaussians from low-resolution data and iteratively optimize them. 
        In each step, we (1) predict values using Normalized Gaussian Splatting (NGS), (2) compute a combined data ($L_{data}$) and physics ($L_{PDE}$) loss, and (3) update the Gaussians parameters.
        The Gaussian density is adjusted via splitting, cloning, and our proposed \emph{merging}. 
        At inference, we use the trained Gaussians for prediction via NGS.
    }
    \label{fig:framework}
\end{figure*}

\section{Related work}


\subsubsection{Super-resolution for 4D Flow MRI.}
Methods to enhance the resolution of 4D flow MRI fall into two main categories.

First, data-driven deep learning, uses architectures like CNNs to learn direct low-to-high-resolution mappings from paired training data~\cite{shit2022srflow, ferdian2023cerebrovascular, long2023super}. While powerful, their reliance on large and diverse training datasets is a significant practical challenge, leading to concerns about generalization to unseen cardiovascular domains~\cite{ericsson2024generalized}.

Second, Physics-Informed Neural Networks (PINNs) offer a compelling alternative by embedding governing laws like the Navier-Stokes equations into the optimization process~\cite{raissi2019physics, fathi2020super}. This allows them to be trained on a single 4D flow MRI scan without requiring large datasets~\cite{fathi2020super, saitta2024implicit}.
A known synergy exists with neural rendering; for example, the \emph{Siren} activation function is now often used in PINNs to better represent complex physical fields.
Despite this, the fundamental reliance of PINNs on a slow-to-train implicit MLP creates a major computational bottleneck for high-dimensional 4D flow MRI data, limiting their clinical feasibility~\cite{saitta2024implicit, callmer2025temporal, cho2023separable}.

\subsubsection{The shift toward explicit representations in rendering.}
The computational bottleneck in PINNs mirrors a similar challenge recently overcome in novel view synthesis. Early methods like neural radiance fields (NeRFs) were also limited by slow, implicit MLP-based models~\cite{nerf}. This limitation was largely solved by 3D Gaussian splatting, which achieved state-of-the-art results with orders-of-magnitude faster training by using an explicit set of 3D Gaussians~\cite{3dgs}. This successful transition from a slow implicit to a fast explicit representation provides the primary motivation for our work: to bring similar efficiency gains to physics-informed super-resolution.

\subsubsection{Physics-informed Gaussian representations.}
A few recent studies have explored the intersection of physics and Gaussian representations, but each has key limitations that our work aims to address. Physics-Informed Gaussian Splatting (PIGS)~\cite{rensen2024pigs} models time-resolved physical fields by moving 3D Gaussians through the time domain. However, its use of unnormalized Gaussian weights means it lacks theoretical convergence guarantees, which, as our experiments will show, can lead to inferior prediction accuracy. In contrast, Physics-Informed Gaussians (PIG)~\cite{kang2024pig} uses a lightweight MLP to process a Gaussian feature embedding. However, by retaining a neural network in its prediction path, it remains an implicit method and, as we found in our experiments, fundamentally shares the same computational overhead as standard PINNs.

\subsubsection{Summary and positioning.} 
A need remains for a theoretically grounded Gaussian splatting approach for efficient physical-field super-resolution. Our work addresses this gap.

\section{Preliminaries}


\subsection{4D flow MRI and governing equations}
\label{sec:4df_mri_background}

\subsubsection{Acquisition principle.}
4D~flow MRI is an extension of phase‑contrast MRI that encodes velocity‐sensitive phase shifts into the MR signal along three orthogonal directions.
For each cardiac phase, the scanner acquires four complex k‑space volumes—one reference and three flow‑encoded scans.
Typical exams capture $20$–$30$ cardiac frames at 1--3\,mm isotropic spatial resolution.  
After inverse Fourier reconstruction, phase unwrapping, and velocity scaling, these volumes provide a time‑resolved velocity field  
$\mathbf{v}(x,y,z,t)=[v_x,v_y,v_z]^{\top}$ defined on the voxel lattice.

\subsubsection{Governing partial differential equations (PDEs).}
Blood flow in large arteries and veins is often approximated as an incompressible Newtonian fluid~\cite{lynch2022effects}.
Subsequently, the velocity field obeys the incompressible Navier–Stokes equation in Eq.~\eqref{navier_stokes}, which describe the motion of a viscous fluid, and the continuity equation, which describes mass conservation in Eq.~\eqref{continuity}.
\begin{align}
\frac{\partial\mathbf{v}}{\partial t}+(\mathbf{v}\cdot\nabla)\mathbf{v}
      &= -\frac{1}{\rho}\nabla p + \nu\nabla^{2}\mathbf{v} + \mathbf{g},
      \label{eq:navier_stokes} 
      \\
\nabla\cdot\mathbf{v} &= 0,                                   \label{eq:continuity}
\end{align}
where $p(x,y,z,t)$ is pressure, $\rho$ the fluid density, $\nu$ the kinematic viscosity, and $\v g$ is the gravity.
Eq.~\eqref{navier_stokes} and Eq.~\eqref{continuity} supply the physics‑informed loss terms later used in training.

\subsubsection{Spatial averaging.}
Each voxel in the acquired velocity field represents a spatial average of all sub-voxel velocities, leading to a loss of fine-scale details.
This motivates the need for super‑resolution methods that recover fine‐grained velocities consistent with Eq.~\eqref{navier_stokes} and Eq.~\eqref{continuity}.

\subsection{3D Gaussian splatting in novel view synthesis}
\label{sec:3dgs_background}

As shown in Fig.~\ref{fig:nerf_pinn_comparison},
3DGS represents a scene with a sparse $N$ set of anisotropic 3D Gaussians $\{ \mathcal{G}_i \}$.
Each Gaussian has center (mean)~${\boldsymbol \mu}_i \in \real^3$,
covariance $\m \Sigma_i = \m R_i\,\mathrm{diag}(h_{i1}^2,h_{i2}^2,h_{i3}^2)\,\m R_i^{\!\top}$,
color $\mathbf{c}_i\!\in\!\real^3$, and opacity $0\leq\alpha_i\leq1$, where $\{h_{ij} \}$ represent the scales of $\mathcal G_i$ along its principal axes and $\m R_i\in\!SO(3)$ represent its rotation.

\subsubsection{Rasterization.}  
After perspective projection, the $i$-th Gaussian appears on the image
plane with 2D mean $\boldsymbol{\mu}'_i$ and covariance
$\m \Sigma'_i$.
For a pixel at screen coordinate $\v x\in\real^2$, Gaussians are sorted
front‑to‑back and blended by closed‑form alpha‑compositing as follows:
\begin{align}
\tilde\alpha_i(\v x)
      &= \alpha_i \exp\! \left( -\tfrac12(\v x-\boldsymbol{\mu}'_i)^{\!\top}
                              \m \Sigma'_i(\v x-\boldsymbol{\mu}'_i) \right), \\
\widehat{\v c}(\v x) &= \sum_{i=1}^{N} \mathbf{c}_i\,\tilde\alpha_i(\v x)
                \prod_{j<i}\bigl(1-\tilde\alpha_j(\v x)\bigr),
\end{align}
where $\widehat{\v c}:\real^3\rightarrow\real^3$ is the rendered RGB color at  pixel $\v x$.

\subsubsection{Optimization pipeline.}  
Training alternates between two stages, namely  
(i) joint gradient-based optimization of Gaussian variables $\{\boldsymbol\mu_i,\m \Sigma_i,\mathbf{c}_i,\alpha_i\}$, and
(ii) adaptive density control to \textit{split/clone} Gaussians in
high‑error regions and \textit{prune} low‑contribution Gaussians.
This yields NeRF‑level quality with training times
orders‑of‑magnitude shorter.

\section{Proposed method}
\label{sec:proposed_method}

We introduce PINGS-X, a framework for physics-informed super-resolution that leverages the explicit representation of 3D Gaussian Splatting. As illustrated in Fig.~\ref{fig:framework}, our approach is built upon three key innovations detailed in this section: Normalized Gaussian Splatting (NGS), the use of axes-aligned Gaussians, and a Gaussian merging procedure.

\subsection{Normalized Gaussian splatting and its convergence}

\subsubsection{Motivation.}
A straightforward application of 3DGS to model a physical field (e.g. velocity and pressure) $\widehat{\v v}:\real^q \rightarrow \real^p$ at any spatiotemporal point~$\mathbf{x}\in\mathbb{R}^q$ is to use an (unnormalized) weighted sum of $N$ Gaussians $\{\mathcal G_i\}$ each with physical property $\v v_i\in\real^p$, center $\boldsymbol \mu_i\in\real^q$ and covariance $\m \Sigma_i := \m R_i\,\mathrm{diag}(h_{i1}^2,\cdots,h_{iq}^2)\,\m R_i^{\!\top}$ ($\m R_i\in\,SO(q)$ is the rotation and $h_{ij}$ is the Gaussian scale about the $j$-th axis) as follows:
\begin{align}
    \label{eq:unnormalized_sum}
    \widehat{\mathbf{v}}(\mathbf{x}) = \sum_{i=1}^{N} z_i(\mathbf{x}) \mathbf{v}_i,
\end{align}
where $z_i(\mathbf{x}) = \exp\left(-\frac{1}{2} (\mathbf{x} - \boldsymbol{\mu}_i)^\top \m \Sigma_i^{-1} (\mathbf{x} - \boldsymbol{\mu}_i) \right)$ can be viewed as the (unnormalized) influence of $\mathcal G_i$. 
This formulation similarly used in PIGS~\cite{rensen2024pigs} is simpler than the original 3DGS rendering equation as it is a direct evaluation in the spatiotemporal domain and thus omits the projection and alpha compositing steps.
However, the unnormalized sum above has a fundamental limitation: it is not a \emph{well-behaved} prediction scheme.

In regions far from any Gaussian center, the prediction $\widehat{\v v}(\v x)$ collapses to zero (Fig.~\ref{fig:1d_visualization}).
To compensate, the model must adopt inefficient strategies, such as learning overly large Gaussians that cause over-smoothing, or constantly adding new Gaussians which can create oscillatory behavior. This instability is further evidenced in our Rosenbrock experiment (Table~\ref{table:rotated_pigs_pings} and Fig.~\ref{fig:rosenbrock}), 
and we also provide a theory confirming that convergence of $\widehat{\v v}$ is not guaranteed, as a distant $\textbf{x}$ 
makes the prediction collapse to zero unless the number of Gaussians is adjusted (Supplementary B). 




\subsubsection{Our solution.}
\begin{figure}[t]%
\includegraphics[width=\columnwidth]{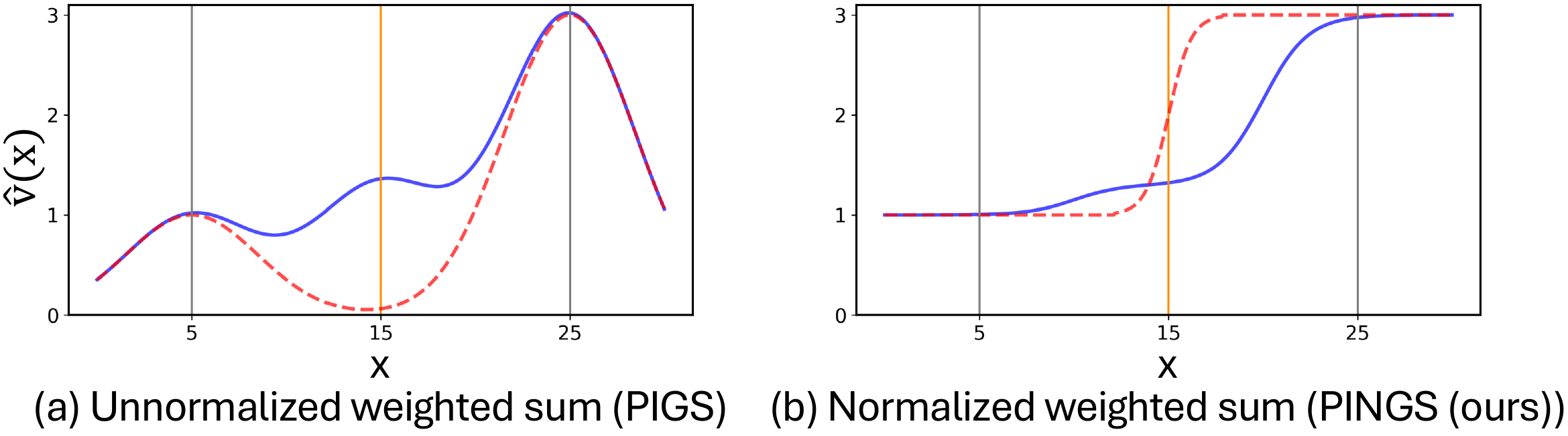}
\small
\caption{
    A 1D visualization comparing unnormalized and normalized sums of Gaussians ($\sigma=12$).
    (a) Unnormalized sum: with only two Gaussians ($x=\{5, 25\}$, $v=\{1, 3\}$), the prediction (dotted red) decays to zero in the middle. Adding a third Gaussian at $x=15$, $v=1.3$ (orange) to fill this gap results in an oscillatory profile (solid blue).
    (b) Normalized sum: we observe a smooth prediction curve between the two Gaussians. 
    Adding the third Gaussian adjusts the curve while maintaining a stable, monotonic result.
}
\label{fig:1d_visualization}
\end{figure}

To address this issue, we introduce Normalized Gaussian splatting (NGS), which redefines the prediction by normalizing the Gaussian weights:
\begin{align}
\label{eq:normalized_sum}
\widehat{\mathbf{v}}(\mathbf{x}) = \sum_{i=1}^{N} w_i(\mathbf{x}) \mathbf{v}_i, ~~\text{where}~~ w_i(\mathbf{x}) = \frac{z_i(\mathbf{x})}{\sum_{j=1}^{N} z_j(\mathbf{x})}.
\end{align}
This ensures the prediction is a convex combination of the Gaussian properties, providing inherent stability as the predicted value is always bounded by the minimum and maximum of the values from the influencing Gaussians. 
Consequently, adding new Gaussians serves as a principled refinement—placing new ``control points'' to capture finer details instead of a corrective measure (Fig.~\ref{fig:1d_visualization}). 
Such stability is the basis for our method's universal approximation capability, which we formalize in the following theorem.

\subsubsection{Universal approximation property: consistency.}
\label{sec:universal}
We formally establish the asymptotic convergence in a statistical sense. 
The convergence rate is explicitly derived in Theorem~\ref{th:univ_appox}.
We provide the proof in the supplementary material (B).

\begin{theorem}
Suppose $(\boldsymbol{\mu}_i,\v v_i)\in\mathbb{R}^{q+p}$, $i=1,\ldots, N$, $q, p\in\mathbb{N}$, are samples from the random vector $(\boldsymbol{\mu},\v v)$, 
whose cumulative distribution function is absolutely continuous with respect to the Lebesgue measure with density $f$. We assume:
\begin{list}{(\roman{enumi})}
  {\usecounter{enumi}
   \setlength{\leftmargin}{1.15em}
   \setlength{\itemindent}{0em}
   \setlength{\labelsep}{0.4em}
   \setlength{\labelwidth}{1.1em}
  }
  \item[(i)] (i.i.d) The samples $(\boldsymbol{\mu}_1,\v v_1), (\boldsymbol{\mu}_2,\v v_2),\ldots, (\boldsymbol{\mu}_N,\v v_N)$ are independent and identically distributed.
\item[(ii)](Smoothness condition) Both $\v v$ and $f$ are $\beta$-times continuously differentiable in some neighborhood of ${\bf{x}}\in\mathbb{R}^q$.
\item[(iii)] (Bandwidth conditions) The matrix $\m \Sigma_i=\m \Sigma_i(N)$ is a sequence of bandwidth matrices such that $[tr(\m \Sigma_i)]^{\beta/2}\rightarrow 0$ and $N\cdot det(\m \Sigma_i)^{1/2}\rightarrow\infty$ as $N\rightarrow\infty$ for all $i$. 
\end{list}
Then, for any ${\bf{x}}\in\mathbb{R}^q$, we have the following convergence:
\begin{align}
\widehat{\mathbf{v}}({\bf{x}})~-~ \mathbf{v}({\bf{x}})~&\rightarrow~0
\end{align}
in probability, as the number of Gaussians $N\rightarrow\infty$, where $\mathbf{v}({\bf{x}})=\mathbb{E}[\v v|\boldsymbol{\mu}={\bf{x}}]$. Specifically, we have
\begin{equation} \scalebox{0.793}{ 
$\displaystyle \widehat{\mathbf{v}}({\bf{x}})-\mathbf{v}({\bf{x}}) = O_p\left(\frac{1}{N}\sum_{i=1}^{N}\big[\operatorname{tr}(\m \Sigma_i)\big]^{\frac{\beta}{2}}+\sqrt{\frac{1}{N^2}\sum_{i=1}^{N}\frac{1}{\operatorname{det}(\m \Sigma_i)^{1/2}}}\right) $ } \label{rate}
\end{equation}
where $O_p(\cdot)$ denotes the stochastic Big-O notation.
\label{th:univ_appox}
\end{theorem}

\subsubsection{Remark.}
 NGS can be viewed as transforming the classic (static) Nadaraya-Watson (NW) estimator~\cite{nadaraya1964estimating,watson1964smooth} into a dynamic, learnable framework. 
 Unlike the static nature of NW, which uses fixed points and a global bandwidth, our approach optimizes the Gaussian centers ($\boldsymbol{\mu}_i$) and learns an individual anisotropic covariance/bandwidth ($\m \Sigma_i$) for each kernel. 
 Coupled with adaptive density control to add or remove Gaussians, this creates a notably more flexible representation for complex physical fields.


\subsection{Axes-aligned splatting for efficient super-resolution}

\subsubsection{Motivation.}
Optimizing the full anisotropic covariance of NGS becomes prohibitive in high-dimensional domains like 4D spatiotemporal space. The primary challenge is parameterizing the higher-dimensional (4D) rotation of each Gaussian, which is non-trivial to optimize with gradient descent.

\subsubsection{Our solution.}
To circumvent this representational complexity, we constrain the Gaussians to be axes-aligned, meaning their covariance matrices are diagonal, i.e. $\m \Sigma_i =diag(h_{i1}^2,\cdots,h_{iq}^2)$.
This provides a simpler and more direct representation that avoids the need to parameterize complex 4D rotations. As a secondary benefit, this also reduces the number of covariance parameters from 10 to 4 for each Gaussian, improving computational efficiency.

\begin{figure}[t]%
\small
\includegraphics[width=\columnwidth]{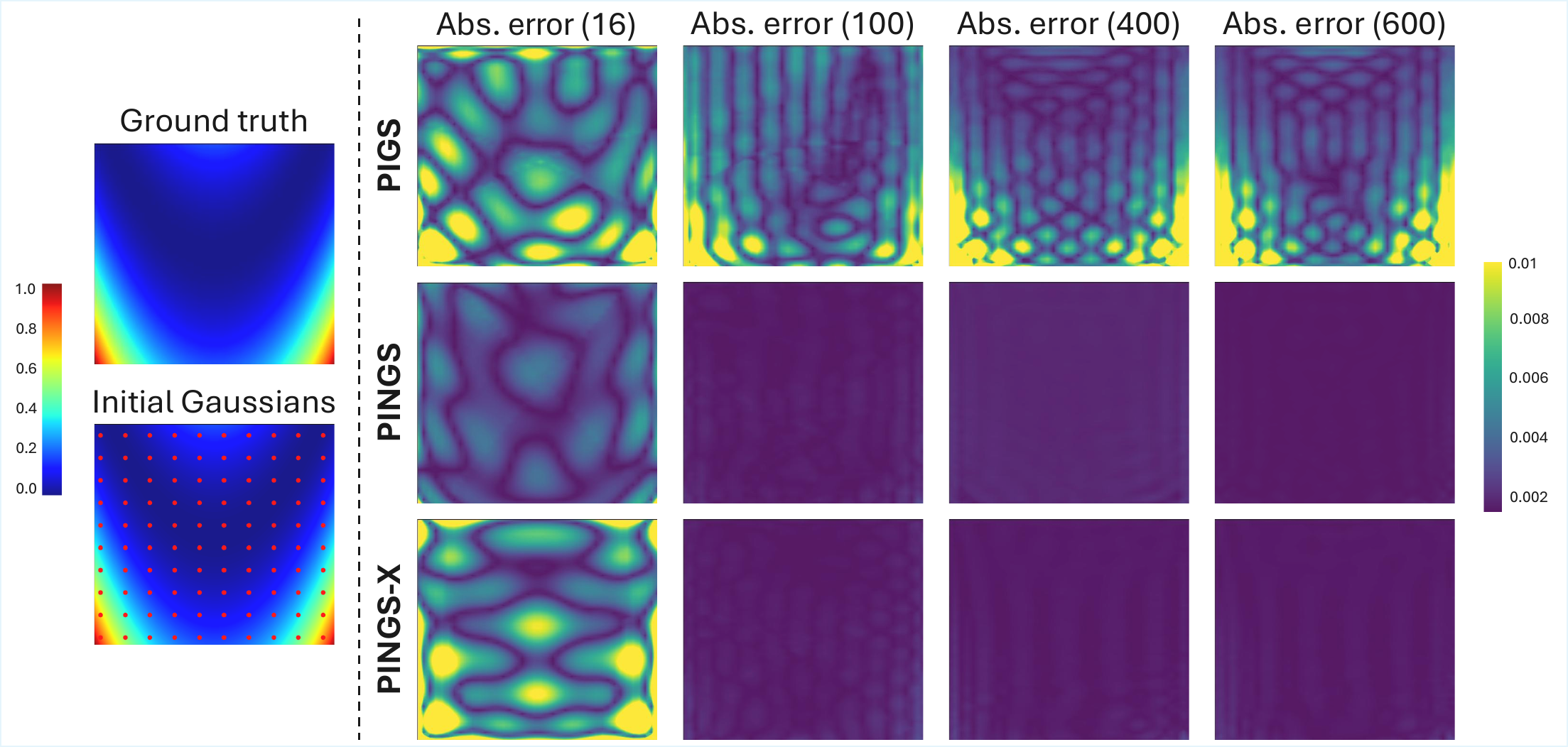}
\caption{
    Absolute error maps for the Rosenbrock function (yellow indicates higher error). 
    Our PINGS/PINGS-X converges as the number of Gaussians (in parentheses) increases, while (unnormalized) PIGS does not.
}
\label{fig:rosenbrock}
\end{figure}

This approach, which we term Normalized and Axes-aligned Gaussian Splatting (NGS-X), preserves the favorable convergence properties of our general method—a claim we validate theoretically below and empirically in Fig.~\ref{fig:rosenbrock}.
\begin{corollary}
In addition to the same assumptions used in Theorem 1, if we further suppose $\m \Sigma_i$ is diagonal with entries $h_{i1}^2,h_{i2}^2,\ldots,h_{iq}^2$ for each $i$, the expression (\ref{rate}) simplifies to
\begin{equation} \scalebox{0.8}{ 
$\displaystyle \widehat{\mathbf{v}}({\bf{x}})-\mathbf{v}({\bf{x}}) = O_p\left(\frac{1}{N}\sum_{i=1}^{N}\sum_{j=1}^{q}h_{ij}^{\beta}+\sqrt{\frac{1}{N^2}\sum_{i=1}^{N}\frac{1}{h_{i1}\cdots h_{iq}}}\right). $ } \label{rratesp}
\end{equation}
\end{corollary}

\noindent\textbf{Remark.} In the case where each bandwidth $h_{ij}$ is of same order of magnitude, i.e. $h_{ij}=O(h)$ $\forall~i,j$, using the bias-variance balancing bandwidth, 
we attain the best possible convergence rate in the minimax sense~\cite{stone1980,stone1982}.


\begin{table}[t]
\centering
\small  
\setlength{\tabcolsep}{1mm} 
\begin{tabular}{c|c|ccc}
\hline\hline
 \# Gauss. & Method & $t$~(s)~$\downarrow$ & Rel. err.~(\%) $\downarrow$ & RMSE~$\downarrow$\\ 
\hline\hline
\multirow{3}{*}{16}
 & PIGS & \underline{61.1} & 3.38 & 7.47e-03 \\
 & PINGS~(ours) & 66.5 & \textbf{1.00} & \textbf{2.22e-03} \\
 & PINGS-X~(ours) & \textbf{48.4}& \underline{3.29} & \underline{7.28e-03} \\
\hline
\multirow{3}{*}{100}
 & PIGS  & \underline{58.2} & 2.45 & 5.42e-03 \\
 & PINGS~(ours) & 68.7 & \textbf{0.10} & \textbf{2.17e-04} \\ 
 & PINGS-X~(ours) & \textbf{50.1} & \underline{0.14} & \underline{3.01e-04} \\
\hline
\multirow{3}{*}{400}
 & PIGS  & \underline{68.2} & 3.14 & 6.94e-03 \\
 & PINGS~(ours) & 74.5 & \underline{0.26} & \underline{5.83e-04} \\ 
 & PINGS-X~(ours) & \textbf{54.2}& \textbf{0.09} & \textbf{1.92e-04} \\
\hline
\multirow{3}{*}{600}
 & PIGS  & \underline{72.6} & 3.53 & 7.82e-03 \\
 & PINGS~(ours)& 80.1 & \textbf{0.04} & \textbf{7.95e-05} \\ 
 & PINGS-X~(ours)& \textbf{62.9}& \underline{0.08} & \underline{1.73e-04} \\
\hline\hline
\end{tabular}
\caption{
    Results for fitting the Rosenbrock function $f(x,y) = (1 - x)^2 + 100 (x - y^2)^2$, analyzing only the effect of the number of Gaussians ($N$) without position updates and adaptive density control of Gaussians. The error for the unnormalized sum (PIGS) remains high and unstable, while our normalized methods (PINGS/PINGS-X) demonstrate clear convergence as $N$ increases. 
}
\label{table:rotated_pigs_pings}
\end{table}

\subsection{Gaussian merging for scalable training}
\label{sec:adap_dc}

\subsubsection{Motivation.}
In 3DGS, the training loss is computed on 2D renderings from multiple camera poses, meaning overlapping Gaussians are implicitly disambiguated by multi-view geometric constraints. Our framework is different, as we directly approximate a high-dimensional physical field without projection or alpha-compositing. With distance-based kernels, this can lead to \emph{degenerate solutions}, where multiple Gaussians accumulate in the same region to represent a single smooth profile. 
Such duplicates inflate memory and slow optimization without improving accuracy. 

Note, pruning is ineffective in our scenario as these redundant Gaussians can be equally impactful. Hence, the appropriate strategy is to \emph{merge} them into a single representative.

\subsubsection{Our solution.}
Our merging procedure identifies Gaussians with similar influence on the prediction. For each Gaussian $\mathcal{G}_i$, we first compute its normalized influence vector $\mathbf{w}_i=[w_i(\mathbf{x}_1),\dots,w_i(\mathbf{x}_K)]^{\!\top}$ at all $K$ training points. We then build an undirected graph where nodes are Gaussians and an edge connects any pair $(i, j)$ whose influence vectors have a cosine similarity exceeding a threshold (0.9).

The connected components of this graph identify clusters of redundant Gaussians, which are then merged in a single pass. The new Gaussian's mean $\boldsymbol \mu$ is the average of the merged means, and its physical property is the predicted field value at that new mean, $\widehat{\v v}(\boldsymbol \mu)$. This merge step is invoked periodically during training (e.g., every 100 epochs), improving training stability and efficiency (Table~\ref{table:Ablation_PDE}).

\subsection{Training procedure}
\label{sec:trainig}

\subsubsection{Initialization of Gaussians.}

We initialize the Gaussians $\{\mathcal G_i\}$ using a simple grid-based sampling strategy to ensure uniform coverage of the spatiotemporal domain.  
Each physical property $\mathbf{v}_i \in \mathbb{R}^p$ is set to the value of the nearest low-resolution data point to the initial mean $\boldsymbol{\mu}_i$.
This straightforward initialization was chosen to keep the setup simple and avoid dependence on complex, data-specific heuristics.

\subsubsection{Loss function.}
We train our model with a composite loss function $L$ that combines a data fidelity term, $L_{data}$, with a physics-informed regularizer, $L_{PDE}$ to yield PINGS-X:
\begin{align}
    L &= L_{data} + \lambda L_{PDE}, \\
    L_{data} &= \frac{1}{K} \sum_{k=1}^{K} 
    \| \boldsymbol \omega \odot (\widehat{\v v}(\v x_k) - \v v_k) \|^2
    \label{eq:loss_data} \\
    L_{PDE} &= \frac{1}{K} \sum_{k=1}^K \sum_{m=1}^M \| \v g_m (\widehat{\v v}(\v x_k)) |^2, \label{eq:loss_pde}
\end{align}
where $L_{data}$ is essentially the mean squared error between the prediction $\widehat{\v v}$ and the low-resolution observation $\v v_k$. 
The binary vector $\boldsymbol{\omega} \in \mathbb{R}^p$ masks unobserved components of the physical properties; for 4D flow MRI, we use it to exclude the pressure term which is not directly measured. 
$L_{PDE}$ is the mean squared error of $M$ physical constraints, $\{ \v g_m \}$. For 4D flow MRI, these are based on the dimensionless form of Eq.~\eqref{navier_stokes} and Eq.~\eqref{continuity}. from the dimensionless Navier-Stokes equations (Supplementary D.1) for which we set $\lambda = 1.0$.

Note, unlike PINNs, which compute derivatives for $L_{PDE}$ via computationally expensive backpropagation, our explicit formulation allows for their efficient, analytical calculation. This avoids the compounding cost of automatic differentiation, making our training process significantly faster.

\subsubsection{Boundary conditions.}
Boundary conditions are implicitly enforced by sampling training data on boundaries.

\section{Experimental results and discussions}

This section experimentally evaluates PINGS-X, detailing the setup, presenting results on synthetic 2D CFD and real 4D flow MRI datasets, and including an ablation study.

\subsection{Datasets}
\label{sec:dataset}

We evaluate our method on both synthetic CFD and real 4D flow MRI datasets. 
For both cases, low-resolution data was generated via spatial averaging of the original high-resolution fields. 
Further details on the datasets and the spatial averaging process are in the supplementary material (C).

\subsubsection{Synthetic CFD datasets.} 

We used three steady-state, incompressible 2D flow simulations: a \emph{lid-driven cavity}, \emph{L-shaped} and \emph{Y-shaped} channels with the last mimicking vascular geometries. 
High-resolution ground truth velocity and pressure fields were generated using standard CFD solvers (OpenFOAM and StarCCM+) with appropriate boundary conditions. 
Low-resolution data was created by spatially averaging these fields, providing a more challenging setup than simple point-sampling. 
For these cases, the model maps 2D coordinates ($q=2$) to a 3D physical property vector containing velocity and pressure ($v_x, v_y, p$), hence $p=3$.



\begin{table}[t]
    \centering
    \small
    \setlength{\tabcolsep}{1mm}
        \centering
        \begin{tabular}{
    c|
    r r|
    r r|
    r r
}
            \hline\hline
            \multirow{3}{*}{\makecell{Methods}} 
                & \multicolumn{2}{c|}{Lid-driven} 
                & \multicolumn{2}{c|}{Y-shaped} 
                & \multicolumn{2}{c}{L-shaped} \\ \cline{2-7}
            & \makecell{$t$ $\downarrow$~\\(min)} & \makecell{Rel. $\downarrow$ \\ err.~(\%)}
            & \makecell{$t$ $\downarrow$~\\(min)} & \makecell{Rel. $\downarrow$ \\ err.~(\%)}
            & \makecell{$t$ $\downarrow$~\\(min)} & \makecell{Rel. $\downarrow$ \\ err.~(\%)} \\ \hline\hline
            Nada.-Watson & $<$1 & 3.03& $<$1 & 5.11& $<$1 & 3.99\\ \hline
            PINN           & \underline{51.4}& 12.20 & \underline{45.0}& 11.54 & \underline{43.6}& 11.84 \\
            Siren          & 59.8& 3.20 & 54.0& 4.34 & 50.3& 2.49 \\
            XPINN (tanh)   & 90.2& 8.47 & 165.0& 8.07 & 132.6& 5.54 \\
            XPINN (sin)    & 105.8& 2.71 & 195.5& \underline{3.89} & 158.0& {2.15} \\
            PIG& 820.1& \underline{2.33}& {292.9}& {7.06}& {297.8}& \underline{1.48}\\ 
            {PINGS-X (ours)} & \textbf{21.9}& \textbf{1.13} & \textbf{4.8}& \textbf{2.62} & \textbf{6.3}& \textbf{1.15} \\ \hline\hline
        \end{tabular}
    \caption{
    Quantitative comparison on synthetic 2D CFD datasets. PINGS-X consistently outperforms other methods.
    }
    \label{tab:Main_NS_Toy}
\end{table}

\begin{table}[t]
    \centering
    \small
    \setlength{\tabcolsep}{4pt} 
    \begin{tabular}{c | rr | rr | rr}
        \hline\hline
        \multirow{3}{*}{\makecell{\# Initial \\ Gaussians}}
            & \multicolumn{2}{c|}{Lid-driven}
            & \multicolumn{2}{c|}{Y-shaped}
            & \multicolumn{2}{c}{L-shaped} \\
        \cline{2-7}
        & \makecell{$t$ $\downarrow$\\ (min)} & \makecell{Rel. $\downarrow$ \\ err.(\%)} 
        & \makecell{$t$ $\downarrow$\\ (min)} & \makecell{Rel. $\downarrow$ \\ err.(\%)} 
        & \makecell{$t$ $\downarrow$\\ (min)} & \makecell{Rel. $\downarrow$ \\ err.(\%)} \\
        \hline\hline
        200  & 35.5 & 1.13 & 5.6 & 2.60 & 6.7 & 1.16 \\
        400  & 22.2 & 1.13 & 5.2 & 2.62 & 6.7 & 1.15 \\
        800  & 9.4  & 1.36 & 5.0 & 2.74 & 6.5 & 1.12 \\
        1200 & 15.7 & 1.15 & 4.7 & 2.51 & 7.0 & 1.17 \\
        1600 & 10.6 & 1.33 & 4.8 & 2.59 & 5.1 & 1.12 \\
        \hline\hline
    \end{tabular}
    
    \caption{
        Sensitivity analysis for the number of initial Gaussians. 
        The performance of PINGS-X is robust across a wide range of initial counts, highlighting the effectiveness of its adaptive density control (split/clone/merge) during training.
    }
    \label{tab:initial_gaussian_main}
\end{table}


\subsubsection{Real 4D flow MRI dataset.} 
We used a high-resolution 4D Flow MRI dataset of a carotid artery phantom that replicates patient vasculature~\cite{ko2019patient}, acquired at 0.35 mm spatial and 25 ms temporal resolution. 
From 34 time frames per cardiac cycle, five frames around the peak systolic phase were used for constructing the 4D dataset. 
From this data, low-resolution inputs for our $\times8$ and $\times64$ super-resolution tasks were generated via spatial averaging. For this task, the model maps 4D spatiotemporal coordinates ($q=4$) to a 4D physical property vector containing the 3D velocity and pressure ($v_x, v_y, v_z, p$), hence $p=4$. 
\begin{figure}[t]%
\centering
    \includegraphics[width=\columnwidth]
    {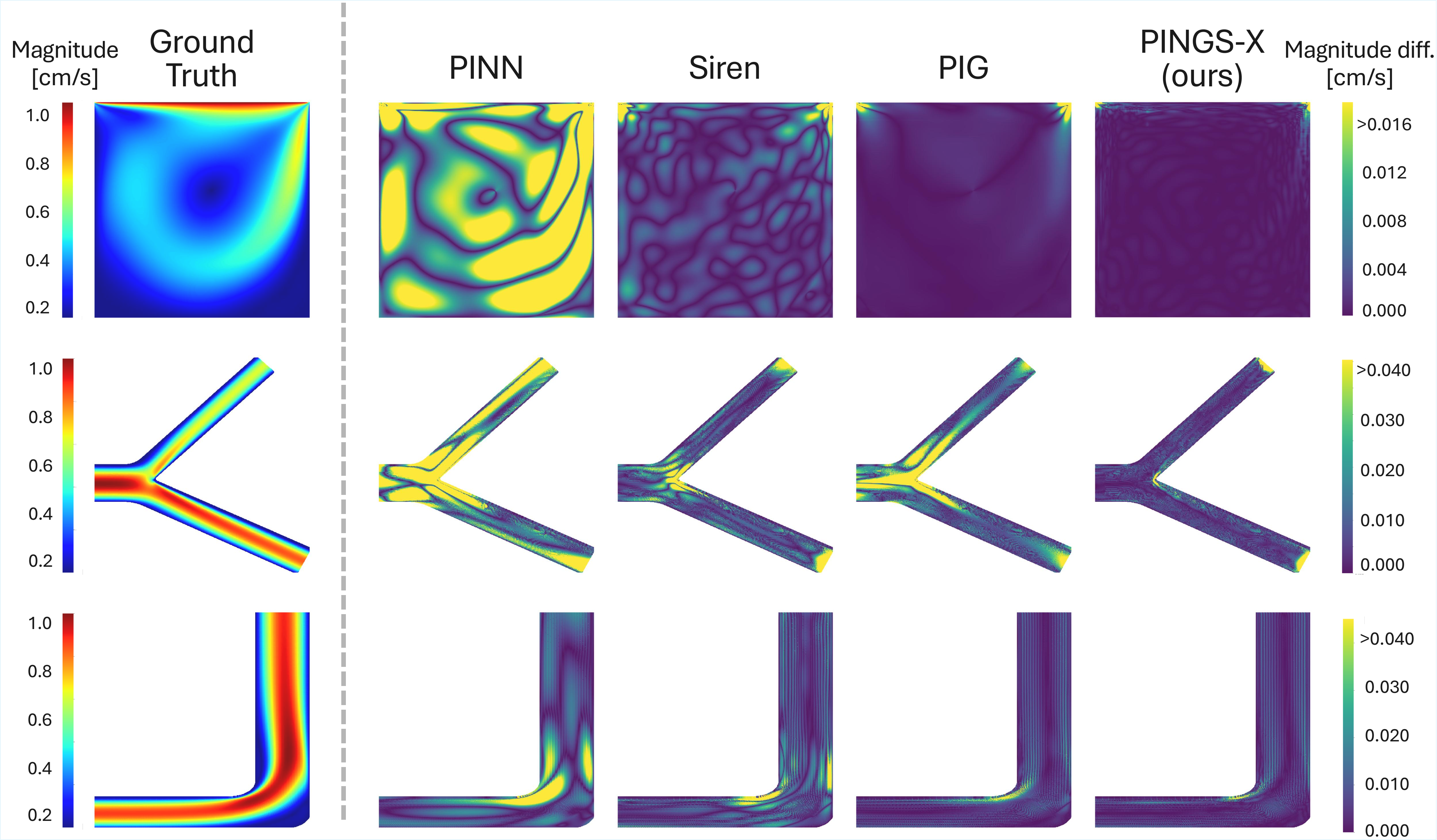}
    \caption{
        Qualitative comparison of ground truth velocity and absolute error maps for the synthetic 2D CFD datasets (Lid-driven, Y-shape, and L-shape). For the error maps (yellow = high error), the color scale is clipped to the mean error of the PINN baseline to aid visual comparison.
    }
    \label{fig:2d_fluid_visualalization}
\end{figure}


\begin{table}[t]
    \centering
    \small
    \setlength{\tabcolsep}{1mm}
    \begin{tabular}{l |rr| rr| rr}
        \hline\hline
        \multirow{3}{*}{Method} & \multicolumn{2}{c|}{Lid-driven} & \multicolumn{2}{c|}{Y-shaped} & \multicolumn{2}{c}{L-shaped} \\
        \cline{2-7}
        & \makecell{$t$ $\downarrow$\\ (min)} & \makecell{Rel. $\downarrow$\\ err.~(\%)}
        & \makecell{$t$ $\downarrow$\\ (min)} & \makecell{Rel. $\downarrow$\\ err.~(\%)}
        & \makecell{$t$ $\downarrow$\\ (hr)}  & \makecell{Rel. $\downarrow$\\ err.~(\%)} \\
        \hline\hline
        PINGS-X & 21.9 & 1.13 & 4.8 & 2.62 & 6.3 & 1.15  \\
        \hline
        w/o norm.       & 23.3 & NC     & 3.4 & NC     & 4.3 & NC  \\
        w/o axes-align. & 70.3 & 1.01   & 23.3 & 2.27  & 22.7 & 1.13   \\
        w/o merging     & $^*$12.1 & $^*$1.38 & $^*$35.1 & $^*$1.66 & $^*$32.8 & $^*$1.12  \\
        \hline\hline
    \end{tabular}
    \caption{
        Ablation study of PINGS-X components. 
        NC denotes non-convergence. 
        $^*$ indicates experiments terminated early ($<$~2000 epochs) due to out-of-memory (OOM) issues.
    }
    \label{table:Ablation_PDE}
\end{table}

\subsection{Experimental settings}
\label{sec:settings}

\paragraph{Baselines and general configuration.}
For the task of physics-informed super-resolution, we compared our method, PINGS-X, against several representative baselines: a standard PINN~\cite{pinn}, PINN with a Siren activation function~\cite{siren}, XPINN~\cite{xpinn} (using both \emph{tanh} and \emph{Siren} activations), and the recent Physics-Informed Gaussians (PIG)~\cite{kang2024pig}, all of which were run for 100,000 epochs. To evaluate against its theoretical foundation, we also included the classic Nadaraya-Watson (NW) estimator as a non-learnable reference. All models were trained using an Adam optimizer. More details are in the supplementary material (D.4).


\subsubsection{Implementation details.}
For synthetic CFD experiments, PINN, Siren, and each XPINN sub-network shared an architecture of 5 hidden layers with 128 neurons and were trained using a full-batch approach. 
Models with Tanh activation (PINN, XPINN~(tanh)) used a learning rate of $10^{-4}$~\cite{wang2023pinnguide}, while those with Siren activation (Siren, XPINN~(sin)) used $5 \times 10^{-6}$ and $\omega_0{=}10$.
For PIG, we used the official implementation with default settings, while PINGS-X was initialized with 400 Gaussians, trained for 10,000 epochs with a learning rate of $10^{-2}$, and applied splitting/cloning and merging every 100 epochs.

For the real 4D flow MRI dataset, we omitted PIG and XPINN due to high training time and NW as it showed much inferior performance.
PINN and Siren baselines utilized larger MLP with 8 hidden layers and 256 neurons, trained with a batch size of 10,000. The learning rates were identical to the synthetic case, with Siren using $\omega_0{=}20$ for improved stability.
For PINGS-X framework, Gaussians were initialized by downsampling input points to 1/10, yielding 4,863 Gaussians for the $\times$8 task and 753 for $\times$64.
Training was conducted for 1,000 epochs with a learning rate of $10^{-2}$ and batch size 10,000, applying split/clone and merge every 100 epochs. More details are in the supplementary material (D).

\subsubsection{Evaluation metrics}
We evaluate models based on wall-clock training time ($t$) and two error metrics computed over the $K$ high-resolution test points. Our primary metric is the relative $L^2$ error, defined as $\sqrt{{\sum_{k=1}^K \| \widehat{\mathbf{v}}(\v x_k) - \mathbf{v}_k \|_2^2} / {\sum_{k=1}^K \| \mathbf{v}_k \|_2^2}}$. 
For the 4D flow MRI experiments, we also report the root mean squared error (RMSE), $\sqrt{\frac{1}{K}\sum_{k=1}^K \| \widehat{\mathbf{v}}(\v x_k) - \mathbf{v}_k \|_2^2}$, in cm/s. 
In both metrics, $\widehat{\mathbf{v}}(\v x_k)$ is the prediction at $\v x_k$ and $\mathbf{v}_k$ is the corresponding ground truth. For these calculations, we exclude the unknown pressure component so the metrics directly reflect the error in the predicted velocity.


\begin{table}[t]
    \centering\small
    \setlength{\tabcolsep}{1mm}
    \begin{tabular}{
        c|rrr|rrr
    }
        \hline\hline
        \multirow{3}{*}{Method} 
            & \multicolumn{3}{c|}{Carotid ($\times8 = \times2^3 $)}
            & \multicolumn{3}{c}{Carotid ($\times64 = \times4^3$)}\\ 
            \cline{2-7}
        & \makecell{$t~\downarrow$~\\(hr)} & \makecell{Rel.~$\downarrow$\\ err.~(\%)} & \makecell{RMSE~$\downarrow$\\(cm/s)}
        & \makecell{$t~\downarrow$~\\(min)} & \makecell{Rel.~$\downarrow$\\ 
        err.~(\%)} & \makecell{RMSE~$\downarrow$\\(cm/s)} \\
        \hline\hline
        PINN           & {30.1} & 25.93 & 2.75 & 259.0 & 39.70 & 4.21 \\
        Siren          & 30.8 & {10.63} & {1.13} & 299.2 & 18.49 & 1.96 \\
        PINGS-X & \textbf{2.6} &  \textbf{8.98} & \textbf{0.95} & \textbf{3.9} & \textbf{17.59} & \textbf{1.87} \\ 
        \hline\hline
    \end{tabular}
    \caption{
        Quantitative comparison on the 4D flow MRI (carotid) dataset for two spatial averaging settings (×8 and ×64). PINGS-X shows superior performance in both training time and prediction error.
    }
    \label{table:Carotid}
\end{table}

\subsection{Comparative analysis}
\label{sec:results}


\subsubsection{Results on synthetic CFD datasets.}
As shown in Table~\ref{tab:Main_NS_Toy}, PINGS-X consistently achieves the lowest relative $L^2$ error across all synthetic datasets. Notably, this superior accuracy is achieved with a dramatic reduction in training time. For
instance, on the Y-shaped dataset, PINGS-X converges in just 4.8 minutes, making it approximately 40 times faster than the next most accurate model (XPINN~(sin)). Fig.~\ref{fig:2d_fluid_visualalization} also qualitatively demonstrates that PING-X achieved the lowest error among all the baseline methods.

\subsubsection{Results on real 4D Flow MRI dataset.}
\begin{figure}[t]
\centering
    \includegraphics[width=0.95\columnwidth]{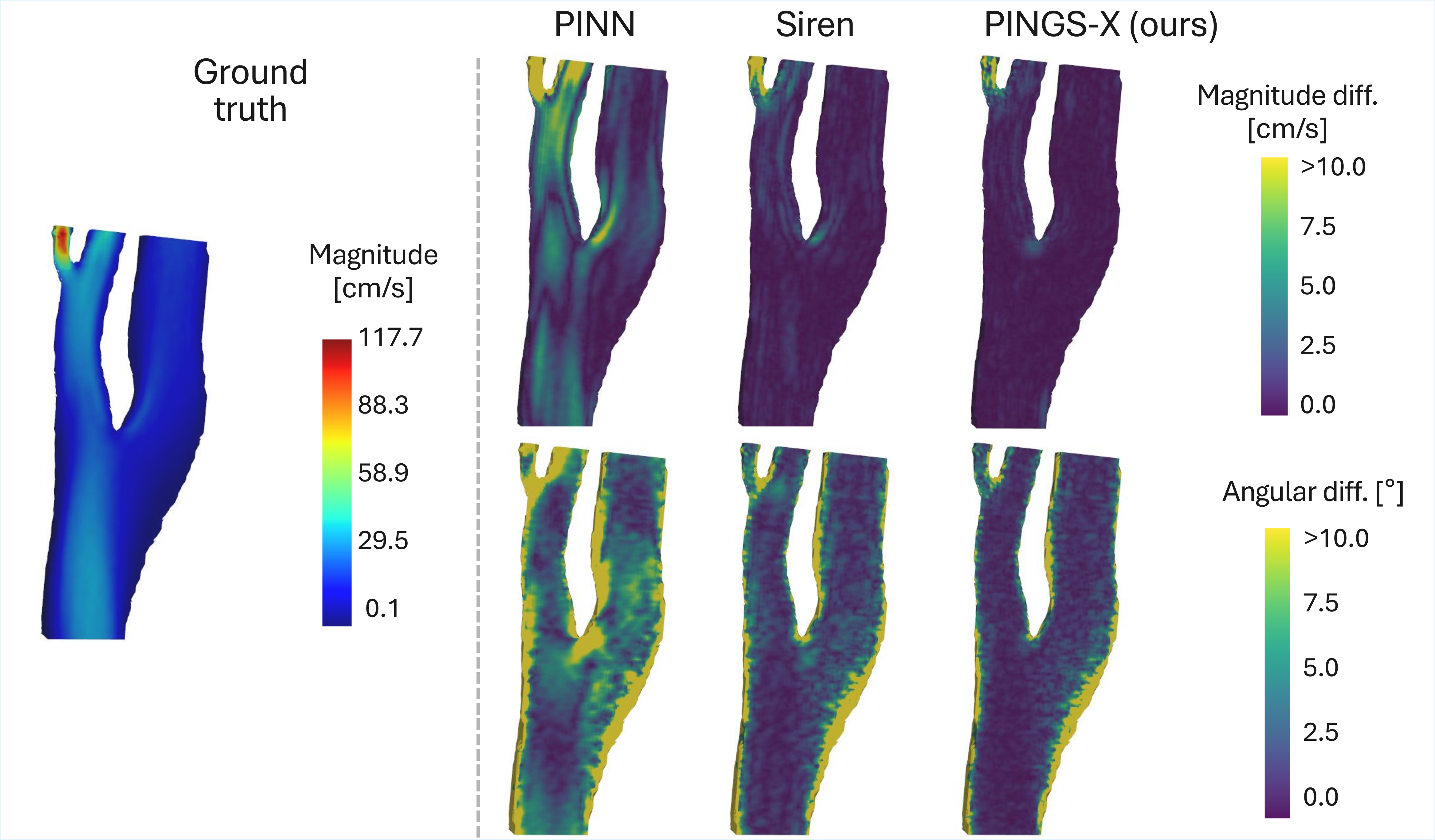}
    \caption{
Visualizations of ground truth velocity field averaged over five time frames and prediction error maps (magnitude and direction) achieved by different methods on the 4D flow MRI dataset for the $\times 8$ spatially averaged data.
    }
    \label{fig:4d_carotid_visualization}
\end{figure}

On the real-world 4D flow MRI dataset, PINGS-X again demonstrates superior performance (Table~\ref{table:Carotid}).
For both the $\times8$ and the more challenging $\times64$ super-resolution tasks, our method converged to a solution with lower error significantly faster than the PINN and Siren baselines. The qualitative results in Fig.~\ref{fig:4d_carotid_visualization} corroborate this, showing error maps for both velocity magnitude and angle that are visibly closer to the ground truth.



\subsection{Ablation study}
\label{sec:ablation}
Our ablation studies (Table~\ref{table:Ablation_PDE}) validate our key design choices. 
Removing normalization prevents the model from converging, confirming its necessity for stability. 
Disabling axes-alignment significantly increases training time for negligible accuracy gain, validating its efficiency benefits. Finally, removing Gaussian merging generally increases both training time and error. 
The effectiveness of our complete adaptive density control (with merging) is further highlighted in Table~\ref{tab:initial_gaussian_main}, which shows that the full PINGS-X framework is robust to the initial number of Gaussians.

\section{Conclusion}

In this work, we introduced \emph{PINGS-X}, a framework for the physics-informed super-resolution of 4D flow MRI data, designed to address the computational expense of PINN-based methods.
Inspired by successes in computer graphics, we developed an explicit, spatiotemporal Gaussian representation with three key innovations: a Normalized Gaussian splatting scheme with theoretical convergence guarantees, an axes-aligned representation for efficient training in (high-dimensional) spatiotemporal space, and a Gaussian merging procedure for robust, scalable training. 
Experiments on synthetic and real-world spatially averaged data show that PINGS-X achieves high accuracy with significantly reduced training time compared to competing methods. By bridging the gap between explicit representations and physics-informed learning, PINGS-X offers a promising direction for efficient and accurate scientific modeling.


\section{Acknowledgments}
This work was partly supported by National Research Foundation of Korea (NRF) grant funded by the Korea government(MSIT) (No.2021R1A2B5B03002103 and No.RS-2022-NR070832), Institute of Information \& communications Technology Planning \& Evaluation (IITP) grant funded by the Korea government(MSIT) (No.RS-2020-II201373, Artificial Intelligence Graduate School Program(Hanyang University)), Hanyang University (HY-202500000003991) and the CoHASS Research Support Grant (No.025637-00001), funded by Nanyang Technological University, Singapore.




\bibliography{main}

\clearpage
\appendix

\setcounter{secnumdepth}{2}

\twocolumn[
\begin{center}
    {\LARGE \textbf{PINGS-X: Physics-Informed Normalized Gaussian Splatting with Axes Alignment for Efficient Super-Resolution of 4D Flow MRI}\par}
    \vspace{1em}
    {\LARGE Supplementary Material\par}
    \vspace{2em}
\end{center}
]

This document provides supplementary material for ``PINGS-X: Physics-Informed Normalized Gaussian Splatting with Axes Alignment for Efficient Super-Resolution of 4D Flow MRI.''
Our code and datasets are publicly available at: \texttt{\url{https://github.com/SpatialAILab/PINGS-X}}.








\subsection*{Table of contents}
\begin{itemize}
    \item[\textbf{A.}] Clarifications
    \item[\textbf{B.}] Proofs of main theoretical results
    \begin{itemize}
        \item[B.1] Asymptotic behavior of the unnormalized sum
        \item[B.2] Proof of Theorem 1 (Convergence of NGS)
        \item[B.3] Proof of Corollary 1 (Conv. of axes-aligned NGS)
    \end{itemize}
    \item[\textbf{C.}] Dataset acquisition and pre-processing
    \begin{itemize}
        \item[C.1] Synthetic 2D CFD datasets
        \item[C.2] Real 4D flow MRI dataset
        \item[C.3] Spatial averaging method
    \end{itemize}
    \item[\textbf{D.}] Additional implementation details
    \begin{itemize}
        \item[D.1] Physics-Informed loss function
        \item[D.2] Gaussian merging algorithm
        \item[D.3] XPINN configuration
        \item[D.4] Hardware and software settings
        \item[D.5] List of hyperparameters
    \end{itemize}
    \item[\textbf{E.}] Limitations and future work
    \item[\textbf{F.}] Additional experimental results
    \begin{itemize}
        \item[F.1] Sensitivity analysis for hyperparameters
        \item[F.2] Effect of different input resolutions        
        \item[F.3] Experimental results on Burgers' equation
    \end{itemize}
    \item[\textbf{G.}] Additional visualizations
    \begin{itemize}
        \item[G.1] Number of Gaussians per epoch  
        \item[G.2] Illustration of adaptive Gaussian dynamics
        \item[G.3] Add. qualitative comparisons on 2D CFD datasets
        \item[G.4] Add. qualitative comparisons on 4D flow MRI dataset
    \end{itemize}
\end{itemize}

\section{Clarifications}

    
\subsubsection{Variance in training time.}
There are very slight differences in the training times reported between the main results (Table 2) and the ablation study (Table 3). This is due to variance in execution time across separate experimental runs. The consistency in the final error metrics confirms that both sets of experiments converged to the same solution.

\section{Proofs of main theoretical results}
\subsection{Asymptotic behavior of the unnormalized sum}
First, note that
\begin{equation} \scalebox{0.88}{ 
$\displaystyle 
\mathbb{E}_\mu\left[ z_i(\mathbf{x}) \, \mathbf{v}_i \right] =
\int \mathbf{v}(\mathbf{u}) \, \exp\left( -\frac{1}{2} (\mathbf{x} - \mathbf{u})' \m \Sigma_i^{-1} (\mathbf{x} - \mathbf{u}) \right) f(\mathbf{u}) d\mathbf{u}. $
}
\nonumber
\end{equation}
We use the change of variables $\mathbf{y} = \m \Sigma_i^{-1/2}(\mathbf{x} - \mathbf{u})$ so that $d\mathbf{u} = \text{det}(\m \Sigma_i)^{1/2} \, d\mathbf{y}$, and Talyor expand $\mathbf{v}(\mathbf{u}) f(\mathbf{u})$ with respect to $\mathbf{x}$. Then the first-order terms integrate to 0 and the second-order terms are $o( | \m \Sigma_i|^{1/2})$. 
This yields $\mathbb{E}[ z_i(\mathbf{x}) \, \mathbf{v}_i] 
= (2\pi)^{q/2} \text{det}(\m \Sigma_i)^{1/2} \, \mathbf{v}(\mathbf{x}) f(\mathbf{x}) 
+ o( \text{det}(\m \Sigma_i )^{1/2} )$, yielding
\begin{equation}
    \widehat{\mathbf{v}}(\mathbf{x})~\approx~ \mathbf{v}(\mathbf{x}) (2\pi)^{q/2} f_{\boldsymbol{\mu}}(\mathbf{x}) \sum_{i=1}^N \text{det}(\m \Sigma_i)^{1/2},
\end{equation}
which is the desired expression. \hfill$\square$\\

\subsubsection{Remark.} The result above formally proves the discussion in our paper that convergence of the prediction $\widehat{\v v}$ is not guaranteed in the unnormalized case. The probability density of $\boldsymbol{\mu}$, denoted $f_{\boldsymbol{\mu}}(\cdot)$, tends to zero as $\textbf{x}$ becomes distant from $\boldsymbol{\mu}$; therefore, unless the number of Gaussians $N$ is sufficiently large, convergence is not guaranteed.

\subsection{Proof of Theorem 1}
For notational convenience, with a slight abuse of notation, we initially use ${\bf{v}}$ to denote an arbitrary $j$th component of ${\bf{v}}\in\mathbb{R}^{p}$, and work in 1-dimension where needed. Also, following the convention, throughout the proof, the bandwidth conditions are understood to hold uniformly in $i\leq N$. Upon noting that
\begin{align}
\mathbf{v}({\bf{x}})&=\mathbb{E}[{\bf{v}}|\boldsymbol{\mu}={\bf{x}}] \nonumber \\
&=\int {\bf{v}}f_{{\bf{v}}|{\bf{x}}}({\bf{v}}|{\bf{x}})d{\bf{v}}~=~\frac{\int {\bf{v}}f_{{\bf{v}},{\bf{x}}}({\bf{x}},{\bf{v}})d{\bf{v}}}{f({\bf{x}})},\label{est}
\end{align}
\onecolumn
we can estimate the numerator of (\ref{est}) by replacing the density $f(\cdot,\cdot)$ in $\int {\bf{v}}f_{{\bf{v}},{\bf{x}}}({\bf{x}},{\bf{v}})d{\bf{v}}$ with its Parzen-Rosenblatt nonparametric estimate $\widehat{f}(\cdot,\cdot)$, yielding $\int {\bf{v}}\widehat{f}_{{\bf{v}},{\bf{x}}}({\bf{x}},{\bf{v}})d{\bf{v}}$, where
\begin{equation}
\widehat{f}_{{\bf{x}},{\bf{v}}}({\bf{x}},{\bf{v}})=\frac{1}{N}\sum_{i=1}^{N}\frac{(2\pi)^{-(q+1)/2}}{\sqrt{s_{i,0}\text{det}(\m \Sigma_i)}}\exp\left(-\frac{1}{2}({\bf{x}}-\boldsymbol{\mu}_i)^{\prime}\m \Sigma_i^{-1}({\bf{x}}-\boldsymbol{\mu}_i)\right)\exp\left(-\frac{1}{2}\frac{({\bf{v}}-{\bf{v}}_i)^2}{s_{i,0}}\right)
\end{equation}
and where $s_{i,0}=s_{i,0}(N)(\rightarrow 0~\text{as}~N\rightarrow\infty)$ is the smoothing bandwidth associated with ${\bf{v}}$.

Consequently, we have for $m_i(\mathbf{x}) = (2\pi)^{-q/2}\exp\left(-\frac{1}{2} (\mathbf{x} - \boldsymbol{\mu}_i)' \m \m \Sigma_i^{-1} (\mathbf{x} - \boldsymbol{\mu}_i) \right)$,
\begin{eqnarray}
\int {\bf{v}}\widehat{f}_{{\bf{x}},v}({\bf{x}},{\bf{v}})d{\bf{v}}&=&\frac{1}{N}\sum_{i=1}^{N}\frac{1}{\sqrt{s_{i,0}\text{det}(\m \Sigma_i)}}\frac{1}{\sqrt{2\pi}}m_i({\bf{x}})\frac{1}{\sqrt{2\pi}}\int {\bf{v}}\exp\left\{-\frac{1}{2}\frac{({\bf{v}}-{\bf{v}}_i)^2}{s_{i,0}}\right\}d{\bf{v}}\notag\\
&=&\frac{1}{N}\sum_{i=1}^{N}\frac{s_{i,0}^{1/2}}{\sqrt{s_{i,0}\text{det}(\m \Sigma_i)}}\frac{1}{\sqrt{2\pi}}m_i({\bf{x}})\int ({\bf{v}}_i+s_{i,0}^{1/2}u)\frac{1}{\sqrt{2\pi}}\exp\left(-\frac{1}{2}u^2\right)du\notag\\
&=&\frac{1}{N}\sum_{i=1}^{N}\frac{1}{\sqrt{\text{det}(\m \Sigma_i)}}\frac{1}{\sqrt{2\pi}}m_i({\bf{x}}){\bf{v}}_i,
\end{eqnarray}
since the Gaussian density integrates to one, and is centred at 0. 

For later reference we propose to use the following as an estimator for the density $f_{\bf{x}}$
\begin{equation}
\widehat{f}_{{\bf{x}}}({\bf{x}})=\frac{1}{N}\sum_{i=1}^{N}\frac{(2\pi)^{-q/2}}{\sqrt{\text{det}(\m \Sigma_i)}}\exp\left(-\frac{1}{2}({\bf{x}}-\boldsymbol{\mu}_i)^{\prime}\m \Sigma_i^{-1}({\bf{x}}-\boldsymbol{\mu}_i)\right).
\end{equation}
This can be understood as the multivariate Parzen-Rosenblatt type estimator with a Gaussian multivariate (radial-symmetric) kernel. Proving the consistency is similar to what follows, and is therefore omitted.

Note that in the estimate of $\mathbf{v}({\bf{x}})=\mathbb{E}({\bf{v}}|\mu={\bf{x}})$, i.e.
\begin{equation}
\widehat{\mathbf{v}}({\bf{x}})~:=~\frac{\sum_{i=1}^{N}{\bf{v}}_im_i({\bf{x}})}{\sum_{j=1}^{N}m_j({\bf{x}})}~=~\sum_{i=1}^{N}{\bf{v}}_iw_i({\bf{x}}),\label{vhat}
\end{equation}
the $w_i({\bf{x}})$ component can be viewed as the weight attached to ${\bf{v}}_i$ in the nonparametric regression problem, where the weights are nonnegative and add up to one. We write
\begin{eqnarray}
\widehat{\mathbf{v}}({\bf{x}})-\mathbf{v}({\bf{x}})&=&\frac{(\widehat{\mathbf{v}}({\bf{x}})-\mathbf{v}({\bf{x}}))\widehat{f}_{\bf{x}}({\bf{x}})}{\widehat{f}_{\bf{x}}({\bf{x})}}\notag\\
&=:&\frac{\widehat{\Psi}({\bf{x}})}{\widehat{f}_{\bf{x}}({\bf{x}})}.
\end{eqnarray}
Substituting the ``regression relation" ${\bf{v}}_i={\mathbf{v}}(\boldsymbol{\mu}_i)+\varepsilon_i$
 into (\ref{vhat}), where $\varepsilon_i\sim^{\text{i.i.d}} (0,\m \Sigma^2(\mu_i))$ and is independent of $\mu_i$, we see that
\begin{equation}
\widehat{\Psi}({\bf{x}})~=~\widehat{\Psi}_1({\bf{x}})+\widehat{\Psi}_2({\bf{x}}), 
\end{equation}
where
\begin{equation}
\widehat{\Psi}_1({\bf{x}})~=~\frac{1}{N}\sum_{i=1}^{N}\frac{(2\pi)^{-q/2}}{\sqrt{\text{det}(\m \Sigma_i)}}[{\mathbf{v}}(\boldsymbol{\mu}_i)-\mathbf{v}({\bf{x}})]\exp\left(-\frac{1}{2}({\bf{x}}-\boldsymbol{\mu}_i)^{\prime}\m \Sigma_i^{-1}({\bf{x}}-\boldsymbol{\mu}_i)\right)
\end{equation}
and
\begin{equation}
\widehat{\Psi}_2({\bf{x}})~=~\frac{1}{N}\sum_{i=1}^{N}\frac{(2\pi)^{-q/2}\varepsilon_i}{\sqrt{\text{det}(\m \Sigma_i)}}\exp\left(-\frac{1}{2}({\bf{x}}-\boldsymbol{\mu}_i)^{\prime}\m \Sigma_i^{-1}({\bf{x}}-\boldsymbol{\mu}_i)\right).
\end{equation}
Using the standard change of variable technique and Taylor expansion, we have
\begin{eqnarray}
\mathbb{E}(\widehat{\Psi}_1({\bf{x}}))&=&\frac{1}{N}\sum_{i=1}^{N}\frac{(\text{det}(\m \Sigma_i))^{-1/2}}{(2\pi)^{q/2}}\int f({\bf{x}}_1)[{\mathbf{v}}({\bf{x}}_1)-\mathbf{v}({\bf{x}})]\exp\left(-\frac{1}{2}({\bf{x}}-{\bf{x}_1})^{\prime}\m \Sigma_i^{-1}({\bf{x}}-{\bf{x_1}})\right)d{\bf{x}_1}\quad\notag\\
&=&\frac{1}{N}\sum_{i=1}^{N}\frac{1}{(2\pi)^{q/2}}\int f({\bf{x}}-(\m \Sigma_i)^{1/2}{\bf{r}})[{\mathbf{v}}({\bf{x}}-(\m \Sigma_i)^{1/2}{\bf{r}})-\mathbf{v}({\bf{x}})]\exp\left(-\frac{1}{2}{\bf{r}}^{\prime}{\bf{r}}\right)d{\bf{r}}\quad\notag\\
&=&\frac{1}{N}\sum_{i=1}^{N}\frac{1}{(2\pi)^{q/2}}\int ((\m \Sigma_i)^{1/2}{\bf{r}})^{\prime}\mathcal{D}_f({\bf{x}})((\m \Sigma_i)^{1/2}{\bf{r}})^{\prime}\mathcal{D}_\mathbf{v}({\bf{x}})+\frac{f({\bf{x}})}{2}((\m \Sigma_i)^{1/2}{\bf{r}})^{\prime}\mathcal{H}_\mathbf{v}({\bf{x}})(\m \Sigma_i)^{1/2}{\bf{r}}+\cdots
\quad\notag\\
&=&\frac{1}{N}\sum_{i=1}^{N}\frac{\kappa_2}{2}\text{tr}\left\{2\m \Sigma_i\mathcal{D}_f({\bf{x}})\mathcal{D}_\mathbf{v}({\bf{x}})+\m \Sigma_if({\bf{x}})\mathcal{H}_\mathbf{v}({\bf{x}})\right\}+O\left(\frac{1}{N}\sum_{i=1}^{N}(\text{tr}(\m \Sigma_i))^{3/2}\right)\notag\\
&=&f({\bf{x}})\sum_{i=1}^{q}\frac{\kappa_2}{2}f({\bf{x}})^{-1}\frac{1}{N}\sum_{i=1}^{N}\text{tr}\left\{2\m \Sigma_i\mathcal{D}_f({\bf{x}})\mathcal{D}_\mathbf{v}({\bf{x}})+\m \Sigma_if({\bf{x}})\mathcal{H}_\mathbf{v}({\bf{x}})\right\}+O\left(\frac{1}{N}\sum_{i=1}^{N}(\text{tr}(\m \Sigma_i))^{3/2}\right)\notag\\
&=&\left(\frac{1}{N}\sum_{i=1}^{N}\text{tr}(\m \Sigma_i)\right)
f({\bf{x}})\sum_{i=1}^{q}\mathcal{B}_i({\bf{x}})+O\left(\frac{1}{N}\sum_{i=1}^{N}(\text{tr}(\m \Sigma_i))^{3/2}\right),\label{aresult1}
\end{eqnarray}
where 
\begin{equation*}
\kappa_2~:=~\frac{1}{(2\pi)^{q/2}}\int{\bf{r}}^{\prime}{\bf{r}}\exp\left(-\frac{1}{2}{\bf{r}}^{\prime}{\bf{r}}\right)d{\bf{r}}.\label{result1}
\end{equation*}

Following the convention in nonparametric statistics, we assume here that $V$ and $f$ are twice continuously differentiable, i.e., $\beta = 2$, for simplicity of exposition. Extensions to the general case can be made straightforwardly by considering higher-order Taylor expansions. Note that $\beta$-times continuous differentiability can be replaced by H\"older continuity of order $\beta$.

We now move on to the variance of $\widehat{\Psi}_1({\bf{x}})$. Due to independence, it follows that\\
\begin{eqnarray}
\text{var}(\widehat{\Psi}_1({\bf{x}}))&=&\frac{1}{N^2}\sum_{i=1}^{N}\text{var}\left\{\frac{(2\pi)^{-q/2}}{\sqrt{\text{det}(\m \Sigma_i)}}[{\mathbf{v}}(\boldsymbol{\mu}_i)-\mathbf{v}({\bf{x}})]\exp\left(-\frac{1}{2}({\bf{x}}-\boldsymbol{\mu}_i)^{\prime}\m \Sigma_i^{-1}({\bf{x}}-\boldsymbol{\mu}_i)\right)\right\}\notag\\
&=&\frac{1}{N^2}\sum_{i=1}^{N}\mathbb{E}\left\{\frac{(2\pi)^{-q/2}}{\sqrt{\text{det}(\m \Sigma_i)}}[{\mathbf{v}}(\boldsymbol{\mu}_i)-\mathbf{v}({\bf{x}})]\exp\left(-\frac{1}{2}({\bf{x}}-\boldsymbol{\mu}_i)^{\prime}\m \Sigma_i^{-1}({\bf{x}}-\boldsymbol{\mu}_i)\right)\right\}^2\notag\\
&&\quad- \frac{1}{N^2}\sum_{i=1}^{N}\left\{\mathbb{E}\left[\frac{(2\pi)^{-q/2}}{\sqrt{\text{det}(\m \Sigma_i)}}[{\mathbf{v}}(\boldsymbol{\mu}_i)-\mathbf{v}({\bf{x}})]\exp\left(-\frac{1}{2}({\bf{x}}-\boldsymbol{\mu}_i)^{\prime}\m \Sigma_i^{-1}({\bf{x}}-\boldsymbol{\mu}_i)\right)\right]\right\}^2\notag\\
&=&\frac{(2\pi)^{-q}}{N^2}\sum_{i=1}^{N}\frac{1}{\text{det}(\m \Sigma_i)}\int f({\bf{x}_{1}})[{\mathbf{v}}({\bf{x}}_{1})-\mathbf{v}({\bf{x}})]^2\exp\left(-\frac{1}{2}({\bf{x}}-{\bf{x}_1})^{\prime}\m \Sigma_i^{-1}({\bf{x}}-{\bf{x}_1})\right)d{\bf{x}}_{1}\notag\\
&&\quad- \frac{1}{N^2}\sum_{i=1}^{N}\left\{\frac{(\text{det}(\m \Sigma_i))^{-1/2}}{(2\pi)^{q/2}}\int f({\bf{x}}_1)[{\mathbf{v}}({\bf{x}}_1)-\mathbf{v}({\bf{x}})]\exp\left(-\frac{1}{2}({\bf{x}}-{\bf{x}_1})^{\prime}\m \Sigma_i^{-1}({\bf{x}}-{\bf{x_1}})\right)d{\bf{x}_1}\right\}^2\notag\\
&=&\frac{(2\pi)^{-q}}{N^2}\sum_{i=1}^{N}\frac{1}{\text{det}(\m \Sigma_i)^{1/2}}\int 
f({\bf{x}_{1}}-(\m \Sigma_i)^{1/2}{\bf{r}})\times[{\mathbf{v}}({\bf{x}_{1}}-(\m \Sigma_i)^{1/2}{\bf{r}})-\mathbf{v}({\bf{x}})]^2\exp(\cdot)d{\bf{r}}\notag\\
&&\quad-\frac{1}{N^2}\sum_{i=1}^{N}\left\{\frac{1}{(2\pi)^{q/2}}\int f({\bf{x}}-(\m \Sigma_i)^{1/2}{\bf{r}})[{\mathbf{v}}({\bf{x}}-(\m \Sigma_i)^{1/2}{\bf{r}})-\mathbf{v}({\bf{x}})]\exp\left(-\frac{1}{2}{\bf{r}}^{\prime}{\bf{r}}\right)d{\bf{r}}\right\}^2\quad\notag\\
&=&O\left(\frac{1}{N^2}\sum_{i=1}^{N}\frac{\text{tr}(\m \Sigma_i)}{\text{det}(\m \Sigma_i)^{1/2}}\right)+\text{smaller order terms}.\label{aresult2}
\end{eqnarray}

\noindent From (\ref{aresult1}) and (\ref{aresult2}) we see that 
\begin{align}
&\mathbb{E}\left[\widehat{\Psi}_1({\bf{x}})-\left(\frac{1}{N}\sum_{i=1}^{N}\text{tr}(\m \Sigma_i)\right)
f({\bf{x}})\sum_{i=1}^{q}\mathcal{B}_i({\bf{x}})\right]^2=~\left(\mathbb{E}\left[\widehat{\Psi}_1({\bf{x}})-\left(\frac{1}{N}\sum_{i=1}^{N}\text{tr}(\m \Sigma_i)\right)
f({\bf{x}})\sum_{i=1}^{q}\mathcal{B}_i({\bf{x}})\right]\right)^2\notag\\
&+\text{var}(\widehat{\Psi}_1({\bf{x}}))~=~O\left(\left(\frac{1}{N}\sum_{i=1}^{N}\text{tr}(\m \Sigma_i)^3\right)^2+\frac{1}{N^2}\sum_{i=1}^{N}\frac{\text{tr}(\m \Sigma_i)}{\text{det}(\m \Sigma_i)^{1/2}}\right),\notag
\end{align}
from which we deduce that
\begin{equation}
\widehat{\Psi}_1({\bf{x}})-\left(\frac{1}{N}\sum_{i=1}^{N}\text{tr}(\m \Sigma_i)\right)
f({\bf{x}})\sum_{i=1}^{q}\mathcal{B}_i({\bf{x}})~=~O_p\left(\left(\frac{1}{N}\sum_{i=1}^{N}\text{tr}(\m \Sigma_i)\right)^{3/2}+\sqrt{\frac{1}{N^2}\sum_{i=1}^{N}\frac{\text{tr}(\m \Sigma_i)}{\text{det}(\m \Sigma_i)^{1/2}}}\right).
\end{equation}
Now, on noting that $\mathbb{E}(\widehat{\Psi}_2({\bf{x}}))=0$, we see that it suffices to consider the second moment. Using the law of iterated expectations, independence of the error, and the arguments used previously, we have
\begin{align}
\mathbb{E}\left\{\widehat{\Psi}_2({\bf{x}})^2\right\}&=~\frac{1}{(2\pi)^q}\frac{1}{N^2}\sum_{i=1}^{N}\sum_{j=1}^{N}\frac{1}{\text{det}(\m \Sigma_i)^{1/2}\text{det}(\m \Sigma_j)^{1/2}}\notag\\
&\qquad\times~ \mathbb{E}\left[\varepsilon_i\varepsilon_j\exp\left(-\frac{1}{2}({\bf{x}}-\boldsymbol{\mu}_i)^{\prime}\m \Sigma_i^{-1}({\bf{x}}-\boldsymbol{\mu}_i)\right)\exp\left(-\frac{1}{2}({\bf{x}}-\boldsymbol{\mu}_j)^{\prime}\m \Sigma_i^{-1}({\bf{x}}-\boldsymbol{\mu}_j)\right)\right]\notag\\
&=~\frac{1}{(2\pi)^q}\frac{1}{N^2}\sum_{i=1}^{N}\frac{1}{\text{det}(\m \Sigma_i)}\times~\mathbb{E}\left[\m \Sigma^2(\boldsymbol{\mu}_i)\exp\left(-({\bf{x}}-\boldsymbol{\mu}_i)^{\prime}\m \Sigma_i^{-1}({\bf{x}}-\boldsymbol{\mu}_i)\right)\right]\notag\\
&=~\Omega({\bf{x}})\cdot \frac{1}{N^2}\sum_{i=1}^{N}\frac{1}{\text{det}(\m \Sigma_i)^{1/2}}+ O\left(\frac{1}{N^2}\sum_{i=1}^{N}\frac{\text{tr}(\m \Sigma_i)}{\text{det}(\m \Sigma_i)^{1/2}}\right),
\end{align}
where $\Omega({\bf{x}})=f({\bf{x}})\m \Sigma^2({\bf{x}})(2\pi)^{-q}\int \exp\left(-{\bf{v}}'{\bf{v}}\right)d{\bf{v}}$.
Therefore, for general $\beta$ we have
\begin{equation}
\widehat{\Psi}({\bf{x}})~=~\widehat{\Psi}_1({\bf{x}})+\widehat{\Psi}_2({\bf{x}})~=~O_p\left(\frac{1}{N}\sum_{i=1}^{N}\big[\text{tr}(\m \Sigma_i)\big]^{\beta/2}+\sqrt{\frac{1}{N^2}\sum_{i=1}^{N}\frac{1}{\text{det}(\m \Sigma_i)^{1/2}}}\right).
\end{equation}
Finally, since $\widehat{\mathbf{v}}({\bf{x}})-\mathbf{v}({\bf{x}})=\widehat{\Psi}({\bf{x}})/\widehat{f}({\bf{x}})$, and because we have $\widehat{f}({\bf{x}})=f({\bf{x}})+o_p(1)$, provided that $f(\cdot)>0$, it follows that
\begin{eqnarray}
\widehat{\mathbf{v}}({\bf{x}})-\mathbf{v}({\bf{x}})&=&O_p\left(\frac{\widehat{\Psi}({\bf{x}})}{f({\bf{x}})+o(1)}\right)\notag\\
&=&O_p\left(\frac{1}{N}\sum_{i=1}^{N}\big[\text{tr}(\m \Sigma_i)\big]^{\beta/2}+\sqrt{\frac{1}{N^2}\sum_{i=1}^{N}\frac{1}{\text{det}(\m \Sigma_i)^{1/2}}}\right),\label{rrate}
\end{eqnarray}
and this implies the convergence in probability (of the $j$th arbitrary component of $\widehat{\mathbf{v}}({\bf{x}})-\mathbf{v}({\bf{x}})$) by Assumption 3. Since the component-wise convergence in probability holds for all $j=1,\ldots, p$, w.l.o.g, the vectorial convergence holds for $\widehat{\mathbf{v}}({\bf{x}})-\mathbf{v}({\bf{x}})$ and so does the convergence with respect to the Euclidean norm, by standard results in probability theory e.g. \cite{Pollard}, as required. This completes the proof.\hfill$\square$\\

\subsubsection{Remark.}
We emphasize that the asymptotic theory in Theorem 1 and the accompanying proof remain valid when $(\boldsymbol{\mu}_1, \mathbf{v}_1), (\boldsymbol{\mu}_2, \mathbf{v}_2), \ldots, (\boldsymbol{\mu}_N, \mathbf{v}_N)$ are not independent and identically distributed (i.i.d.), but instead satisfy ergodic or mixing conditions. Under these alternative dependence structures, the main results still hold with minor modifications to the regularity conditions and asymptotic variance. We also note that, beyond consistency, the asymptotic normality of $\widehat{\mathbf{v}}(\cdot)$ can be established. Proofs of these extensions are available upon request.

\twocolumn

\subsection{Proof of Corollary 1}
When $\m \Sigma_i=\text{diag}(h_{i1}^2,h_{i2}^2,\ldots, h_{iq}^2)$, the trace of $\m \Sigma_i$ becomes the sum of $h_{ij}^2$ over $j=1,\ldots,q$, and the determinant becomes the product of $h_{ij}^2$ over $j$, hence the desired expression follows. Moreover, when $h_{ij}=O(h)$ for all $k,j$, the order for the leading term of the mean squared error becomes 
\begin{equation}
O\left(qh^{2\beta}+1/(Nh^q)\right)~=~O\left(N^{-\frac{2\beta}{2\beta+q}}\right),\notag
\end{equation}
upon assuming the optimal bandwidth $h= O(N^{-\frac{1}{2\beta+q}})$.
\hfill$\square$\\
\begin{figure*}[t]
\centering
    \includegraphics[width=0.9\textwidth]{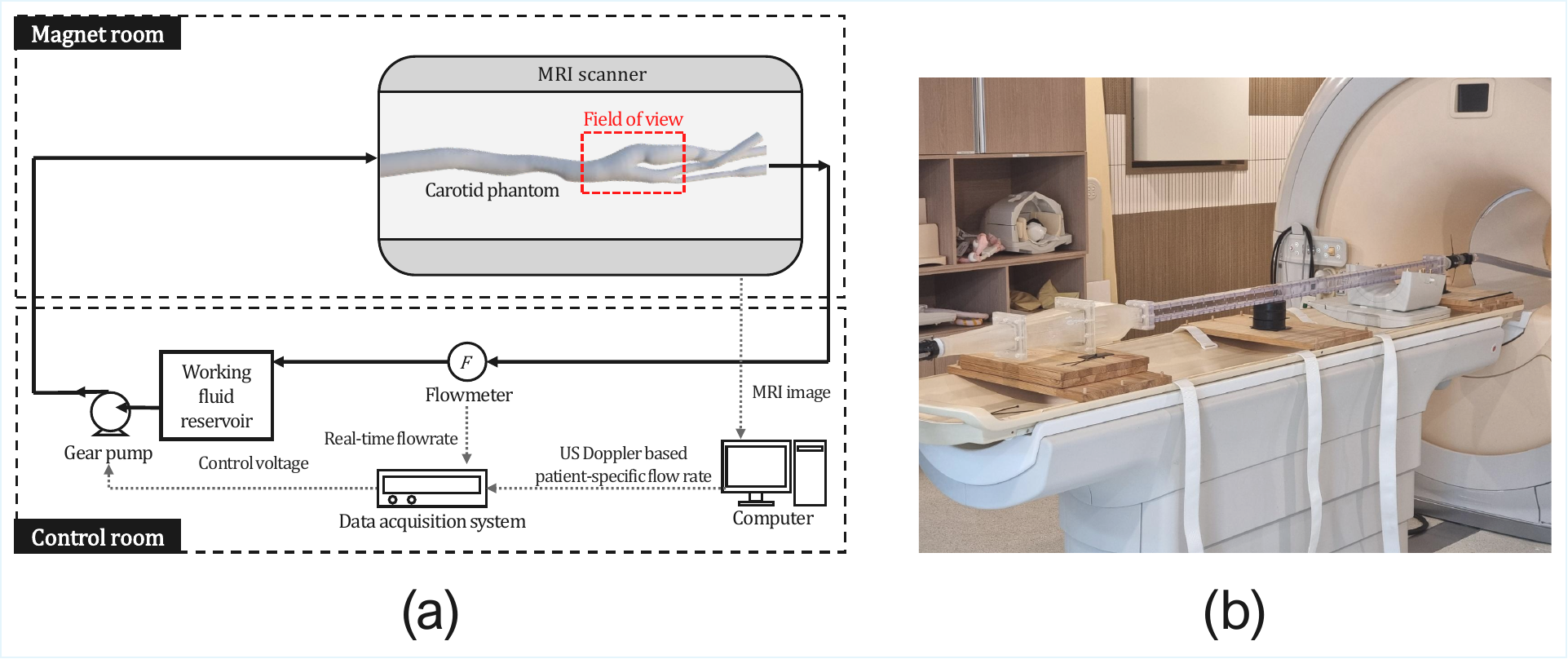}
    \caption{
         Visualization of the 4D flow MRI data-acquisition process. (a) shows the overall experimental setup, where the fluid was controlled using a computer and a centrifugal pump, and measurements were taken using a commercial MRI scanner. (b) is a photograph of the actual experiment-in-progress.
        }
    \figlabel{4d_flow_mri_acquisition}
\end{figure*}

\section{Dataset acquisition and pre-processing}
We provide additional details regarding the processes involved for acquiring both synthetic CFD and real 4D flow MRI datasets used for our experiments.

\subsection{Synthetic 2D CFD datasets}
\subsubsection{Lid-driven cavity.}
We generated synthetic data for incompressible fluid flow using CFD solvers. First, we utilized the commonly used lid-driven cavity flow~\cite{abdelmigid2017revisiting}, where the lid velocity was set to 1~m/s in a square domain of size 0.05~m, resulting in a Reynolds number of 500. No-slip boundary conditions were applied on the three stationary walls, while the top lid moved at a constant velocity. The pressure gradient normal to all boundaries was set to zero. The simulation was carried out using the open-source software OpenFOAM on a structured grid with $400 \times 400$ cells.

\subsubsection{Y-shaped/L-shaped channels.}
Additionally, fluid flow data in L-shaped and Y-shaped channels were obtained to validate our method on flow geometries resembling blood flow in vessels. Both channels have an inlet height of 0.02~m, where a parabolic velocity profile was imposed such that the average inlet velocity is 1~m/s. With the kinematic viscosity of the fluid defined as $1 \times 10^{-4}~\mathrm{m}^2/\mathrm{s}$, this corresponds to a fully developed laminar inflow with a Reynolds number of 200 at the inlet.

At the outlet, a Dirichlet boundary condition was applied for pressure, corresponding to a reference pressure, and a Neumann condition was imposed for velocity. No-slip (Dirichlet) boundary conditions with zero velocity were applied along the walls, and a zero-gradient condition was used for pressure in the direction normal to the walls.

The simulations were performed under steady-state conditions, assuming time-invariant flow. Numerical discretization schemes with at least second-order accuracy were employed. The computational meshes were generated using the automatic 2D polyhedral meshing algorithm in StarCCM+ version~16.1. The total number of cells was approximately 76{,}000 for the L-shaped channel and 80{,}000 for the Y-shaped channel. The numerical solutions converged to steady-state with a relative residual below $10^{-14}$.

For the Y-shaped and L-shaped datasets, this raw data was re-gridded onto a uniform grid to create structured grid data, which was used as the high-resolution ground truth.
This resulted in 55,274 points for the Y-shaped data and 57,105 points for the L-shaped data. The number of data points used in the experiments for each resolution level is in Table~\ref{tab:fluid_points}.


\begin{table}[t]
\centering
\small
\setlength{\tabcolsep}{1mm}

    \begin{tabular}{l|c|c|c|c}
    \hline\hline
    Dataset  & \multicolumn{3}{c|}{2D CFD data} & \makecell{4D flow\\MRI data} \\ 
    \hline\hline
    Shape 
        & \makecell{Lid-driven} 
        & \makecell{Y-shape} 
        & \makecell{L-shape}
        & \makecell{Carotid}\\ \hline
    Original & 160,000 & 55{,}274 & 57{,}105 & 69{,}008 \\ \hline
    Spat. aver.~($\times4$) & 40{,}000 & 13{,}854 & 14{,}343 & -\\ \hline
    Spat. aver.~($\times8$) & - & - & - & 9{,}726\\ \hline
    Spat. aver.~($\times16$) & 10{,}000 & 3{,}458 & 3{,}552 & -\\ \hline
    Spat aver.~($\times64$) & 2{,}500  & 868     & 881     & 1{,}505\\ \hline\hline
    \end{tabular}
    \caption{Number of data points for the high-resolution (original) and low-resolution (spat. aver.) versions of each dataset used in our experiments. Note $\times 16$ and $\times 8$ are the settings reported for synthetic 2D CFD data and 4D flow MRI datasets respectively in the main paper.}
    \label{tab:fluid_points}
\end{table}

\subsection{Real 4D flow MRI dataset}
We utilized 4D flow MRI data measured in a carotid artery phantom that replicates the vascular structure of an actual patient, as measured in (Ko et al. 2019). The 4D flow MRI measurements were performed using a 4.7 Tesla MRI system (BioSpec 47/40, Bruker, Germany) equipped with a standard single-channel birdcage RF coil. Data acquisition was carried out using FLOWMAP, a commercially available phase-contrast MRI sequence integrated into the system console. The imaging was conducted with an isotropic spatial resolution of 0.35 mm and a temporal resolution of 25 ms. For each cardiac cycle, 34 time frames were obtained; among these, five time frames centered around the peak systolic phase were selected for analysis in this study. The Reynolds number for the carotid data was calculated based on the hydraulic diameter and the mean velocity at the corresponding cross-section of the common carotid artery (CCA).
An illustration of the overall acquisition process is provided in  Fig.~\figref{4d_flow_mri_acquisition}.

\subsection{Spatial averaging method}

To create physically realistic low-resolution data for our super-resolution tasks, we used a spatial averaging method. This approach is designed to simulate the partial volume effect inherent in MRI acquisitions, where the signal in a single low-resolution voxel represents the spatial average of the underlying sub-voxel properties. This is a more physically plausible downsampling model than simple sub-sampling. The specific methods are as follows.

\subsubsection{2D synthetic CFD datasets.}
To create low-resolution data for the 2D super-resolution tasks (e.g., with a $\times$4 factor), we applied 2D spatial averaging. 
Specifically, the velocities of four high-resolution points in a $2\times2$ rectangular region were averaged, and a new low-resolution point was created at the center of this region with the averaged velocity value.
To generate data with a higher averaging factor, such as $\times16$, this spatial averaging process was applied recursively. First, a $\times4$ averaged dataset was created from the high-resolution data. Then, another $\times4$ spatial averaging was applied to these newly generated low-resolution data to create the final $\times16$ averaged data.

\subsubsection{4D flow MRI dataset.}
For the 4D super-resolution experiments (e.g., with a $\times8$ factor), low-resolution data was generated by averaging the velocities of eight high-resolution points forming a $2\times2\times2$ rectangular cuboid. A new low-resolution point was then placed at the center of the cuboid with the averaged velocity.
For the $\times64$ task, this process was repeated by applying another $\times8$ spatial averaging to the already averaged $\times8$ low-resolution data.



\begin{algorithm*}[t]
    \caption{Proposed merging algorithm for axes-aligned Gaussians}
    \begin{algorithmic}
        \State \textbf{Input}: Initial set of Gaussians $\mathcal G := \{\mathcal{G}_i\} = \{\boldsymbol{\mu}_i, \mathbf{h}_i, \mathbf{v}_i\}_{i=1}^{N}$, training (spatiotemporal) data points $X = \{\mathbf{x}_k\}_{k=1}^{K}$
        \State \textbf{Output}: Merged set of Gaussians $\mathcal{G}_\text{merged}$
        \newline
    
        \For{$i = 1, \cdots, N$}
            \State $\m \Sigma_i^{-1} \leftarrow \mathtt{diag}(h_{i1}^{-2}, h_{i2}^{-2}, \cdots h_{iq}^{-2})$
            \For{$k = 1, \cdots, K$}
                \State $z_{ik} \leftarrow \exp(-\frac{1}{2}(\v {x}_k-\boldsymbol{\mu}_i)^T \m {\Sigma}_i^{-1} (\v {x}_k-\boldsymbol{\mu}_i))$                
                \Comment Influence of $\mathcal G_i$ on data point $k$
            \EndFor
            \State $\v z_i \leftarrow [z_{i1}, z_{i2}, \cdots z_{iK}]\tr$. 
            \Comment Form an \emph{influence vector} for $\mathcal G_i$.
        \EndFor
        \newline

        \For{$i = 1, \cdots, N$}
            \For{$j = i+1, \cdots, N$}
                \State $\mathtt{sim}(i,j) \leftarrow \v z_i \cdot \v z_j / \| \v z_i \| \|\v z_j \|$.        
                \Comment Calculate similarity of influences between Gaussians $\mathcal G_i$ and $\mathcal G_j$.
            \EndFor
        \EndFor
        \newline               
        
        \State $E \gets \{(i, j) \mid S_{ij} > \tau_{\text{merge}} ~\cap~ i < j\}$ \Comment{Create edge set for similarities above threshold ($\tau_{\text{merge}}=0.9$)}
        \State $C \gets \Call{FindConnectedComponents}{E}$ \Comment{Find clusters (connected components) from the edge set}
        \State $\mathcal{G}_{\text{remove}} \gets \emptyset, \mathcal{G}_{\text{add}} \gets \emptyset$
        \ForAll{cluster $c$ in $C$}
            \State $\mathbf{\mu}_{\text{new}} \gets \frac{1}{|c|} \sum_{i \in c} \mathbf{\mu}_i$ \Comment{Average the parameters within the cluster}
            \State $\mathbf{h}_{\text{new}} \gets \frac{1}{|c|} \sum_{i \in c} \mathbf{h}_i$
            \State $\mathbf{v}_{\text{new}} \gets \Call{PredictVelocities}{\mathcal{G}, \boldsymbol{\mu}_{\text{new}}}$ \Comment{Re-predict physical properties at the new position}
            \State $\mathcal{G}_{\text{add}} \gets \mathcal{G}_{\text{add}} \cup \{(\mathbf{\mu}_{\text{new}}, \mathbf{h}_{\text{new}}, \mathbf{v}_{\text{new}})\}$
            \State $\mathcal{G}_{\text{remove}} \gets \mathcal{G}_{\text{remove}} \cup \{\mathcal{G}_i \mid i \in c\}$
        \EndFor
        \newline
        
        \State $\mathcal{G}_{\text{merged}} \gets (\mathcal{G} \setminus \mathcal{G}_{\text{remove}}) \cup \mathcal{G}_{\text{add}}$
        \newline
        
        \State \textbf{return} $\mathcal{G}_{\text{merged}}$
    \end{algorithmic}
    \label{alg:merging_detailed}
\end{algorithm*}

\section{Additional implementation details}

\subsection{Physics-informed loss function}

The $L_{PDE}$ term in our framework is derived from the incompressible Navier-Stokes equation in Eq.~\eqref{navier_stokes}, which describe the motion of a viscous fluid, and the continuity equation, which describes mass conservation in Eq.~\eqref{continuity}.  
To enhance numerical stability, we utilize the \emph{dimensionless} form of these equations.  
The standard incompressible Navier-Stokes equations are:
\begin{align}
\frac{\partial\mathbf{u}}{\partial t}+(\mathbf{u}\cdot\nabla)\mathbf{u}
      &= -\frac{1}{\rho}\nabla p + \nu\nabla^{2}\mathbf{u} + \mathbf{g},
      \label{eq:navier_stokes_supple} 
      \\
\nabla\cdot\mathbf{u} &= 0.                                   \label{eq:continuity_supple}
\end{align}
where $\mathbf{u}\in\real^3$ is the velocity vector, $p$ is the pressure, $\rho$ is the fluid density, $\nu$ is the kinematic viscosity, and $\mathbf{g}\in\real^3$ is the gravity vector.
To non-dimensionalize these equations, we introduce characteristic scales: a characteristic length $L$, and a characteristic velocity $U$.  
We then define the following dimensionless variables:
\begin{align}
    \mathbf{x^{*}} = \frac{[x, y, z]\tr}{L}, \quad
    \nabla^{*} = L\nabla, \quad
    \mathbf{u^{*}} = \frac{\mathbf{u}}{U}, \nonumber \\
    t^{*} = \frac{t}{L/U}, \quad
    p^{*} = \frac{p}{\rho U^{2}}.
\end{align}

In this framework, we adopt a commonly used strategy in computational fluid dynamics (CFD), where the gravitational term $\mathbf{g}$ is absorbed into the pressure field by defining a \emph{modified pressure}:
\begin{align}
    \tilde{p} = p - \v g \cdot (L \v x^*).
\end{align}
This effectively removes the explicit gravity term from the momentum equation, as the hydrostatic contribution is now embedded in the pressure gradient. \begin{align}
    \tilde{p}^{*} = \frac{\tilde{p}}{\rho U^{2}} = \frac{p - \rho\, \mathbf{g} \cdot (L \mathbf{x}^{*})}{\rho U^{2}}.
\end{align}

Substituting these into the momentum equation and simplifying, we obtain the dimensionless Navier-Stokes equation:
\begin{align}
\frac{\partial\mathbf{u^{*}}}{\partial t^{*}}+(\mathbf{u^{*}}\cdot\nabla^{*})\mathbf{u^{*}}
      + \nabla^{*} \tilde{p}^{*} - \frac{1}{Re}\nabla^{2}\mathbf{u^{*}} = 0.
      \eqlabel{dimensionless_ns}
\end{align}
where $Re$ is the Reynolds number $(Re = UL/\nu)$.  
The continuity equation remains unchanged in its dimensionless form:
\begin{align}
    \nabla^{*}\cdot\mathbf{u^{*}} = 0.
    \eqlabel{dimensionless_continuity}
\end{align}
Our $L_{PDE}$ is simply defined as the square of the LHS of \eqref{dimensionless_ns} summed with the square of the LHS of \eqref{dimensionless_continuity}.

\subsubsection{Remark.}
Note the gravity term is omitted in Eq.~\eqref{dimensionless_ns} because we follow a standard practice in CFD, where the velocity field is computed based on the modified pressure, which already incorporates gravitational effects. Once the velocity and dynamic pressure field are obtained, the absolute pressure can be recovered by simply adding back the hydrostatic pressure term $\rho \mathbf{g} \cdot (L \v {x}^*)$ if needed.  
This approach not only simplifies the numerical formulation, but also aligns with the principle of \emph{dynamic similarity}, wherein gravity does not explicitly appear in the dimensionless governing equations for incompressible flows.

\subsection{Gaussian merging algorithm}
We detail our \emph{Gaussian merging} algorithm in Algorithm~\ref{alg:merging_detailed}.

\subsection{XPINN configuration}
\begin{figure}[t]
\centering
    \includegraphics[width=1.0\columnwidth]{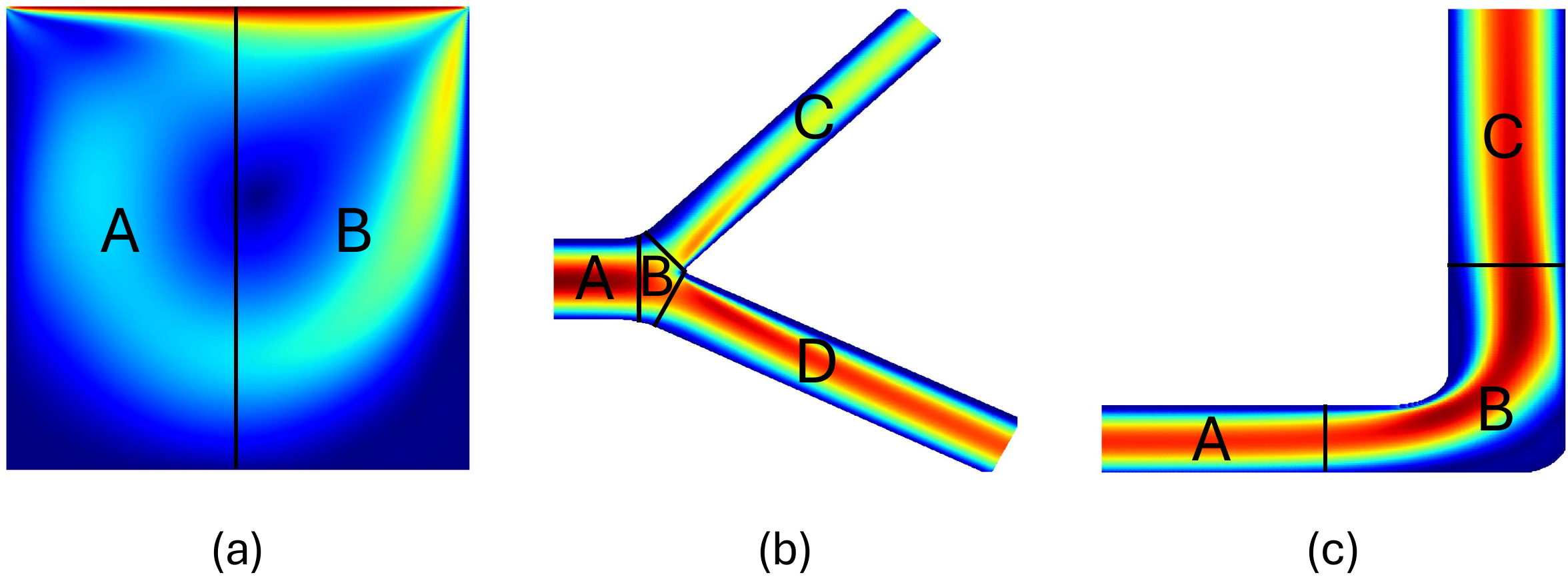}
    \caption{
         Visualization of the domain decomposition for the XPINN model on the three synthetic CFD datasets. 
        (a) The lid-driven cavity domain is divided into 2 sub-domains (A, B). 
        (b) The Y-shaped channel domain is divided into 4 sub-domains (A, B, C, D). 
        (c) The L-shaped channel domain is divided into 3 sub-domains (A, B, C). 
        Each sub-domain is handled by a separate neural network within the XPINN framework.
        }
    \label{fig:XPINN_boundary}
\end{figure}

For the XPINN framework, the number of sub-networks was configured to align with the geometric complexity of each problem. Specifically, we set the number of sub-networks to 2 for the lid-driven cavity problem, 3 for the L-shaped domain, and 4 for the Y-shaped problem. To segment the spatial regions for these individual sub-networks, we have used simple decomposition boundaries shown in Fig.~\ref{fig:XPINN_boundary} throughout our experiments.

\subsection{Hardware and software settings}
All models were trained on a single workstation featuring an Intel Xeon Gold 6342 2.8GHz CPU, 512 GB RAM, and a single NVIDIA RTX A6000 GPU. The experiments utilized Python 3.9.23, Pytorch 2.7.1+cu128, and the Ubuntu 20.04 operating system. 

\subsection{List of hyperparameters}
We have included a list of hyperparameters in PINGS-X and their values in Table~\ref{tbl:hyperparams_pingsx}.
We have also provided a list of hyperparameters used for other baseline and state-of-the-art models in Table~\ref{tbl:hyperparams_others}.

\begin{table}[t]
\centering\small
\begin{tabular}{l | c | c }
    \hline\hline
    Parameter  & 2D CFD & 4D flow MRI \\ \hline\hline 
    Optimizer  & {Adam} & {Adam}   \\ 
    {Learning rate}  & {$10^{-2}$} & {$10^{-2}$}   \\  
    {Epochs}  & {10,000} & {1,000}   \\  
    {Batch size}  & {Full-batch} & {10,000}   \\  
    {Initial Gaussian}  & {400} & {$1/10$ of LR points}   \\  
    {Densification step}  & {100 epochs} & {100 epochs}   \\  
    {Densification threshold}  & {2.0} & {2.0}   \\  
    {split/clone threshold}  & {$2.0\times10^-4$} & {$2.0\times10^-4$}   \\  
    {Merging step }  & {100 epochs} & {100 epochs}   \\  
    {Merging threshold}  & {0.9} & {0.9}   \\  
    {PDE loss weight $(\lambda)$}  & {1.0} & {1.0}   \\  
    {Z-threshold}  & {$10^{-4}$} & {$10^{-4}$} \\
    \hline\hline
\end{tabular}
\caption{
    A list of hyperparameters used in PINGS-X.
    The z-threshold is used to ignore Gaussians with an influence score below this value at each data point $\v x_k$.
    This is carried out to disregard those with a negligible contribution to the final prediction solely for computational efficiency.
}
\label{tbl:hyperparams_pingsx}
\end{table}

\begin{table}[t]
\centering\scriptsize
\begin{tabular}{l | l | c | c }
    \hline\hline
    Model & Parameter  & 2D CFD & 4D flow MRI \\ \hline\hline 
    \multirow{7}{*}{PINN} & {Hidden layers} & {5} & {8} \\  
    {} & {Neurons} & {128} & {256} \\  
    {} & {Activation} & {Tanh} & {Tanh} \\  
    {} & {Learning rate} & {$10^{-4}$} & {$10^{-4}$} \\  
    {} & {Optimizer} & {Adam} & {Adam} \\  
    {} & {Epochs} & {100,000} & {100,000} \\  
    {} & {Batch Size} & {Full-batch} & {10,000} \\
    \hline
    \multirow{8}{*}{Siren} & {Hidden layers} & {5} & {8} \\  
    {} & {Neurons} & {128} & {256} \\  
    {} & {Activation} & {sin} & {sin} \\  
    {} & {Learning rate} & {$5.0\times 10^{-6}$} & {$5.0 \times 10^{-6}$} \\  
    {} & {$\omega_0$} & {10} & {20} \\  
    {} & {Optimizer} & {Adam} & {Adam} \\  
    {} & {Epochs} & {100,000} & {100,000} \\  
    {} & {Batch Size} & {Full-batch} & {10,000} \\
    \hline
    \multirow{8}{*}{XPINN (Tanh)} & {Hidden layers} & {5} & {--} \\  
    {} & {Neurons} & {128} & {--} \\  
    {} & {Activation} & {Tanh} & {--} \\  
    {} & {Learning rate} & {$10^{-4}$} & {--} \\  
    {} & {Sub-networks} & \makecell{2 (Lid-driven),\\ 3 (L-shaped), \\4 (Y-shaped)} & {--} \\  
    {} & {Optimizer} & {Adam} & {--} \\  
    {} & {Epochs} & {100,000} & {--} \\  
    {} & {Batch Size} & {Full-batch} & {--} \\
    \hline
    \multirow{9}{*}{XPINN (sin)} & {Hidden layers} & {5} & {--} \\  
    {} & {Neurons} & {128} & {--} \\  
    {} & {Activation} & {sin} & {--} \\  
    {} & {Learning rate} & {$5\times 10^{-6}$} & {--} \\  
    {} & {$\omega_0$} & {10} & {--} \\  
    {} & {Sub-networks} & \makecell{2 (Lid-driven), \\3 (L-shaped), \\4 (Y-shaped)} & {--} \\  
    {} & {Optimizer} & {Adam} & {--} \\  
    {} & {Epochs} & {100,000} & {--} \\  
    {} & {Batch Size} & {Full-batch} & {--} \\
    \hline
    {PIG} & {Configuration} & {\makecell{Official\\implementation}} & {--} \\
    \hline\hline
\end{tabular}
\caption{
    A list of hyperparameters used in other models.
}
\label{tbl:hyperparams_others}
\end{table}

\section{Limitations and future work} 
PINGS-X introduces several hyperparameters for initialization and adaptive density control. 
While our sensitivity analysis indicates the model is robust to these settings, automating their selection is a key area for future improvement. 
Furthermore, the current general framework remains generic, and could be fine-tuned for specific anatomies to potentially enhance its accuracy for targeted clinical applications.

\section{Additional experimental results}


\begin{table}[t]
    \centering
    \scriptsize
    \setlength{\tabcolsep}{1mm}
    \centering
    \begin{tabular}{c|c
        |rr|rr|rr
    }
        \hline\hline
         \multirow{3}{*}{\makecell{Densification  \\Thres.}}& \multirow{3}{*}{\makecell{Merging  \\Thres.}}& \multicolumn{2}{c|}{Lid driven}
            & \multicolumn{2}{c|}{Y-shape}
            & \multicolumn{2}{c}{L-shape}
            \\
         \cline{3-8}
        &
        & \makecell{$t$ $\downarrow$~\\(min)} & \makecell{Rel. $\downarrow$\\ err.~(\%)}
        & \makecell{$t$ $\downarrow$~\\(min)} & \makecell{Rel. $\downarrow$\\ err.~(\%)}
        & \makecell{$t$ $\downarrow$~\\(min)} & \makecell{Rel. $\downarrow$\\ err.~(\%)}
        \\
        \hline\hline
  1.5
& 0.85
& 8.2& 1.46& 3.9& 2.90& 3.7& 1.22
\\
  2.0
& 
0.85
& 6.3& 1.91& 3.8& 2.79& 3.8& 1.34
\\
  2.5
& 0.85
& 5.7& 2.09& 4.0& 2.80& 4.3& 1.29
\\
  3.0
& 
0.85
& 5.0& 3.00& 3.9& 2.93& 3.4& 1.38
\\
  1.5
& 0.90
& 43.1& 1.12& 5.3& 2.46& 8.8& 1.16
\\
         \textbf{2.0}
& 
\textbf{0.90}
& \textbf{22.2}& \textbf{1.13}& \textbf{5.2}& \textbf{2.62}& \textbf{6.7}& \textbf{1.15}\\
         2.5
& 0.90
& 23.6& 1.11& 5.5& 2.46& 6.0& 1.11\\
         3.0
& 0.90
& 17.7& 1.12& 4.7& 2.49& 4.7& 1.17\\
         1.5
& 0.95 
& 135.6& 1.06& 20.6& 2.25& 60.4& $^*$1.14\\
         2.0
& 0.95 
& 92.8& 1.07& 16.7& 2.34& 29.7& 1.16\\
  2.5
& 0.95 
& 76.9& 1.05& 17.6& 2.27& 13.6& 1.13\\
         3.0& 0.95 
& 59.6& 1.04& 13.9& 2.30& 11.3& 1.15\\
        \hline\hline
    \end{tabular}
    \caption{
        Quantitative comparison under different adaptive threshold settings on the 2D synthetic CFD datasets.
        $^*$ Denotes experiments that were terminated at 5200 epochs due to OOM.
    }
    \label{tbl:split_merge_ablation}
\end{table}

\subsection{Sensitivity analysis for hyperparameters}
In Table~\ref{tbl:split_merge_ablation}, we provide performance of PINGS-X across a range of different hyperparameter values used for (i) Gaussian densifications and (ii) merging.
Note we perform densification for Gaussians contributing to highest errors on the training data points.
For this purpose, we calculate for each Gaussian the weighted sum of losses across the training points. More specifically, the error at each data point $k$ is weighted by the normalized influence of $\mathcal G_k$, i.e. $\varepsilon_i := \sum_{k=1}^K L_k z_{ik} / \| \v z_i \|_2$, where $L_k$ is the total loss at the spatiotemporal point $\v x_k$, $z_ik$ is the unnormalized influence of Gaussian $\mathcal G_i$ and $\v z_i$ is the influence vector of $\mathcal G_i$.
After accumulating these (weighted) errors, $\{ \varepsilon_i \}$, across all Gaussians, any Gaussians exceeding the median of this error multiplied by the \emph{densification threshold} is densified (split or cloned).

On the other hand, the merging threshold defines the value of minimum cosine similarity required for merging between a pair of Gaussians.
This shows the performance of PINGS-X is relatively stable across a wide range of densification and merge threshold settings. 

\subsection{Effect of different input resolutions}

In Table~\ref{table:supp_spatial_1}, we provide experimental results for different resolutions of LR data using our synthetic 2D CFD datasets.
Overall, PINGS-X achieves state-of-the-art performance in terms of reduced training time ($t$) and reduced relative $L_2$ error (Rel.~err.) across different resolutions.

\subsection{Experimental results on Burger's equation}
For additional analysis, we have also compared the super-resolution performance on the well-known (1+1)-D Burgers' equation.
For this task, the low-resolution data is processed by simplying sampling the high-resolution data (no spatial averaging involved).
We obtained the public data from \texttt{\url{https://github.com/maziarraissi/PINNs/tree/master/appendix/Data}}.
Table~\ref{table:burgers} again demonstrates PINGS-X reducing both the training time and the relative $L_2$ error compared to other methods.
\begin{table}[t]
    \centering
    \small
    \begin{subtable}{1.0\linewidth}
        \centering
        \setlength{\tabcolsep}{1mm}
        \begin{tabular}{c|c|c}
            \hline\hline
            \multirow{2}{*}{Method} & \multicolumn{2}{c}{Burgers} \\
            \cline{2-3}
             & $t$~(min) $\downarrow$ & Rel. err.~(\%) $\downarrow$ \\
            \hline\hline
            Nada.-Watson    & $<$1 & 3.95 \\ \hline
            PINN            & \underline{19.1} & 0.91 \\
            Siren           & 22.1 & 0.91 \\
            XPINN~(tanh)    & 63.8 & 1.09 \\
            XPINN~(sin)     & 74.5 & \underline{0.48} \\
            PIG             & {252.3} & {1.14} \\
            {PINGS-X (ours)} & \textbf{12.6} & \textbf{0.45} \\
            \hline\hline
        \end{tabular}
    \end{subtable}
    \caption{
        Comparison of training time and relative $L_2$ error (\%) on the dataset generated from the (1+1)-D Burgers' equation.
        For PINGS-X, we use the same set of hyperparameters as for the synthetic 2D CFD datasets (i.e. initial \# Gaussians: 400, split threshold: 2.0, merge threshold: 0.9).
    }
    \label{table:burgers}
\end{table}

\begin{table}[t]
    \centering
    \small
     \setlength{\tabcolsep}{1mm}
    \begin{subtable}{1.0\linewidth}
        \centering
        \begin{tabular}{
            c
            | r r
            | r r
            | r r
        }
            \hline\hline
            \multirow{3}{*}{Method}
                & \multicolumn{2}{c|}{Lid driven}
                & \multicolumn{2}{c|}{Y-shape}
                & \multicolumn{2}{c}{L-shape} \\
            \cline{2-7}
            & \makecell{$t$~ $\downarrow$\\(min)} & \makecell{Rel. $\downarrow$\\ err.~(\%)}
            & \makecell{$t$~ $\downarrow$\\(min)} & \makecell{Rel. $\downarrow$\\ err.~(\%)}
            & \makecell{$t$~ $\downarrow$\\(min)} & \makecell{Rel. $\downarrow$\\ err.~(\%)} \\
            \hline\hline
            \makecell{Nada.-Watson} & 1.5& \underline{1.24}& $<$1 & \underline{2.33} & $<$1 & \underline{1.68} \\ \hline
            PINN            & \underline{167.7}& 12.10 & \underline{67.6}& 11.09 & \underline{68.9} & 12.09 \\
            Siren           & 194.3 & 3.21 & 78.7 & 3.89 & 87.2 & 2.48 \\
            XPINN (tanh)    & 194.6& 8.60 & 168.7& 8.09 & 130.2 & 5.79 \\
            XPINN (sin)     & 230.9& 2.66 & 206.5 & 3.51 & 152.8& 2.14 \\
            PIG             & {843.0} & {3.24} & {840.7} & {9.51} & {841.6} & {2.15} \\
            {PINGS-X (ours)} & \textbf{70.7}& \textbf{0.63}& \textbf{10.1}& \textbf{1.52}& \textbf{14.3}& \textbf{1.03}\\
            \hline\hline
        \end{tabular}
        \caption{Results with spatial averaging factor $\times$ 4.}
    \end{subtable}


    \begin{subtable}{1.0\linewidth}
        \centering
        \begin{tabular}{
            c
            | r r
            | r r
            | r r
        }
            \hline\hline
            \multirow{3}{*}{Method}
                & \multicolumn{2}{c|}{Lid driven}
                & \multicolumn{2}{c|}{Y-shape}
                & \multicolumn{2}{c}{L-shape} \\
            \cline{2-7}
            & \makecell{$t$~ $\downarrow$\\(min)} & \makecell{Rel. $\downarrow$\\ err.~(\%)}
            & \makecell{$t$~ $\downarrow$\\(min)} & \makecell{Rel. $\downarrow$\\ err.~(\%)}
            & \makecell{$t$~ $\downarrow$\\(min)} & \makecell{Rel. $\downarrow$\\ err.~(\%)} \\
            \hline\hline
            \makecell{Nada.-Watson} & $<$1 & 9.51 & $<$1 & 10.13 & $<$1 & 8.53 \\ \hline
            PINN            & \underline{62.0} & 11.87 & \underline{59.8}& 12.54 & \underline{61.7}& 11.83 \\
            Siren           & 69.5& 3.06 & 69.8 & 5.15 & 69.5 & 2.69 \\
            XPINN (tanh)    & 120.3 & 8.16 & 209.9 & 10.26 & 172.1& 5.37 \\
            XPINN (sin)     & 138.6 & \underline{2.62} & 267.6 & \underline{4.73} & 213.0& \underline{2.37} \\
            PIG             & {215.8} & {2.80} & {88.8} & {6.44} & {89.1} & \textbf{2.03} \\
            {PINGS-X (ours)} & \textbf{1.5}& \textbf{2.55}& \textbf{1.2}& \textbf{4.21}& \textbf{1.3}& {2.41}\\
            \hline\hline
        \end{tabular}
        \caption{Results with spatial averaging factor $\times$ 64.}
    \end{subtable}



    \caption{
         Quantitative comparison on synthetic 2D CFD datasets to analyze the effect of input resolution.
    }
    \label{table:supp_spatial_1}
\end{table}

\section{Additional visualizations}
\subsection{Number of Gaussians per epoch}
\begin{figure}[t]
  \centering
  \begin{subfigure}[b]{1.0\columnwidth}

    \includegraphics[width=\textwidth]{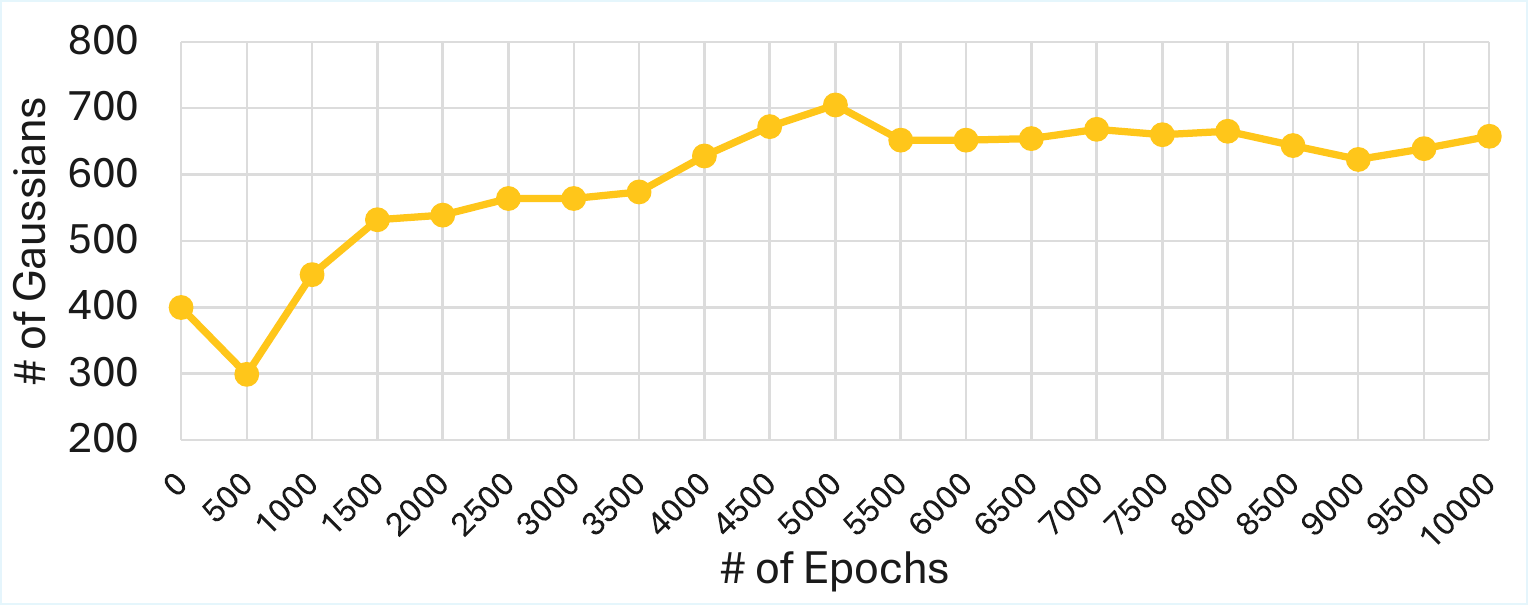}
    \caption{Lid-driven}
    \label{fig:gaussian_number_lid}
  \end{subfigure}
  
  \begin{subfigure}[b]{1.0\columnwidth}
    \centering
    \includegraphics[width=\textwidth]{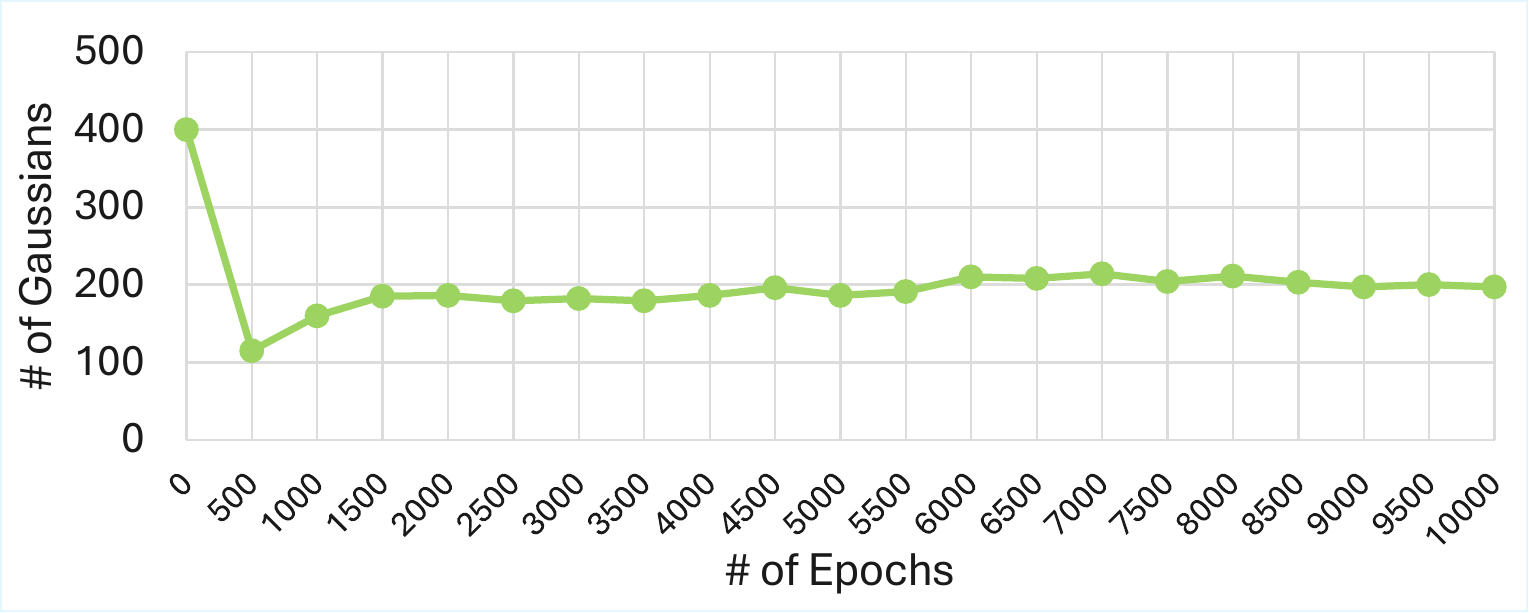}
    \caption{Y-shape}
    \label{fig:gaussian_number_y}
  \end{subfigure}
  
  \begin{subfigure}[b]{1.0\columnwidth}
    \centering
    \includegraphics[width=\textwidth]{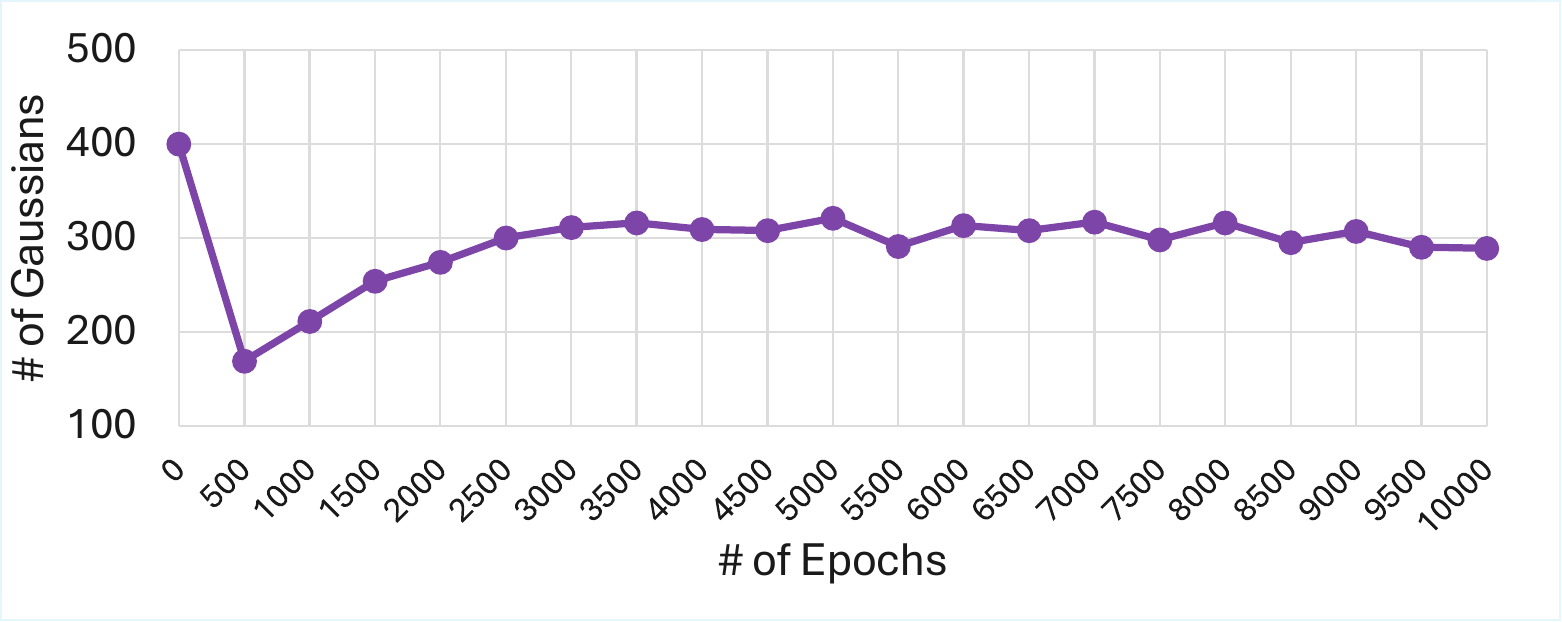}
    \caption{L-shape}
    \label{fig:gaussian_number_l}
  \end{subfigure}
    \caption{Evolution of the number of Gaussians during training for synthetic 2D CFD dataset. The plots illustrate the total count of active Gaussians for the (a) Lid-driven, (b) Y-shape, and (c) L-shape cases, sampled at 100-epoch intervals. It demonstrates the stability of our adaptive density method, where the number of Gaussians stabilizes within a consistent range rather than growing uncontrollably. This equilibrium is maintained by the dynamic balance between our densification operations, which add Gaussians in high-error regions, and the merge operation, which prunes redundant components.}
     \label{fig:gaussian_number_2d}
\end{figure}

\begin{figure}[t]
  \centering
  \begin{subfigure}[b]{1.0\columnwidth}
    \includegraphics[width=\textwidth]{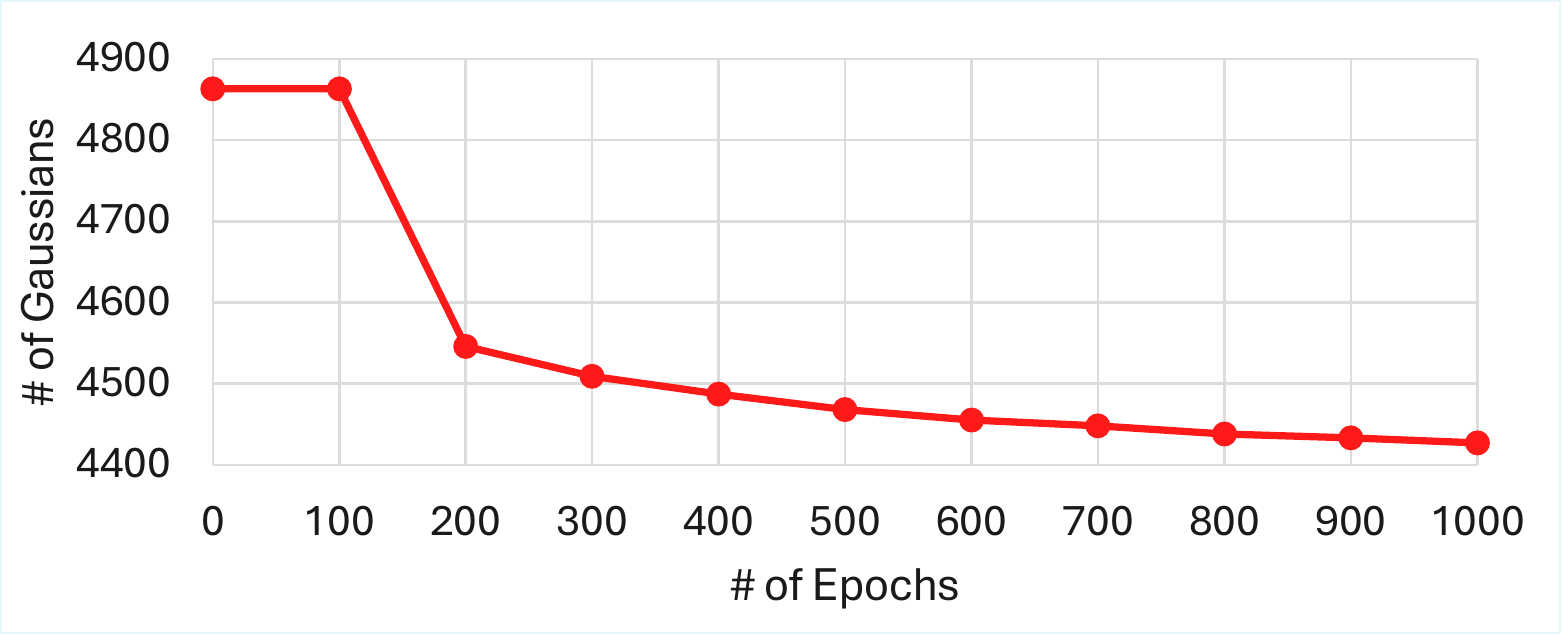}
    \caption{Spatial averaging $\times$ 8}
    \label{fig:gaussian_number_carotid_X8}
  \end{subfigure}
  
  \begin{subfigure}[b]{1.0\columnwidth}
    \centering
    \includegraphics[width=\textwidth]{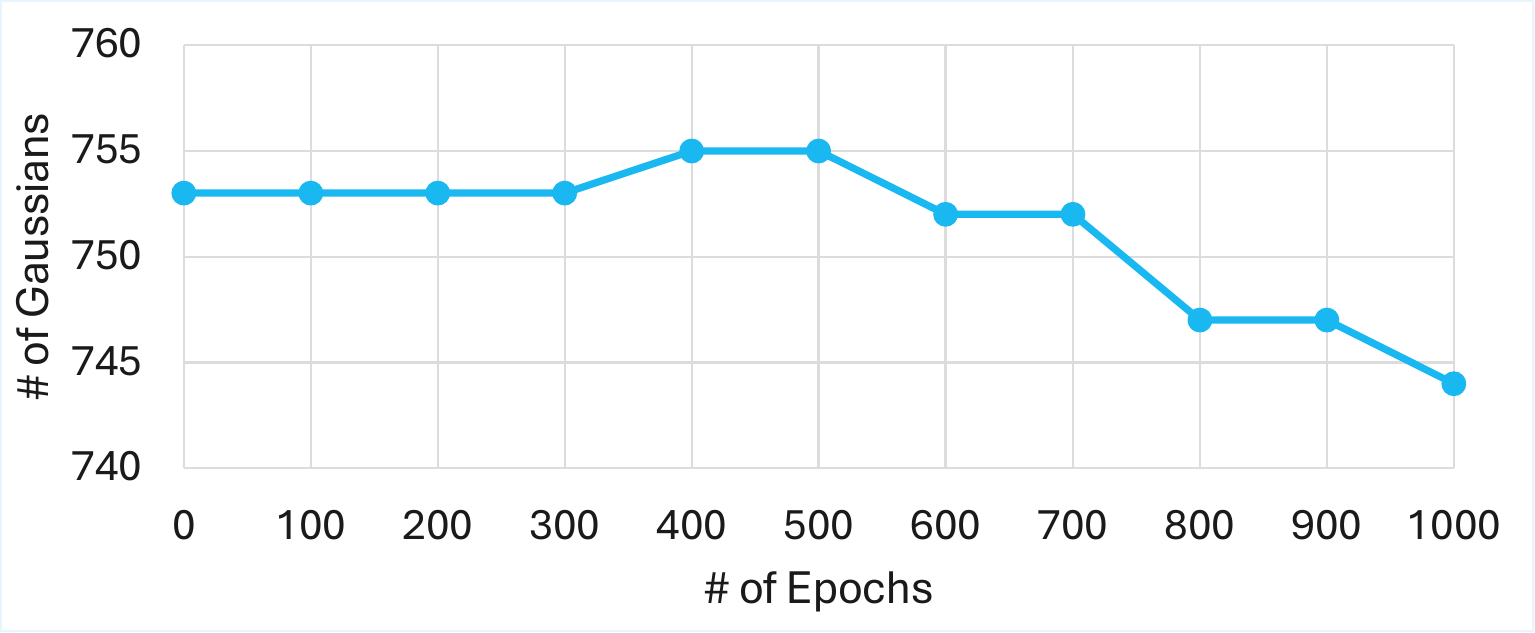}
    \caption{Spatial averaging $\times$ 64}
    \label{fig:gaussian_number_carotid_X64}
  \end{subfigure}
    \caption{Evolution of the number of Gaussians during training for 4D flow MRI dataset (Spatial averaging $\times$ 8, $\times$ 64). The plot illustrates the total count of Gaussians, sampled at 100-epoch intervals. It demonstrates the stability of our adaptive density method, where the number of Gaussians stabilizes within a consistent range rather than growing uncontrollably. This equilibrium is maintained by the dynamic balance between our densification operations, which add Gaussians in high-error regions, and the merge operation, which prunes redundant components.}
     \label{fig:gaussian_number_carotid}
\end{figure}

To analyze the behavior of our adaptive density control, we tracked the total number of Gaussians during training for all datasets. As shown for both the synthetic CFD (Fig.~\ref{fig:gaussian_number_2d}) and real 4D flow MRI datasets (Fig.~\ref{fig:gaussian_number_carotid}), the number of Gaussians does not grow monotonically. Instead, after an initial adjustment phase, the count stabilizes and fluctuates within a consistent range. This indicates a dynamic equilibrium, maintained by the balance between the densification operations that add Gaussians in high-error regions and our merge operation that removes redundant components. This controlled behavior highlights the effectiveness of our adaptive framework, which remains stable even for the highly sparse ($\times$64) real-world data.

\subsection{Illustration of adaptive Gaussian dynamics}
\begin{figure*}[t]
\centering
    \includegraphics[width=1.0\textwidth]{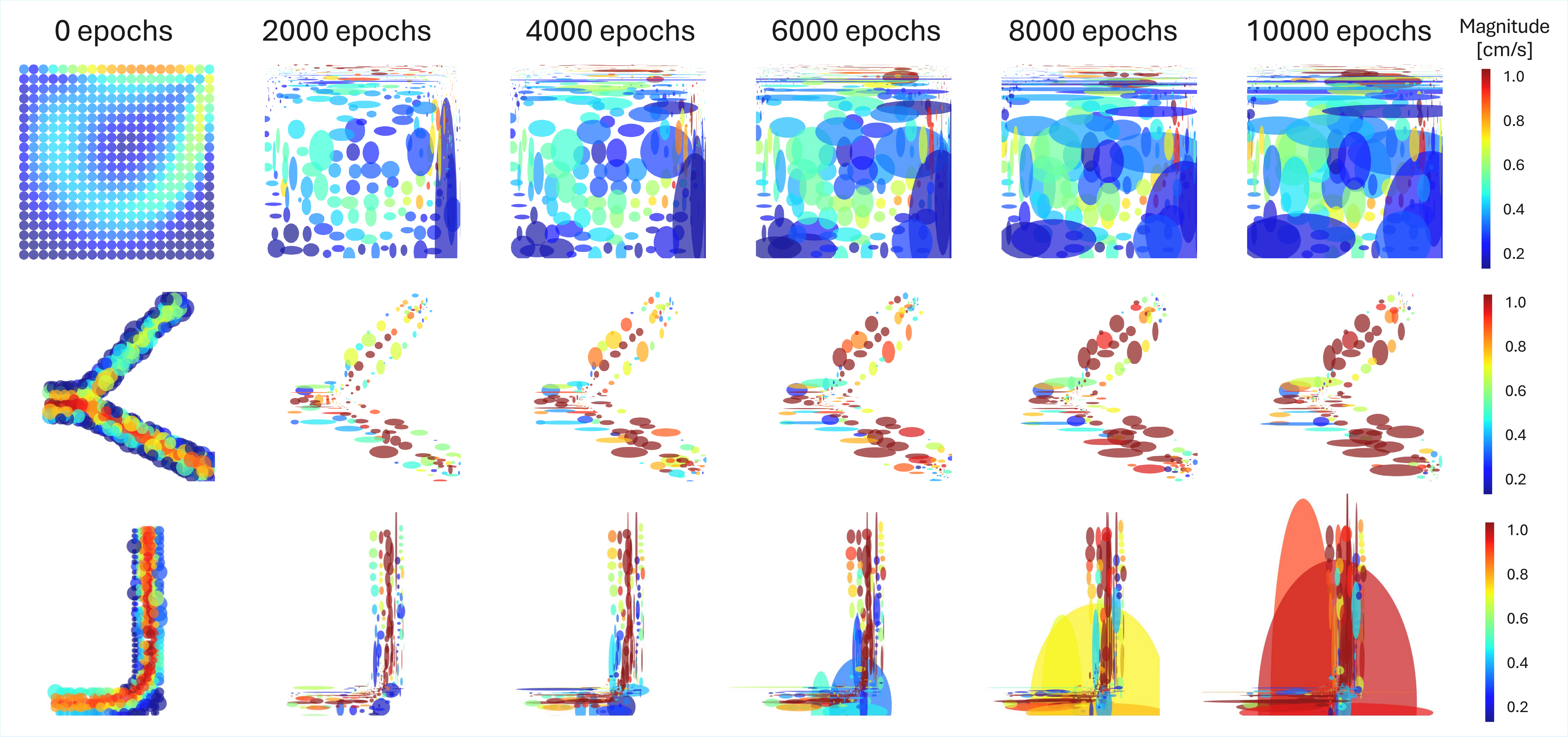}
    \caption{
         Visualization of the learned Gaussians representation's evolution on synthetic 2D CFD datasets. It showcases (top) Lid-driven, (middle) Y-shape, and (bottom) L-shape results. The color scale for each visualization is normalized to the minimum and maximum values of its respective ground truth. Note that for the Y-shape and L-shape cases, the display scale has been adjusted to enhance the clarity of flow features.
        }
    \label{fig:visualization_2D_fluid_dataset_gaussian}
\end{figure*}

\begin{figure*}[t]
\centering
    \includegraphics[width=1.0\textwidth]{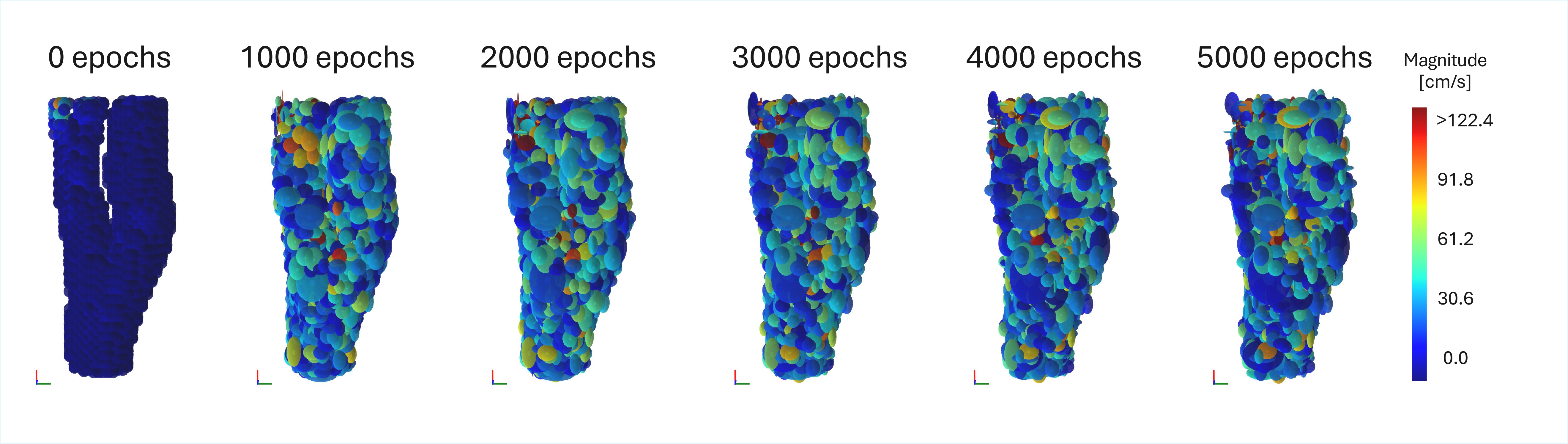}
    \caption{
         Visualization of the learned Gaussian representation's evolution on a real 4D flow MRI carotid artery dataset (Spatial averaging $\times$ 8). The color scale for the ground truth and prediction is fixed to the ground truth’s minimum and maximum values.
        }
    \label{fig:carotid_gaussian}
\end{figure*}

We visualize the spatiotemporal evolution of the Gaussian distribution to illustrate the adaptive nature of our framework.
For the synthetic CFD datasets (Fig.~\ref{fig:visualization_2D_fluid_dataset_gaussian}), the Gaussians begin on a uniform grid but are dynamically repositioned and refined during training. The final distribution shows a high concentration of Gaussians in regions with complex flow features, such as the primary vortex in the Lid-driven case and high-gradient areas in the channel flows. This adaptive behavior is also evident on the real 4D flow MRI dataset (Fig.~\ref{fig:carotid_gaussian}), where the framework dynamically allocates Gaussians to conform to the vessel anatomy and capture time-varying patterns within the carotid bifurcation.

\subsection{Additional qualitative comparisons on synthetic 2D CFD datasets} 
\begin{figure*}[t]
\centering
    \includegraphics[width=1.0\textwidth]{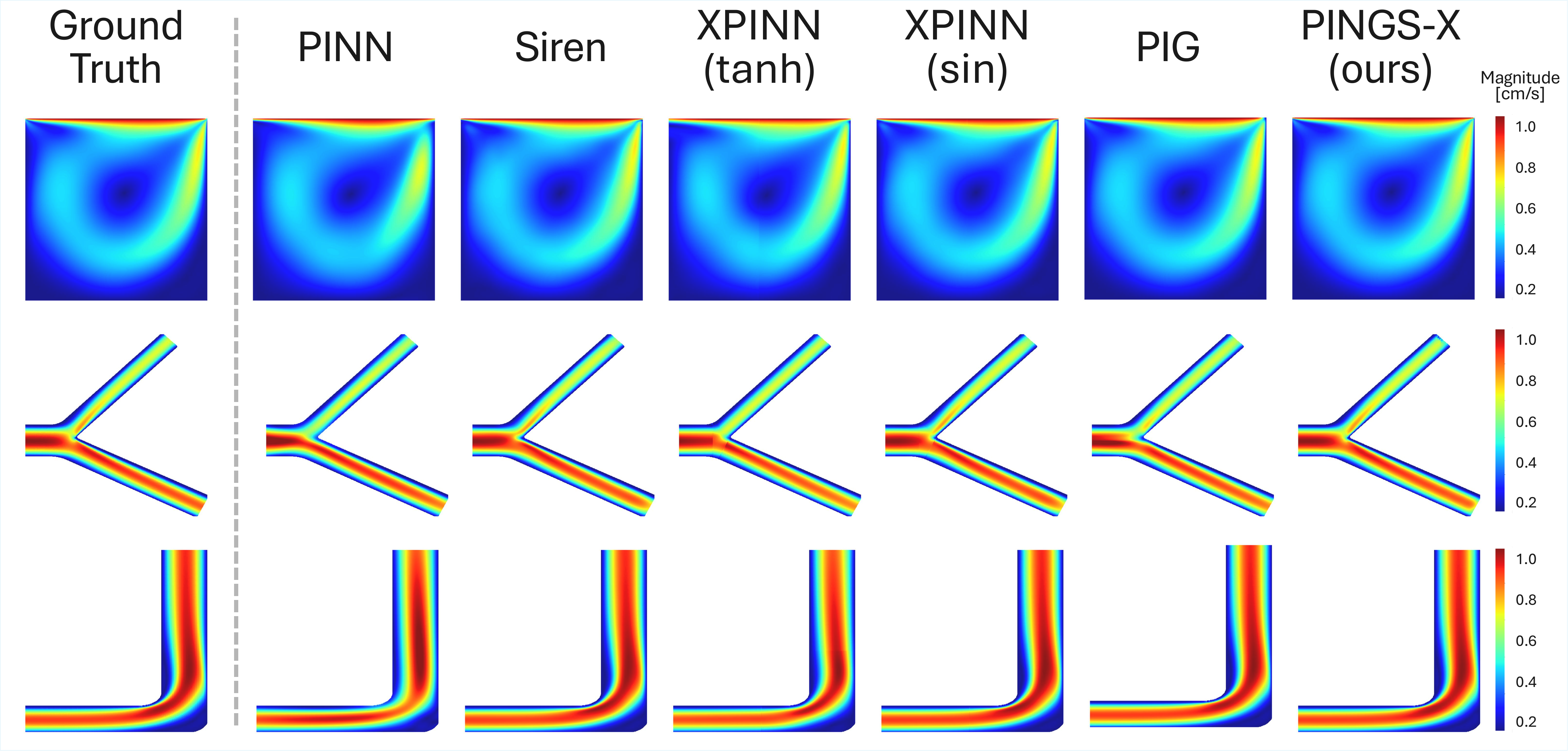}
    \caption{
         Qualitative comparison of ours with five existing models (PINN, Siren, XPINN(tanh), XPINN(sin), PIG) for 2D fluid dataset (Spatial averaging $\times16$). It showcases (top) Lid-driven, (middle) Y-shape, and (bottom) L-shape results. For the error maps (yellow = high error), the color scale is clipped to the color scale of ground truth to aid visual comparison
        }
    \label{fig:visualization_2D_fluid_dataset_pred}
\end{figure*}

\begin{figure*}[t]
\centering
    \includegraphics[width=1.0\textwidth]{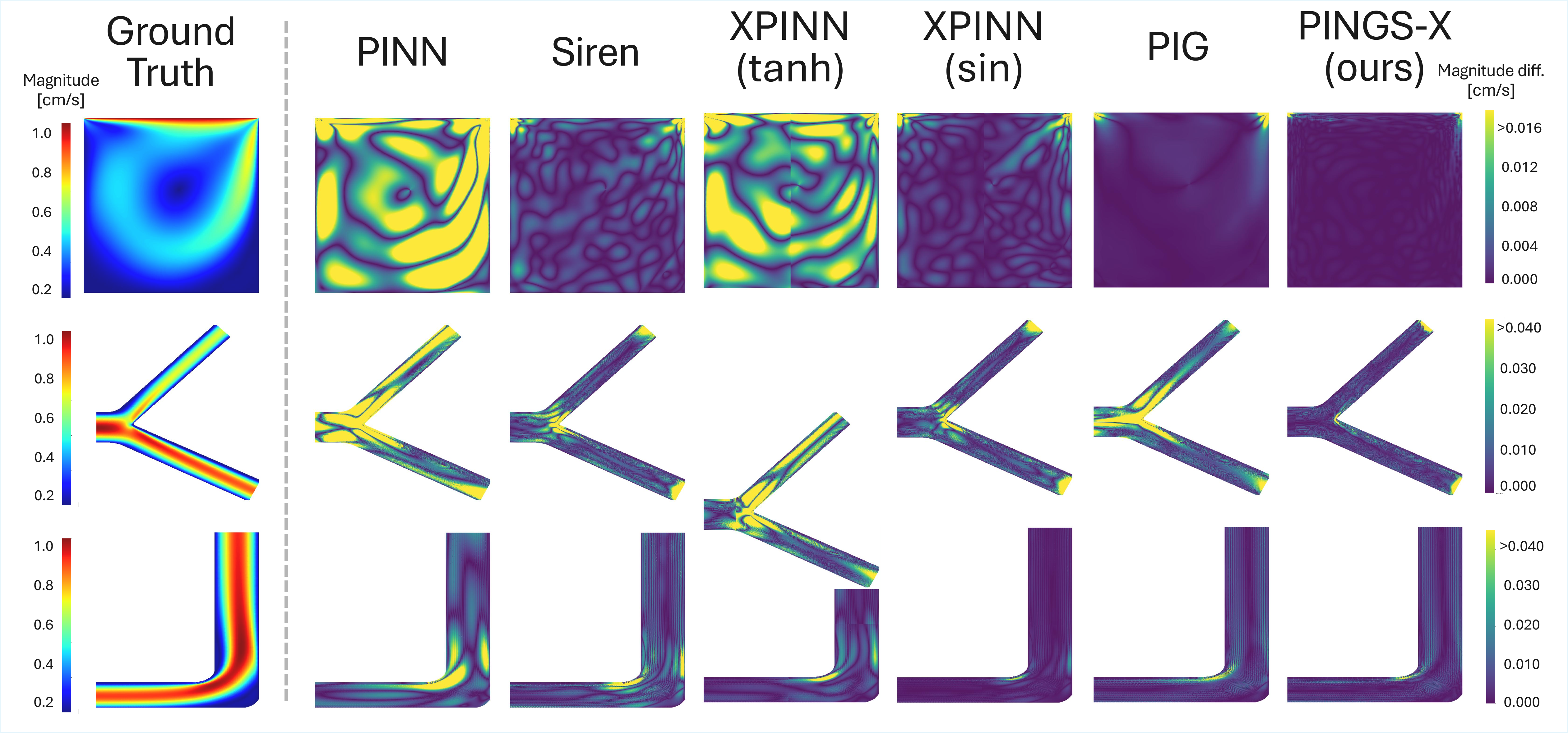}
    \caption{
         Qualitative comparison of ours with five existing models (PINN, Siren, XPINN(tanh), XPINN(sin), PIG) for 2D fluid dataset (Spatial averaging $\times16$). It showcases (top) Lid-driven results, (middle) Y-shape results, and (bottom) L-shape results. For the error maps (yellow = high error), the color scale is clipped to the mean error of the PINN baseline (most erroneous) and minimum to zero to aid visual comparison.
        }
    \label{fig:visualization_2D_fluid_dataset_error}
\end{figure*}

To qualitatively assess our model, we present visual comparisons on the 2D fluid datasets with a spatial averaging factor of $\times$16. We provide two visualizations for a comprehensive analysis. 
In Fig.~\ref{fig:visualization_2D_fluid_dataset_pred}, the color scale is scaled relative to the ground truth's value range.
For a more direct comparison between models, in Fig.~\ref{fig:visualization_2D_fluid_dataset_error}, the error maps are clipped to the mean error of the PINN baseline to provide a fair comparison against the worst-performing baseline. In both figures, across all three test cases, the error maps for our proposed model appear visibly lower than those of other models.

\subsection{Additional (per-time frame) qualitative analysis on  real 4D flow MRI dataset}

\begin{figure*}[t]
  \centering
  \begin{subfigure}[b]{0.55\textwidth}
    \centering
    \includegraphics[width=0.9\textwidth]{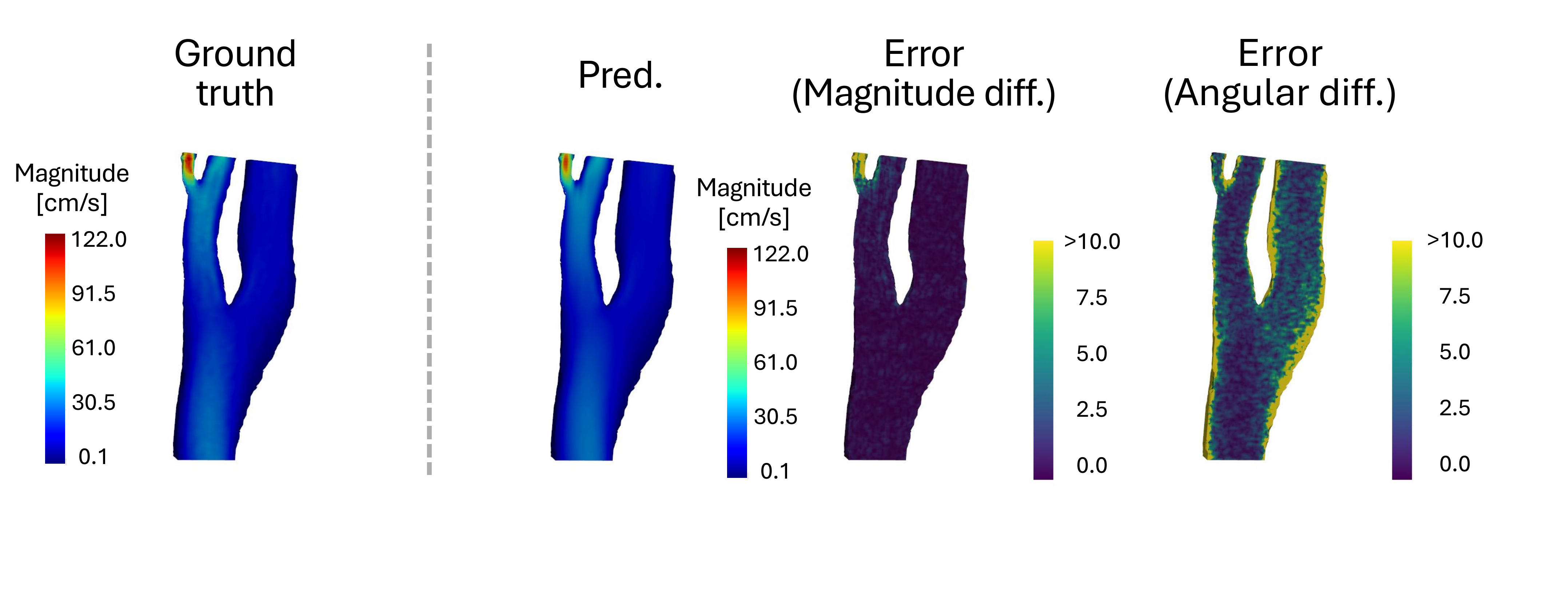}
    \caption{PINGS-X (timeframe=10)}
    \label{fig:sub1}
  \end{subfigure}
  
  \begin{subfigure}[b]{0.55\textwidth}
    \centering
    \includegraphics[width=0.9\textwidth]{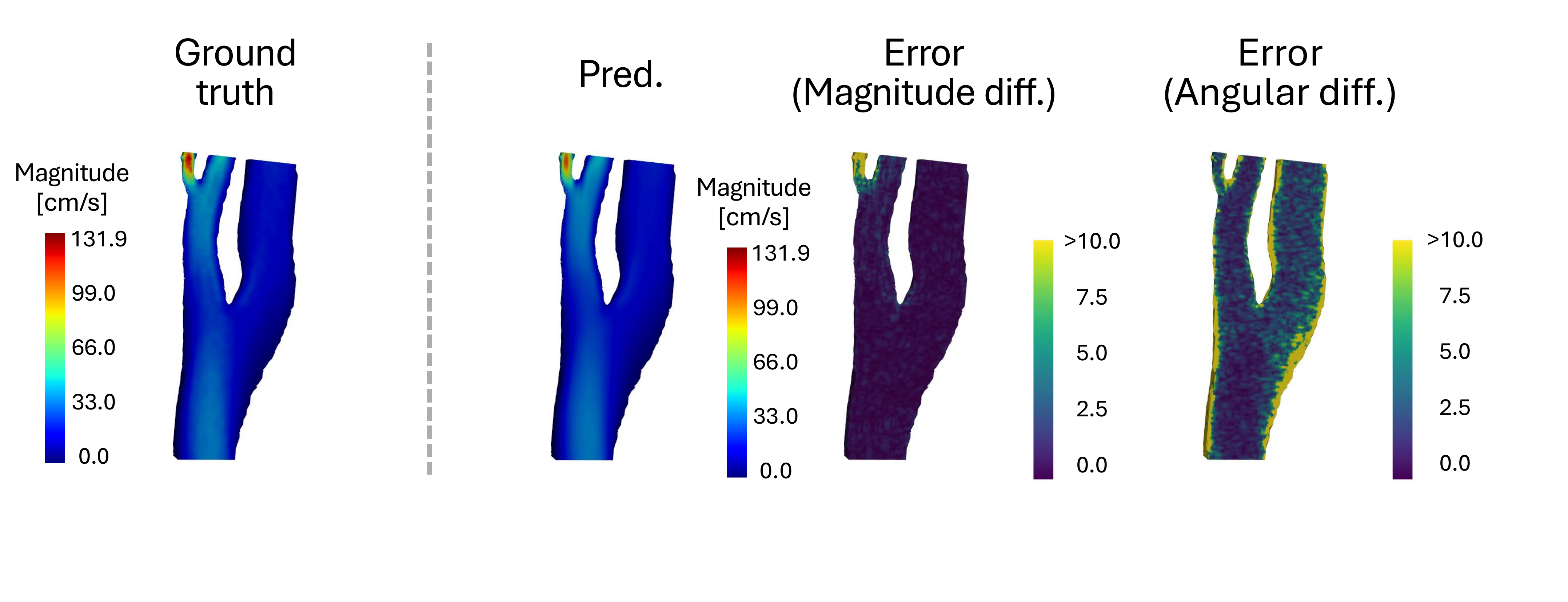}
    \caption{PINGS-X (timeframe=11)}
    \label{fig:sub2}
  \end{subfigure}
  
  \begin{subfigure}[b]{0.55\textwidth}
    \centering
    \includegraphics[width=0.9\textwidth]{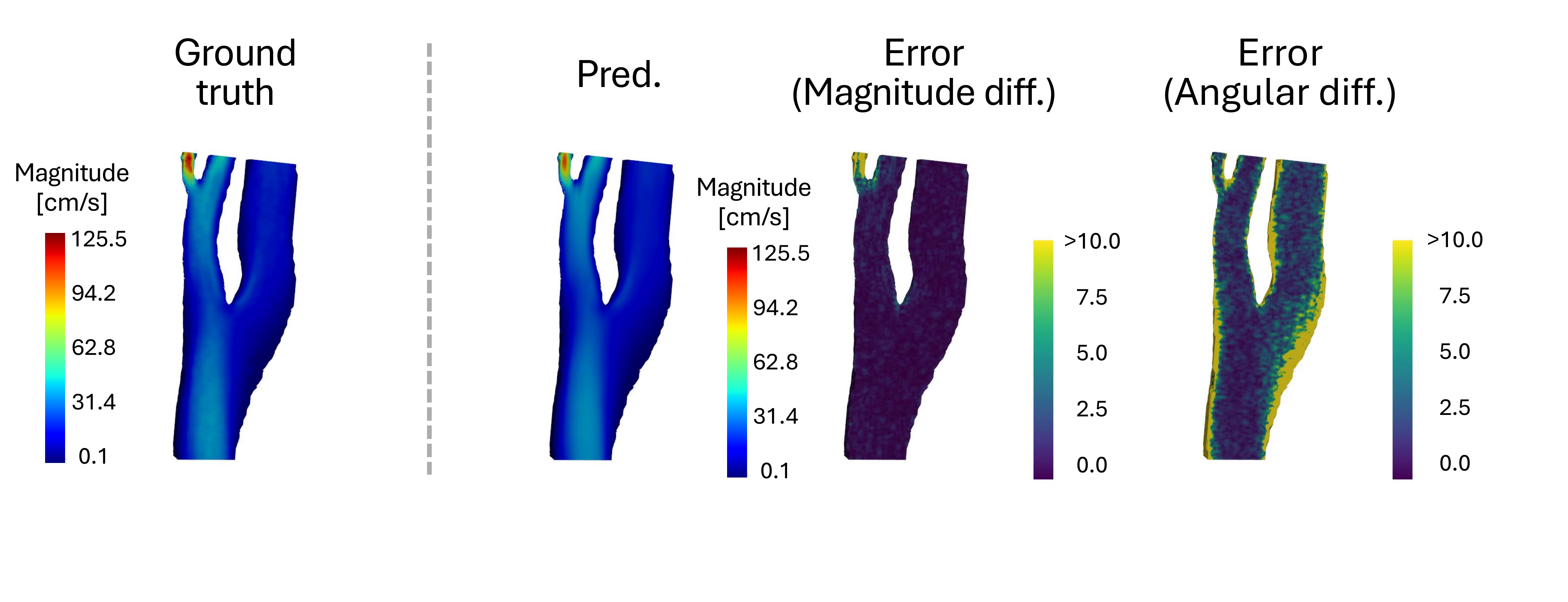}
    \caption{PINGS-X (timeframe=12)}
    \label{fig:sub3}
  \end{subfigure}
  
  \begin{subfigure}[b]{0.55\textwidth}
    \centering
    \includegraphics[width=0.9\textwidth]{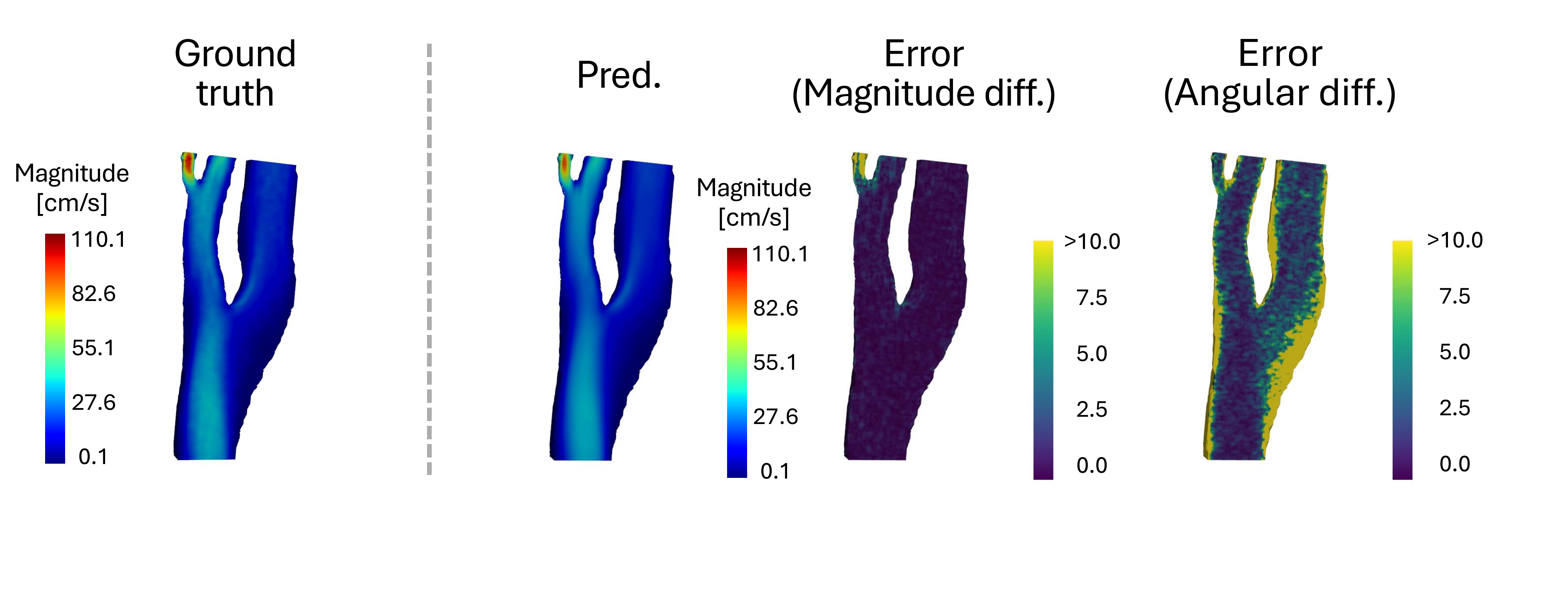}
    \caption{PINGS-X (timeframe=13)}
    \label{fig:sub4}
    \end{subfigure}

  \begin{subfigure}[b]{0.55\textwidth}
    \centering
    \includegraphics[width=0.9\textwidth]{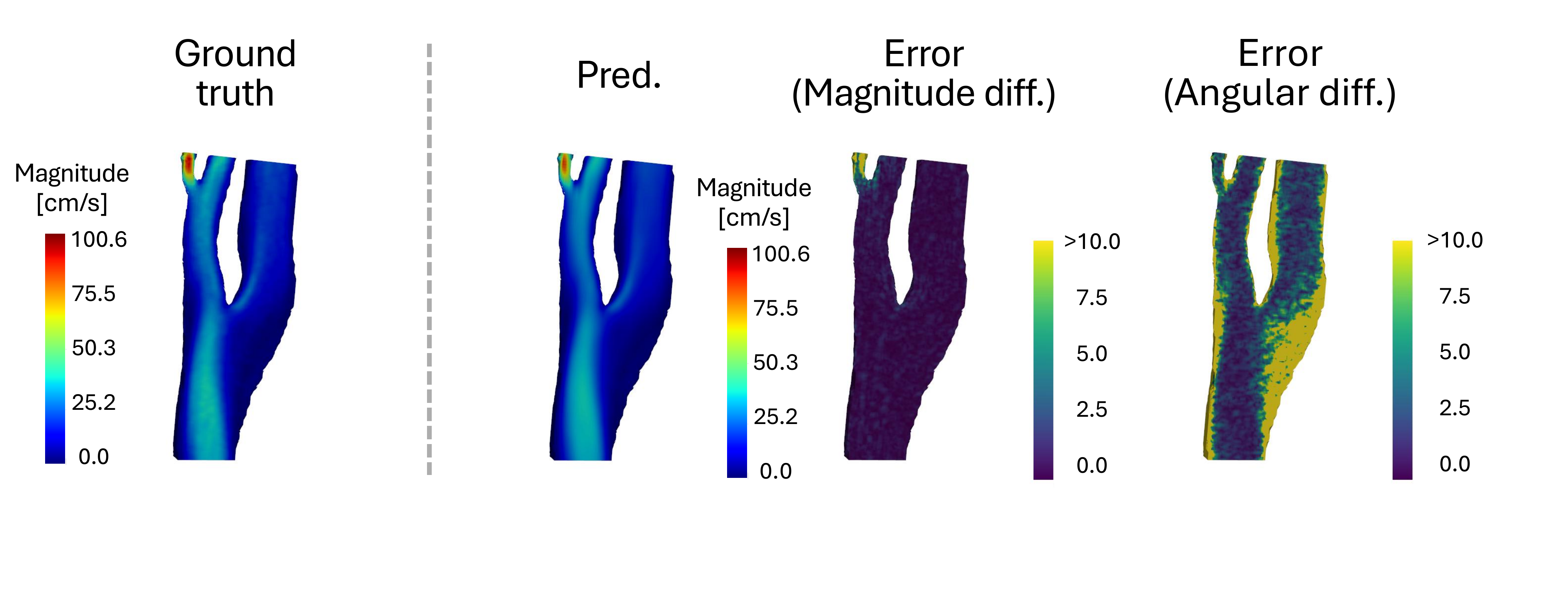}
    \caption{PINGS-X (timeframe=14)}
    \label{fig:sub5}
  \end{subfigure}
  \caption{Visualizations of our PINGS-X model on $\times 8$ spatially averaged 4D flow MRI dataset over five time frames during the peak systolic phase (10 -- 14). The figure visualizes ground truth velocity field and prediction error maps (magnitude and direction). The color scale for the ground truth and prediction is fixed to the ground truth's minimum and maximum values. For error maps, color scale is fixed at [0,10] for both error maps (yellow = high error).}
  \label{fig:timeframe_X8}
\end{figure*}

\begin{figure*}[t]
  \centering
  \begin{subfigure}[b]{0.55\textwidth}
    \centering
    \includegraphics[width=0.9\textwidth]{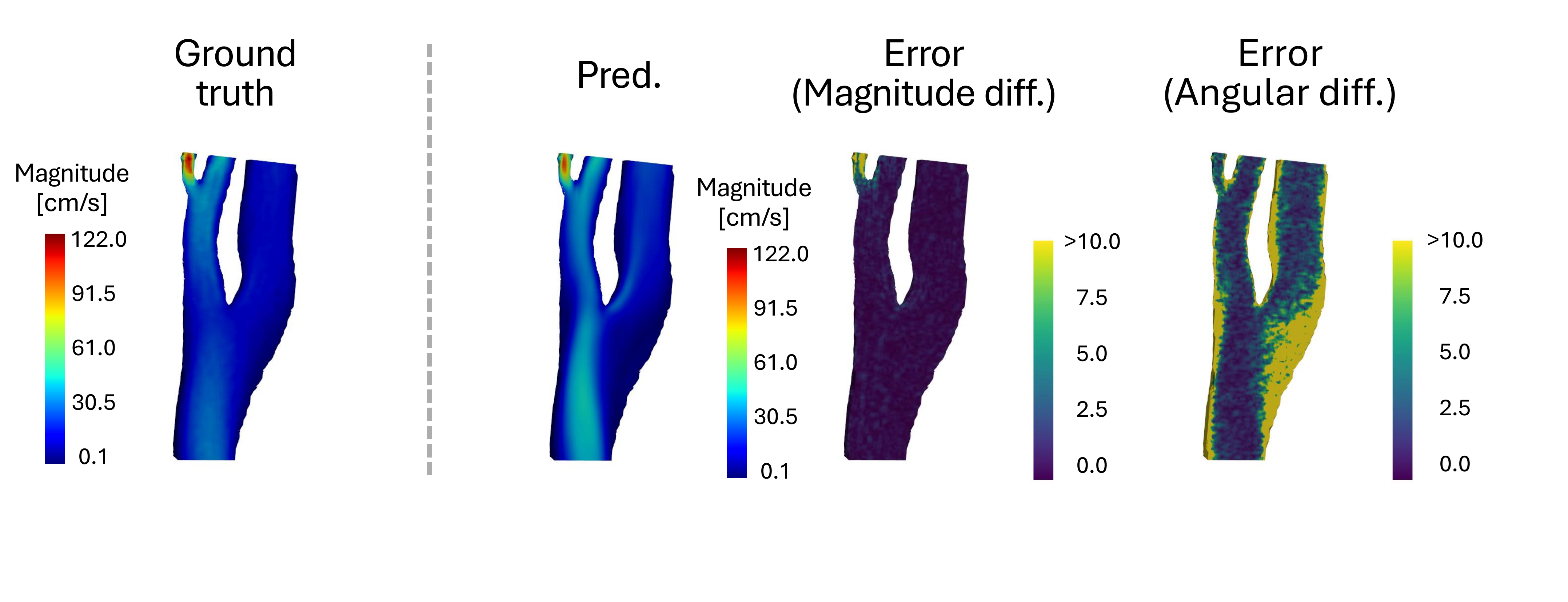}
    \caption{PINGS-X (timeframe=10)}
    \label{fig:sub6}
  \end{subfigure}
  
  \begin{subfigure}[b]{0.55\textwidth}
    \centering
    \includegraphics[width=0.9\textwidth]{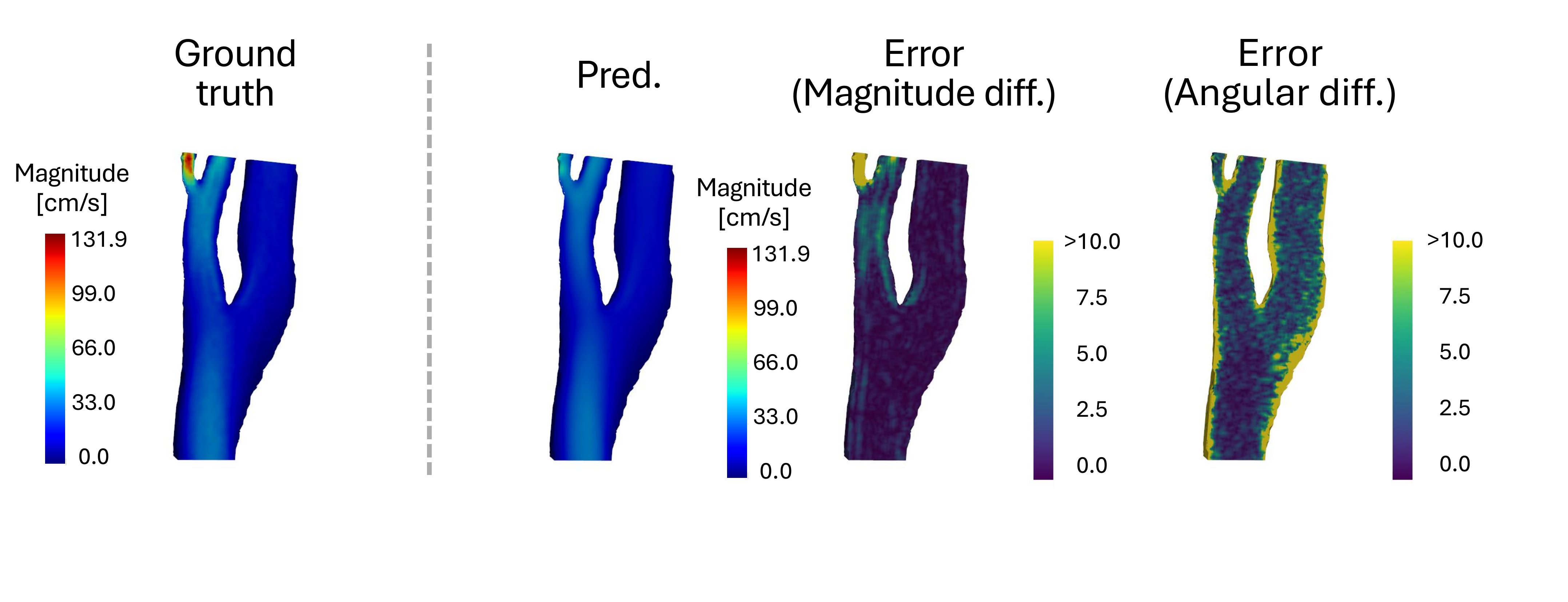}
    \caption{PINGS-X (timeframe=11)}
    \label{fig:sub7}
  \end{subfigure}
  
  \begin{subfigure}[b]{0.55\textwidth}
    \centering
    \includegraphics[width=0.9\textwidth]{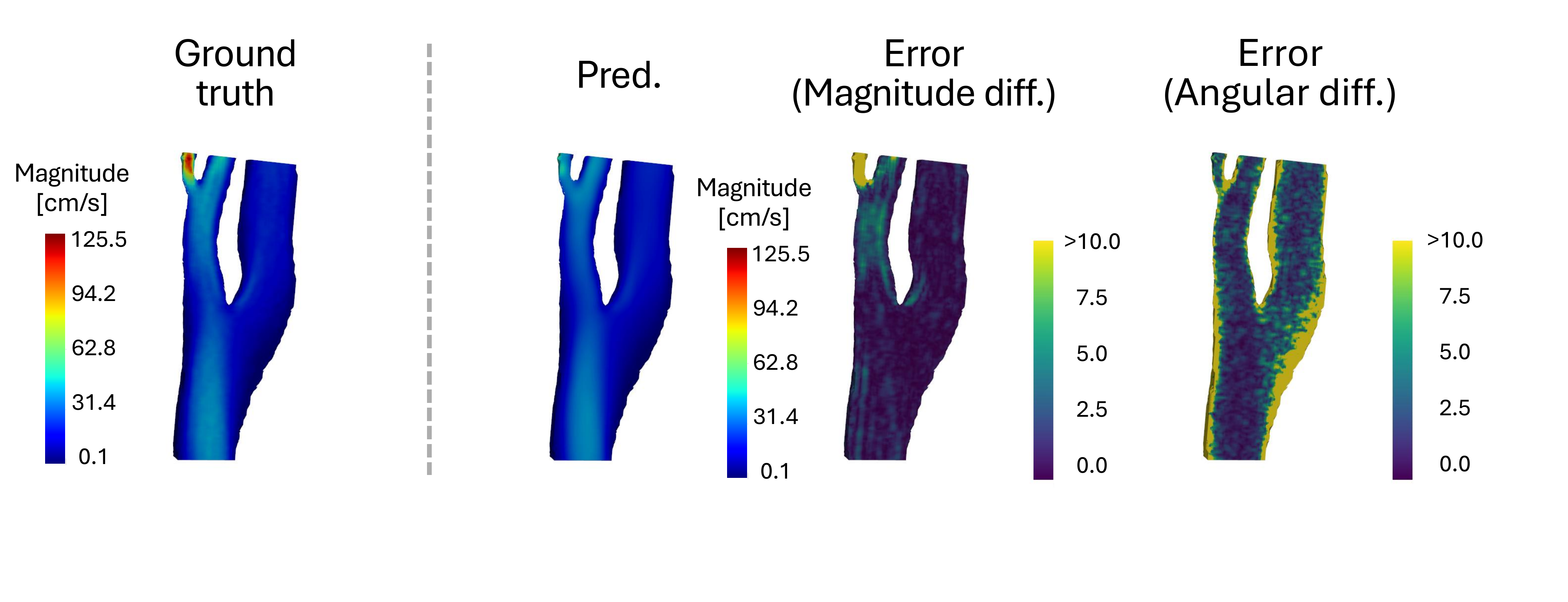}
    \caption{PINGS-X (timeframe=12)}
    \label{fig:sub8}
  \end{subfigure}
  
  \begin{subfigure}[b]{0.55\textwidth}
    \centering
    \includegraphics[width=0.9\textwidth]{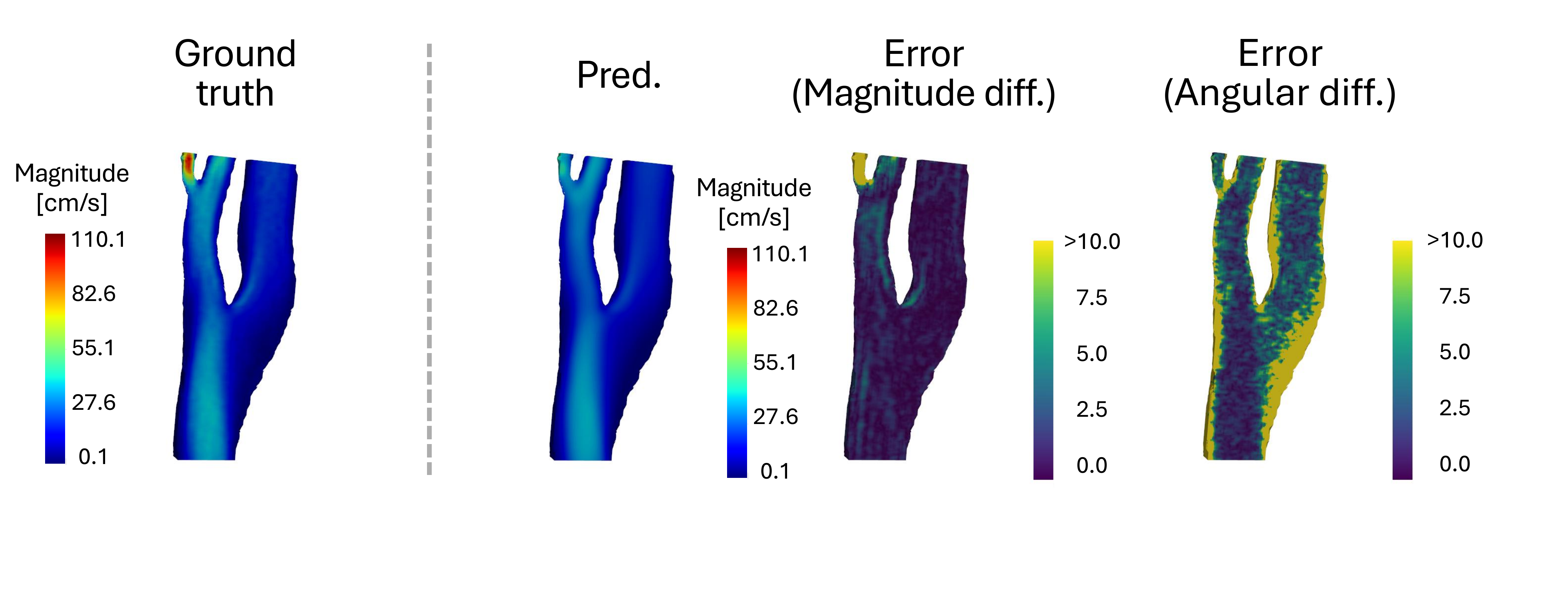}
    \caption{PINGS-X (timeframe=13)}
    \label{fig:sub9}
  \end{subfigure}
  
  \begin{subfigure}[b]{0.55\textwidth}
    \centering
    \includegraphics[width=0.9\textwidth]{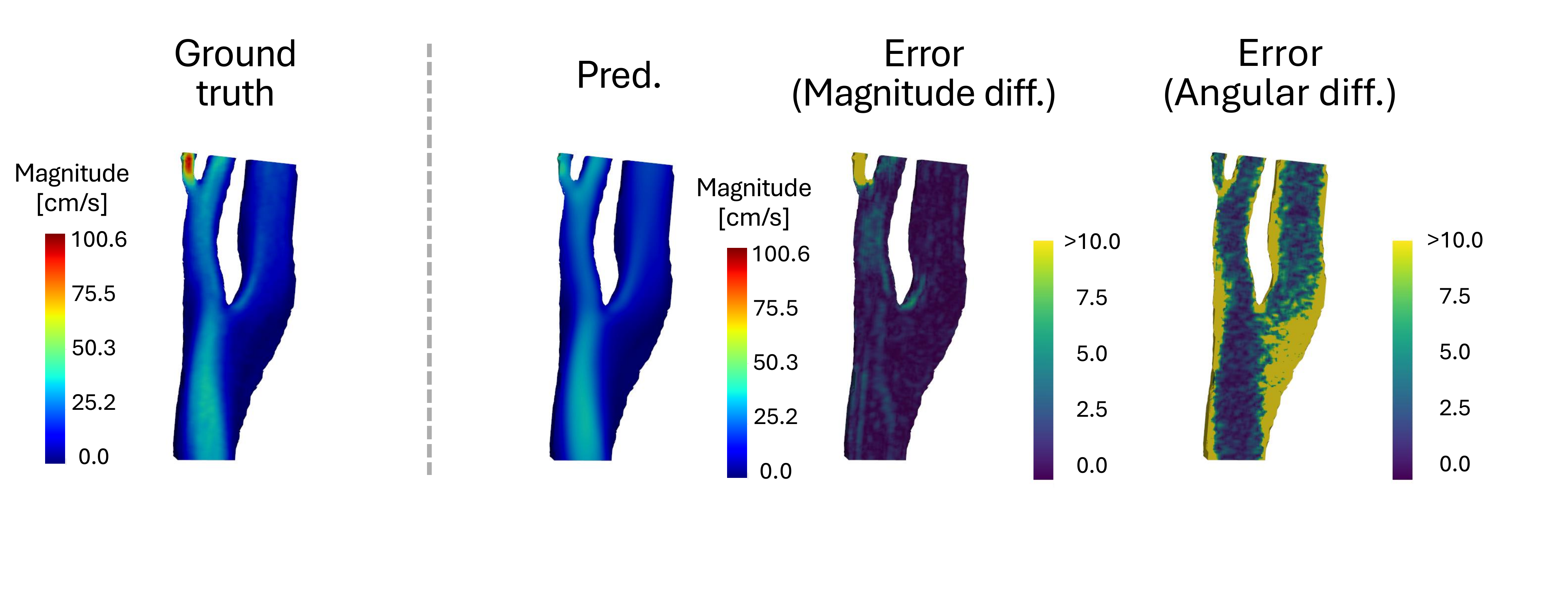}
    \caption{PINGS-X (timeframe=14)}
    \label{fig:sub10}
  \end{subfigure}
    \caption{
   Visualizations of our PINGS-X model on $\times 64$ spatially averaged 4D flow MRI dataset over five time frames during the peak systolic phase (10 -- 14). The figure visualizes ground truth velocity field and prediction error maps (magnitude and direction). The color scale for the ground truth and prediction is fixed to the ground truth's minimum and maximum values. For error maps, color scale is fixed at [0,10] for both error maps (yellow = high error). 
    }
     \label{fig:timeframe_X64}
\end{figure*}

\begin{figure*}[t]
  \centering
  \begin{subfigure}[b]{0.55\textwidth}
    \centering
    \includegraphics[width=0.9\textwidth]{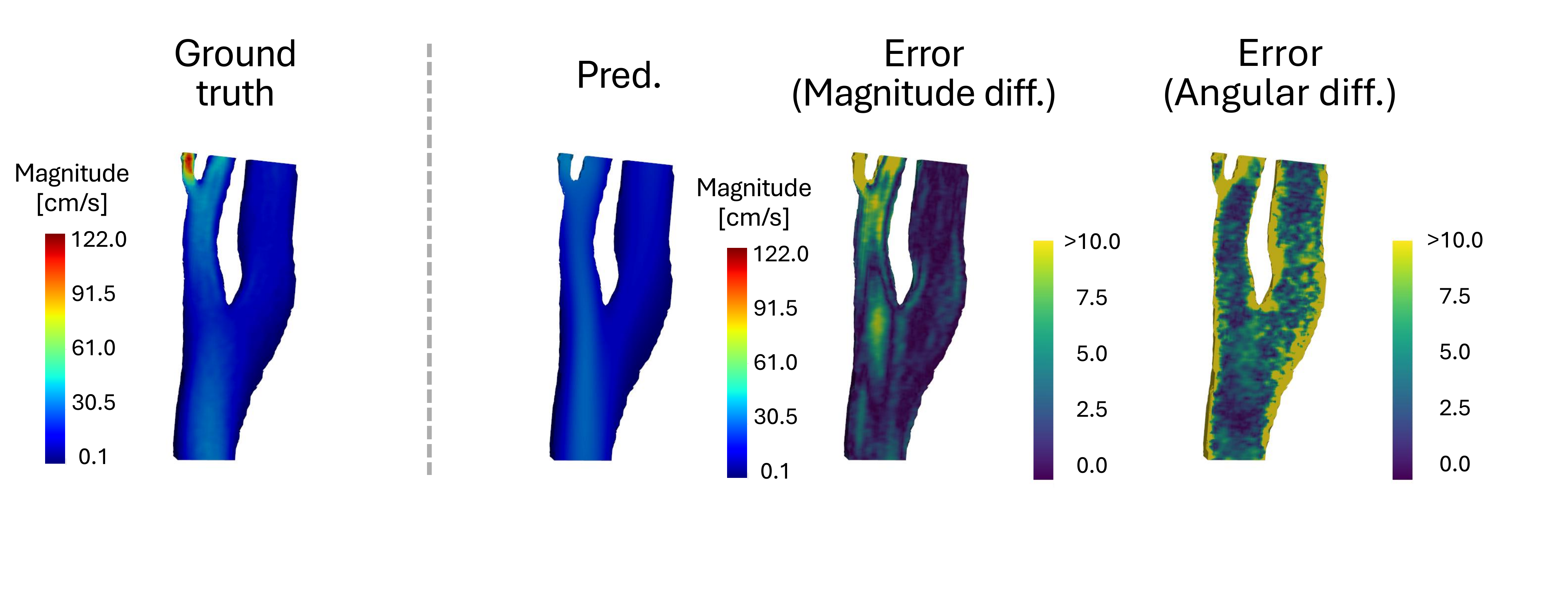}
    \caption{PINN (timeframe=10)}
    \label{fig:sub6_PINN}
  \end{subfigure}
  \begin{subfigure}[b]{0.55\textwidth}
    \centering
    \includegraphics[width=0.9\textwidth]{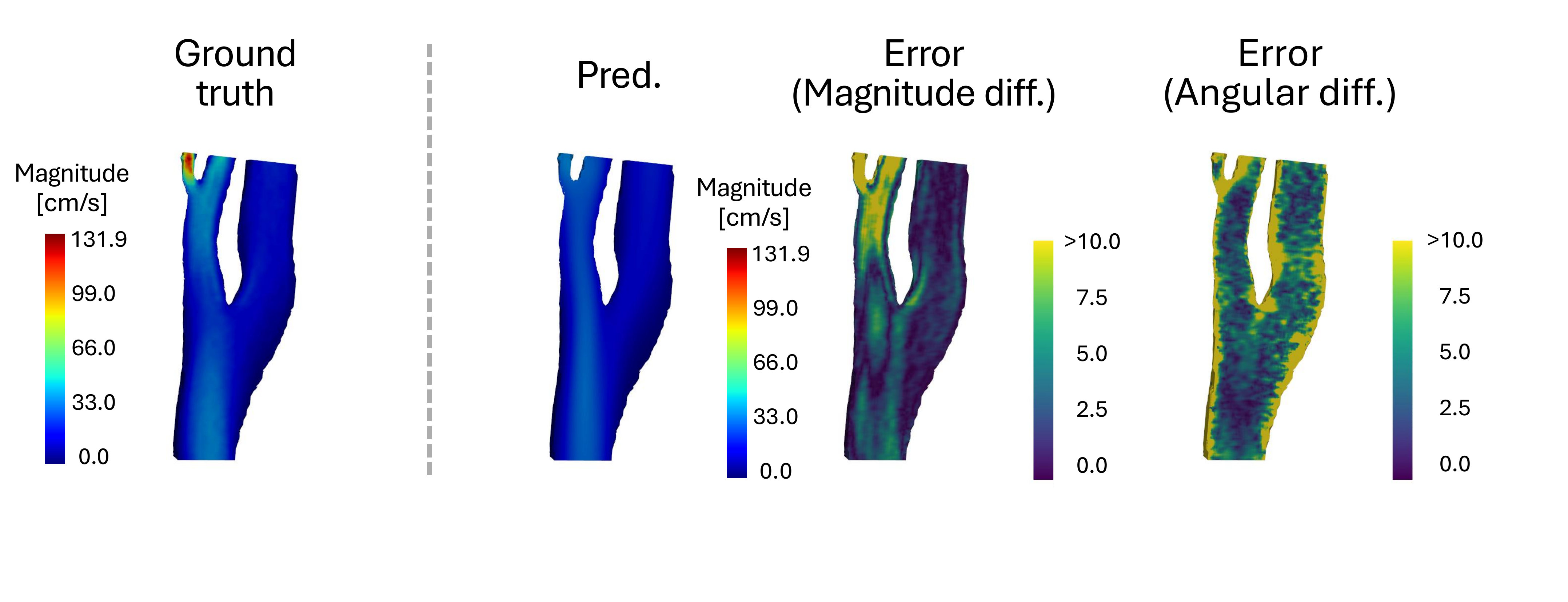}
    \caption{PINN (timeframe=11)}
    \label{fig:sub7_PINN}
  \end{subfigure}
  \begin{subfigure}[b]{0.55\textwidth}
    \centering
    \includegraphics[width=0.9\textwidth]{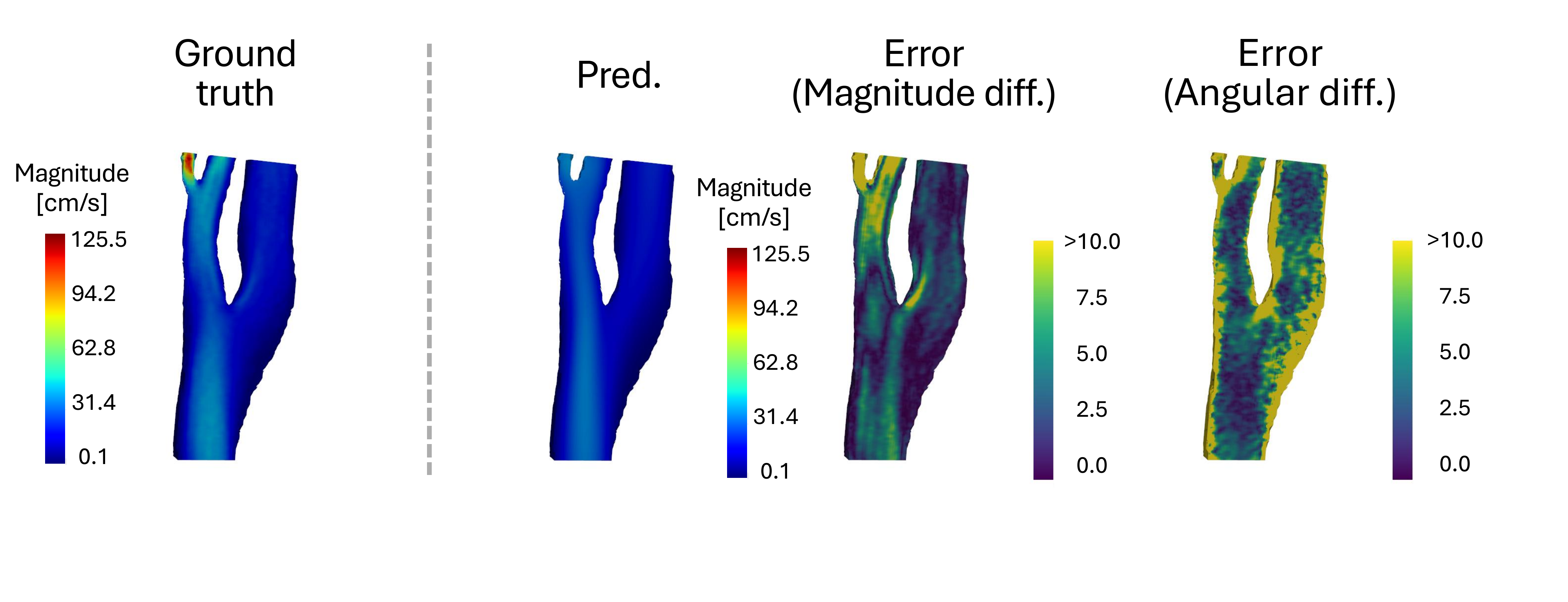}
    \caption{PINN (timeframe=12)}
    \label{fig:sub8_PINN}
  \end{subfigure}
  \begin{subfigure}[b]{0.55\textwidth}
    \centering
    \includegraphics[width=0.9\textwidth]{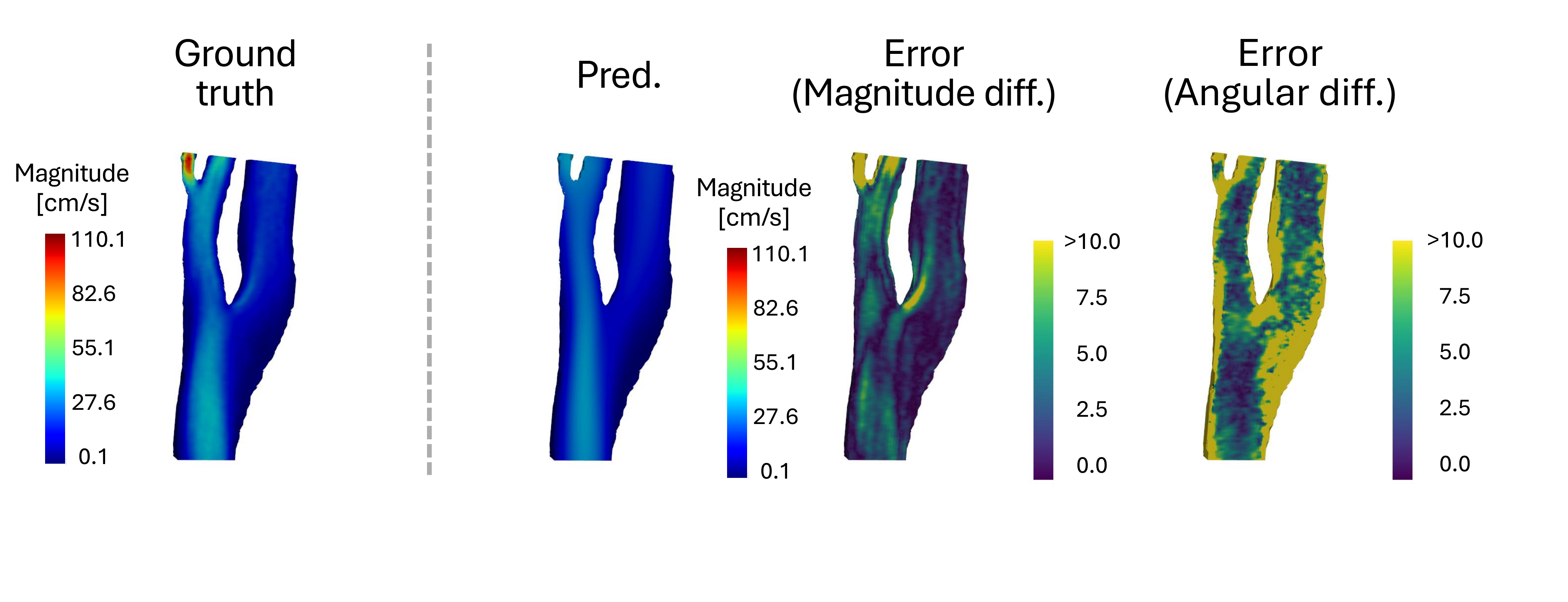}
    \caption{PINN (timeframe=13)}
    \label{fig:sub9_PINN}
  \end{subfigure}
  \begin{subfigure}[b]{0.55\textwidth}
    \centering
    \includegraphics[width=0.9\textwidth]{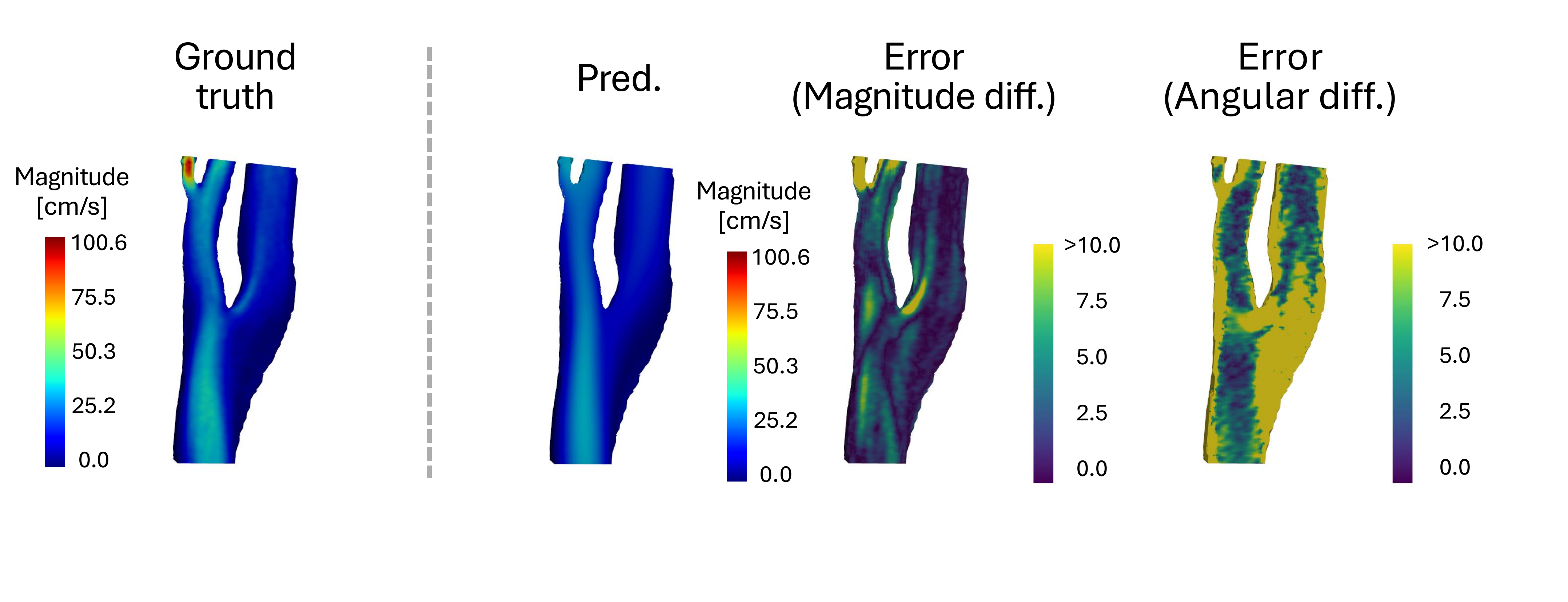}
    \caption{PINN (timeframe=14)}
    \label{fig:sub10_PINN}
  \end{subfigure}
    \caption{Visualizations of the PINN model on $\times 8$ spatially averaged 4D flow MRI dataset over five time frames during the peak systolic phase (10 -- 14). The figure visualizes ground truth velocity field and prediction error maps (magnitude and direction). The color scale for the ground truth and prediction is fixed to the ground truth's minimum and maximum values. For error maps, color scale is fixed at [0,10] for both error maps (yellow = high error). }
     \label{fig:timeframe_X8_PINN}
\end{figure*}

\begin{figure*}[t]
  \centering
  \begin{subfigure}[b]{0.55\textwidth}
    \centering
    \includegraphics[width=0.9\textwidth]{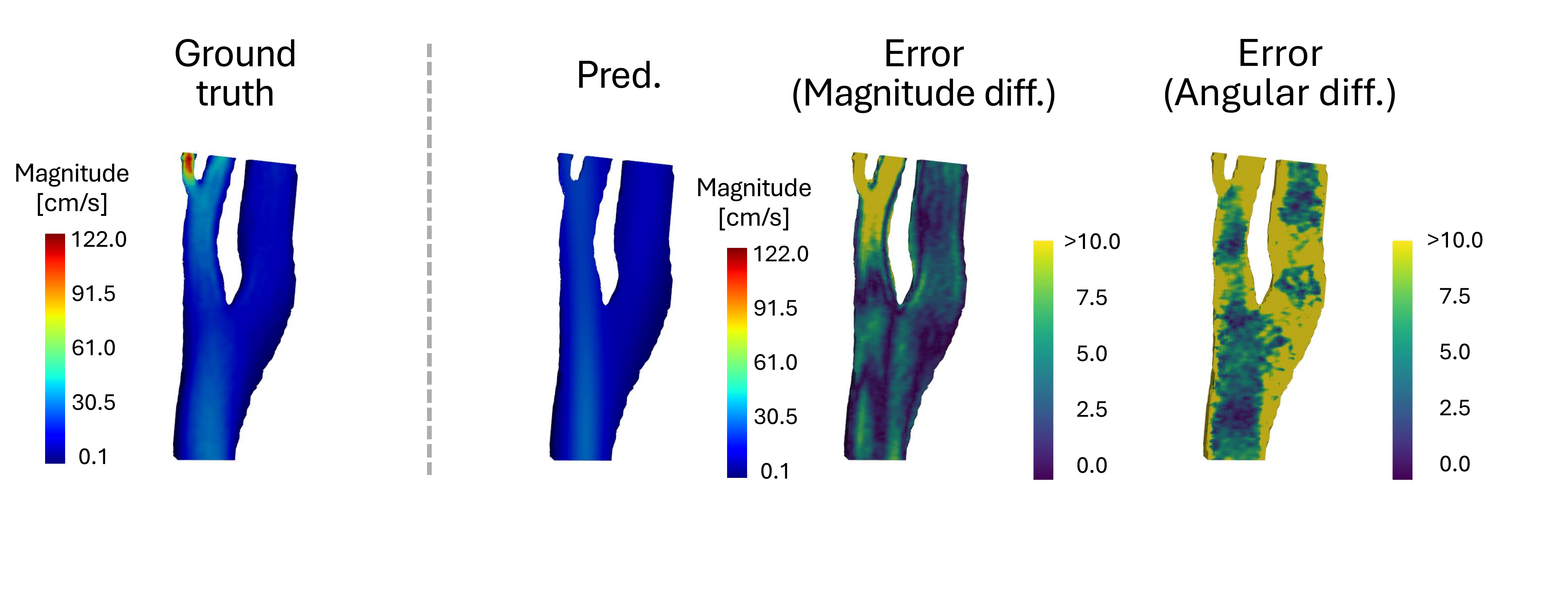}
    \caption{PINN (timeframe=10)}
    \label{fig:sub1_PINN}
  \end{subfigure}
  
  \begin{subfigure}[b]{0.55\textwidth}
    \centering
    \includegraphics[width=0.9\textwidth]{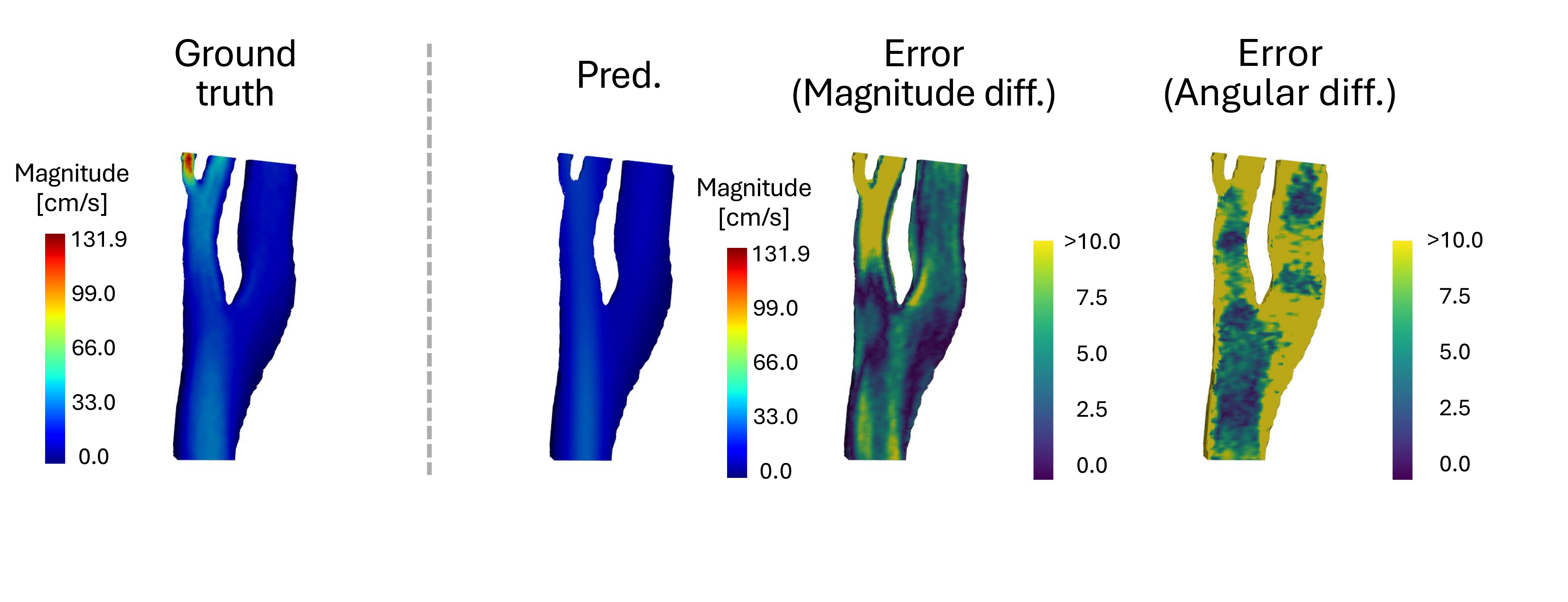}
    \caption{PINN (timeframe=11)}
    \label{fig:sub2_PINN}
  \end{subfigure}
  
  \begin{subfigure}[b]{0.55\textwidth}
    \centering
    \includegraphics[width=0.9\textwidth]{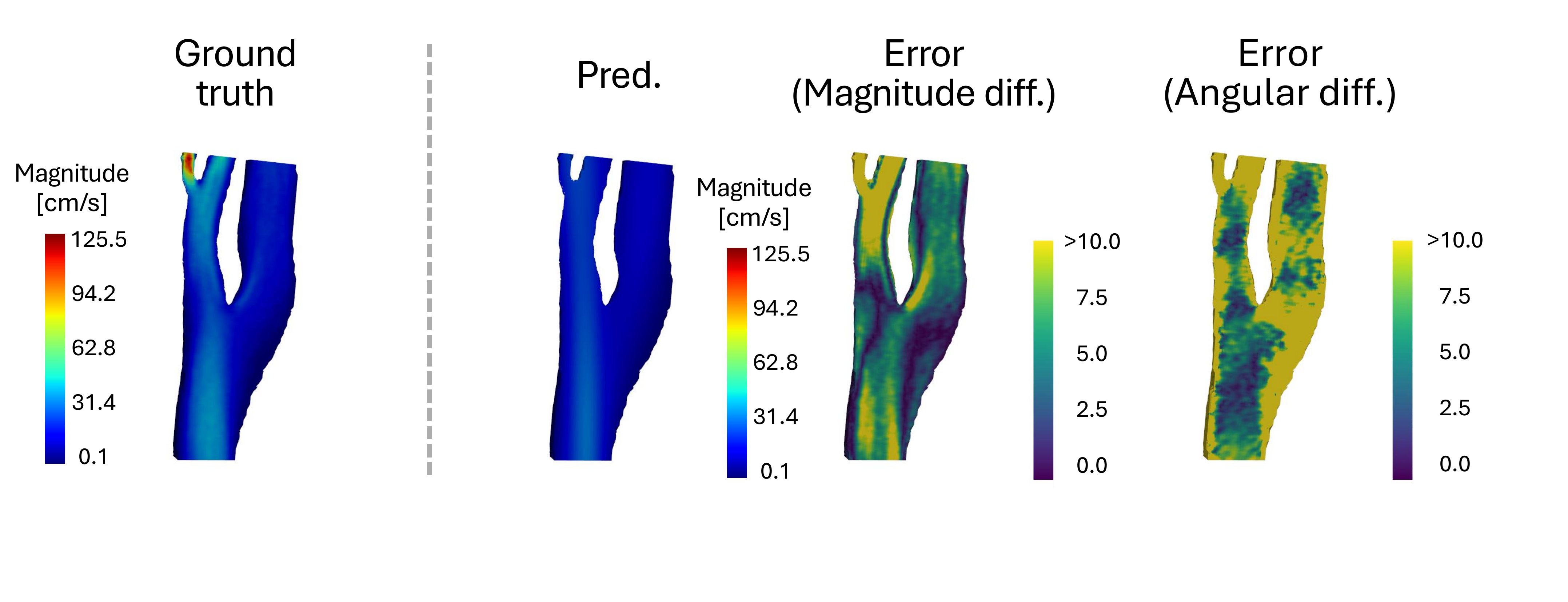}
    \caption{PINN (timeframe=12)}
    \label{fig:sub3_PINN}
  \end{subfigure}
  
  \begin{subfigure}[b]{0.55\textwidth}
    \centering
    \includegraphics[width=0.9\textwidth]{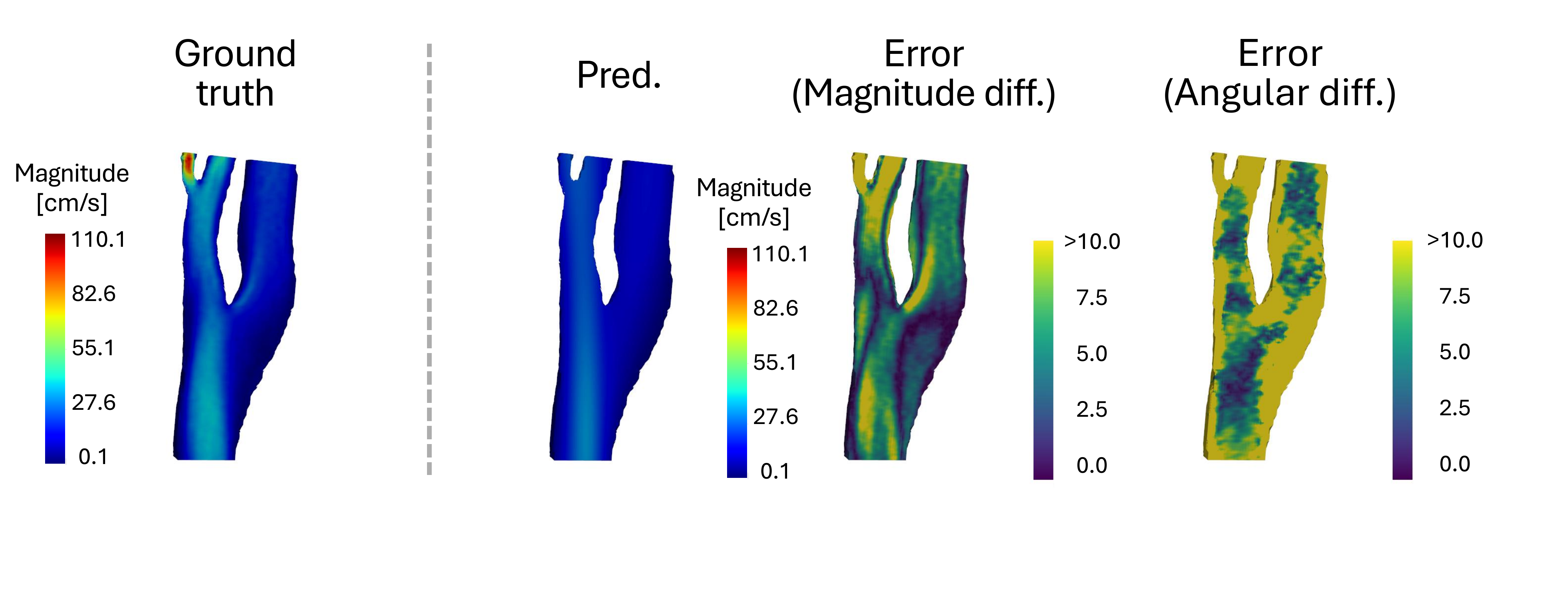}
    \caption{PINN (timeframe=13)}
    \label{fig:sub4_PINN}
  \end{subfigure}
  
  \begin{subfigure}[b]{0.55\textwidth}
    \centering
    \includegraphics[width=0.9\textwidth]{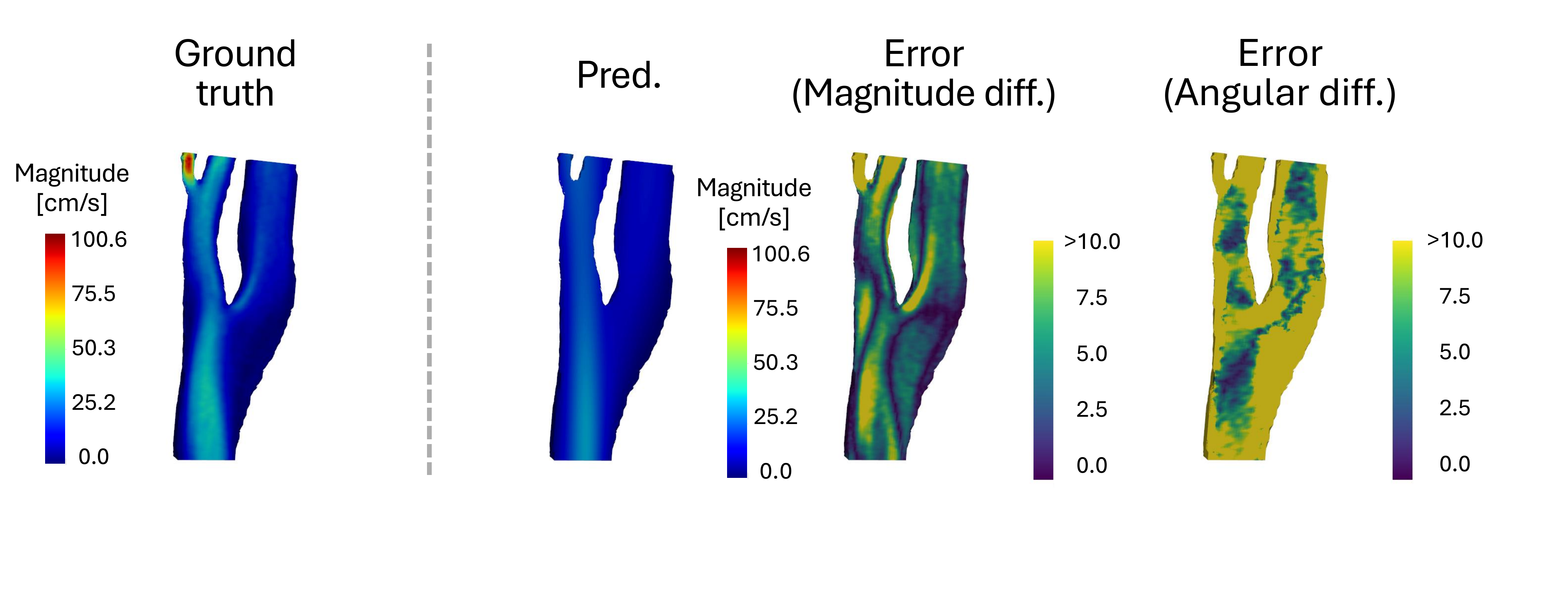}
    \caption{PINN (timeframe=14)}
    \label{fig:sub5_PINN}
  \end{subfigure}
    \caption{Visualizations of the PINN model on $\times 64$ spatially averaged 4D flow MRI dataset over five time frames during the peak systolic phase (10 -- 14). The figure visualizes ground truth velocity field and prediction error maps (magnitude and direction). The color scale for the ground truth and prediction is fixed to the ground truth's minimum and maximum values. For error maps, color scale is fixed at [0,10] for both error maps (yellow = high error).}
     \label{fig:timeframe_X64_PINN}
\end{figure*}

\begin{figure*}[t]
  \centering
  \begin{subfigure}[b]{0.55\textwidth}
  \centering
    \includegraphics[width=0.9\textwidth]{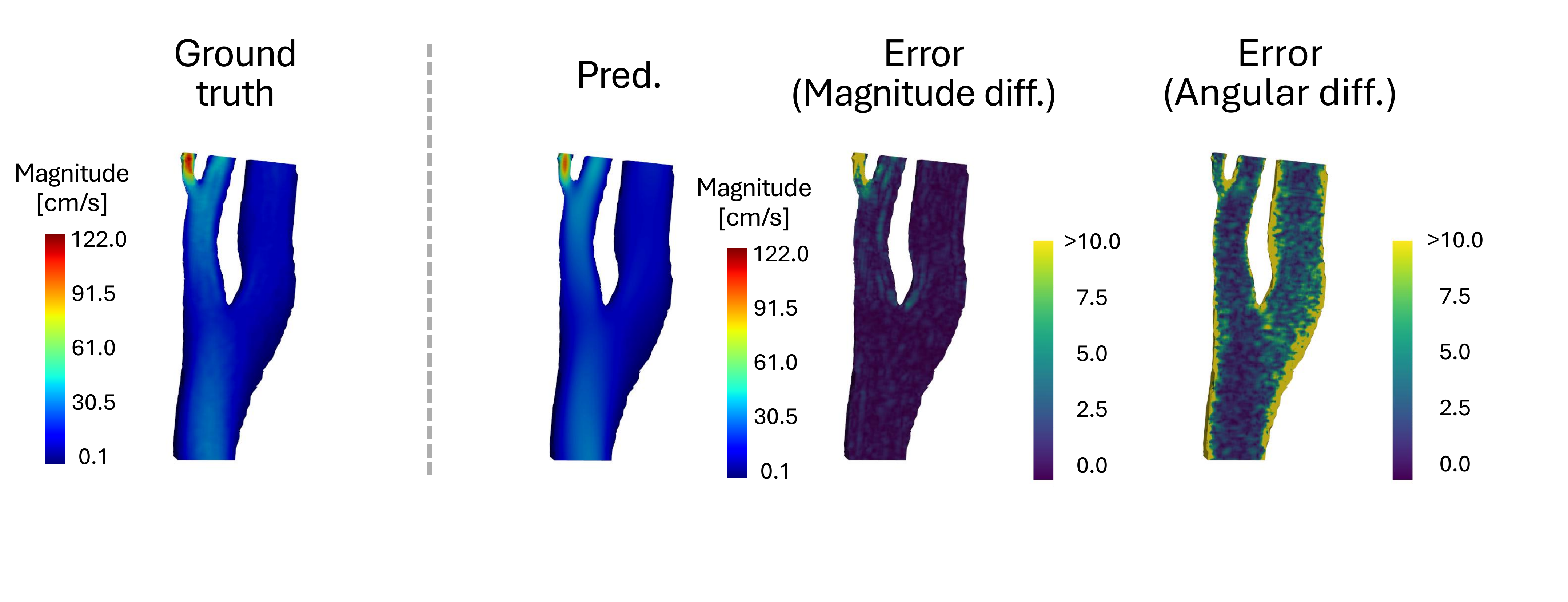}
    \caption{Siren (timeframe=10)}
    \label{fig:sub1_Siren}
  \end{subfigure}
  
  \begin{subfigure}[b]{0.55\textwidth}
    \centering
    \includegraphics[width=0.9\textwidth]{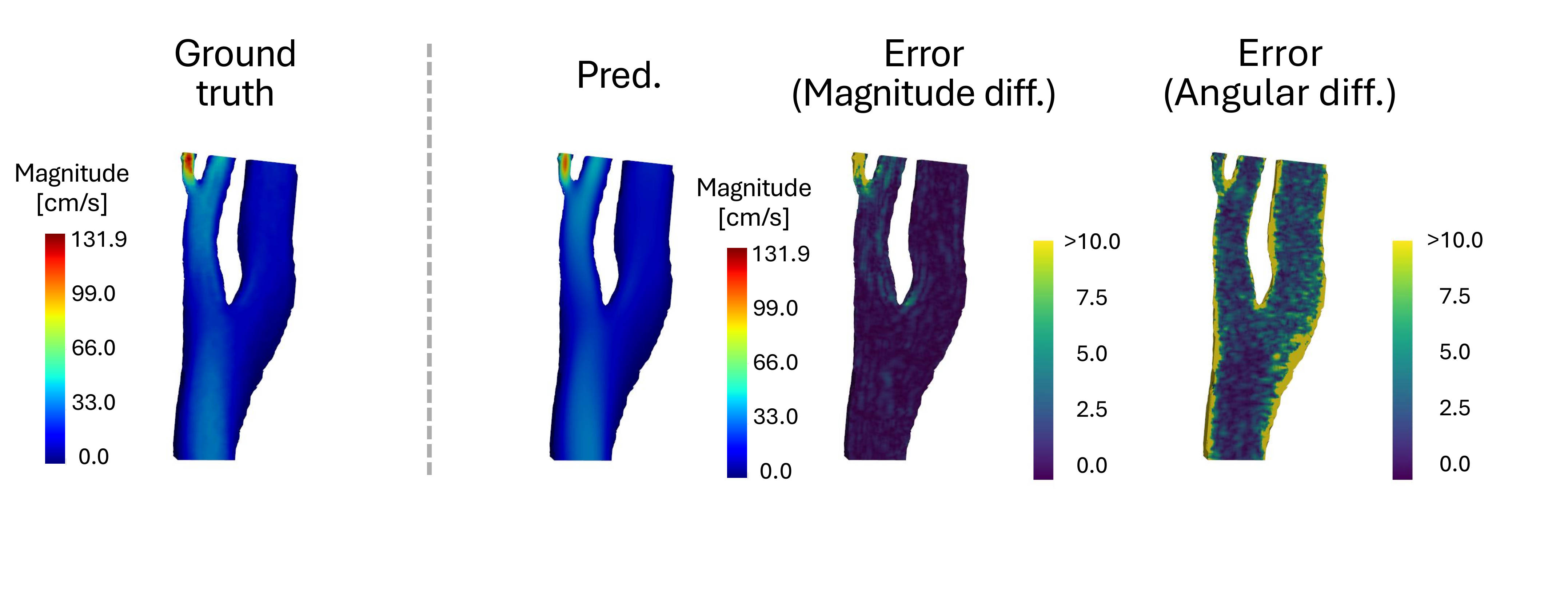}
    \caption{Siren (timeframe=11)}
    \label{fig:sub2_Siren}
  \end{subfigure}
  
  \begin{subfigure}[b]{0.55\textwidth}
    \centering
    \includegraphics[width=0.9\textwidth]{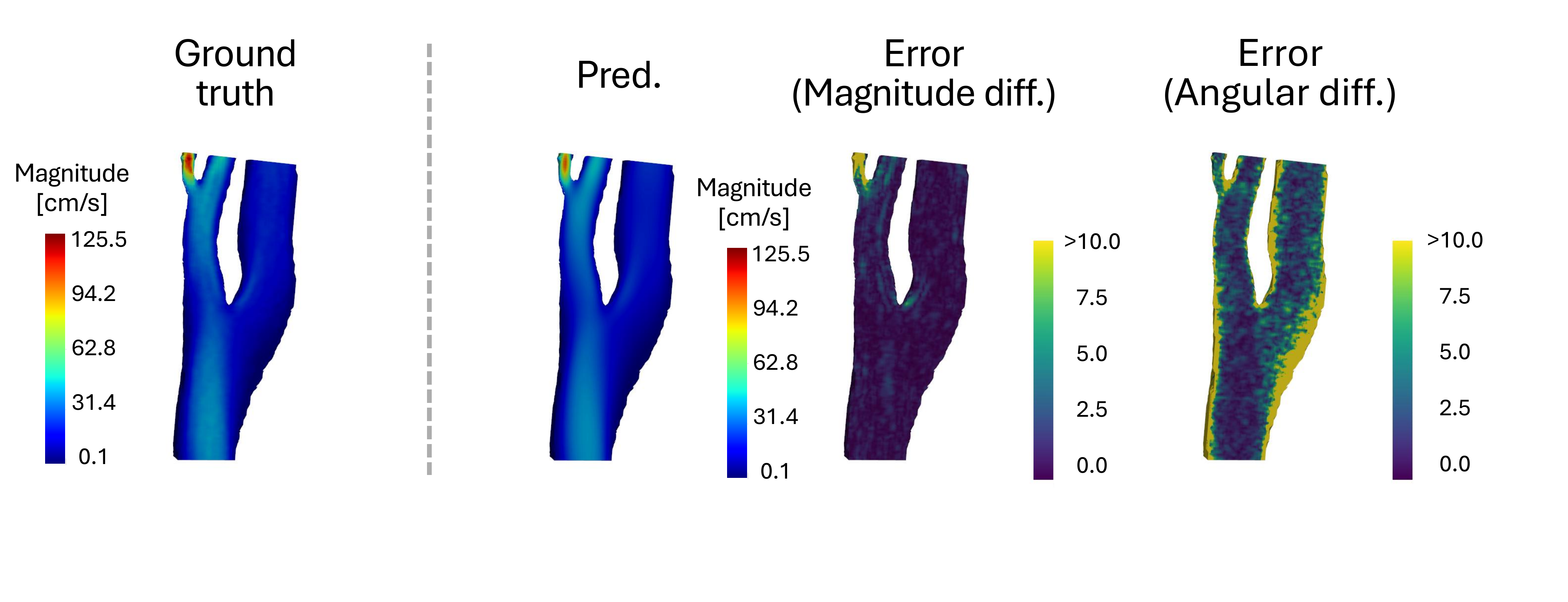}
    \caption{Siren (timeframe=12)}
    \label{fig:sub3_Siren}
  \end{subfigure}
  
  \begin{subfigure}[b]{0.55\textwidth}
    \centering
    \includegraphics[width=0.9\textwidth]{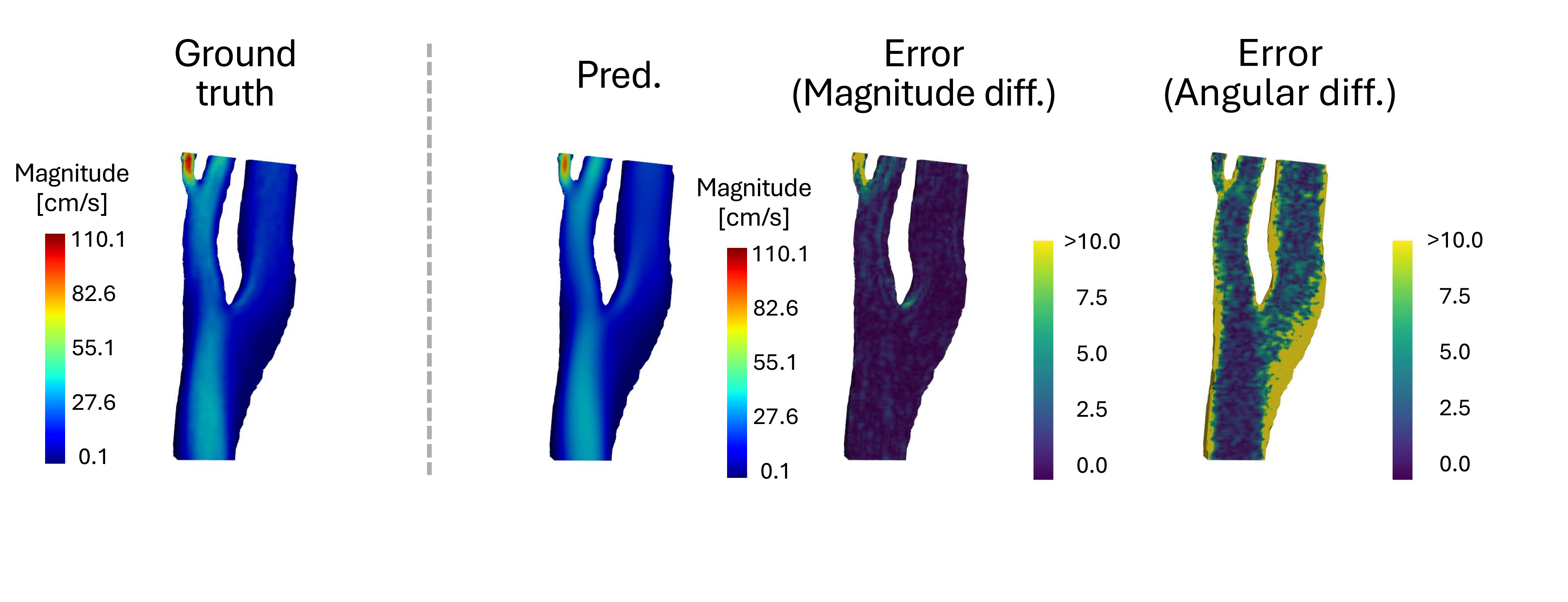}
    \caption{Siren (timeframe=13)}
    \label{fig:sub4_Siren}
  \end{subfigure}
  
  \begin{subfigure}[b]{0.55\textwidth}
    \centering
    \includegraphics[width=0.9\textwidth]{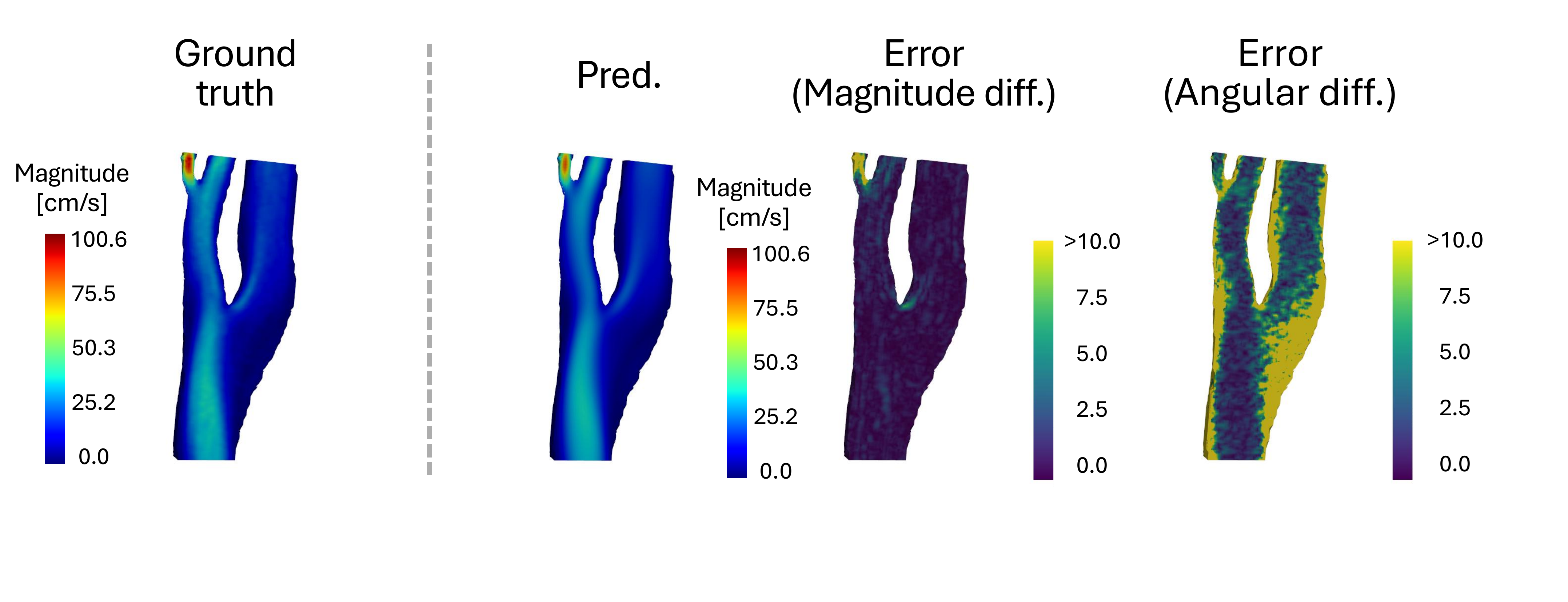}
    \caption{Siren (timeframe=14)}
    \label{fig:sub5_Siren}
  \end{subfigure}
    \caption{Visualizations of the Siren model on $\times 8$ spatially averaged 4D flow MRI dataset over five time frames during the peak systolic phase (10 -- 14). The figure visualizes ground truth velocity field and prediction error maps (magnitude and direction). The color scale for the ground truth and prediction is fixed to the ground truth's minimum and maximum values. For error maps, color scale is fixed at [0,10] for both error maps (yellow = high error).}
     \label{fig:timeframe_X8_Siren}
\end{figure*}

\begin{figure*}[t]
  \centering
  \begin{subfigure}[b]{0.55\textwidth}
  \centering
    \includegraphics[width=0.9\textwidth]{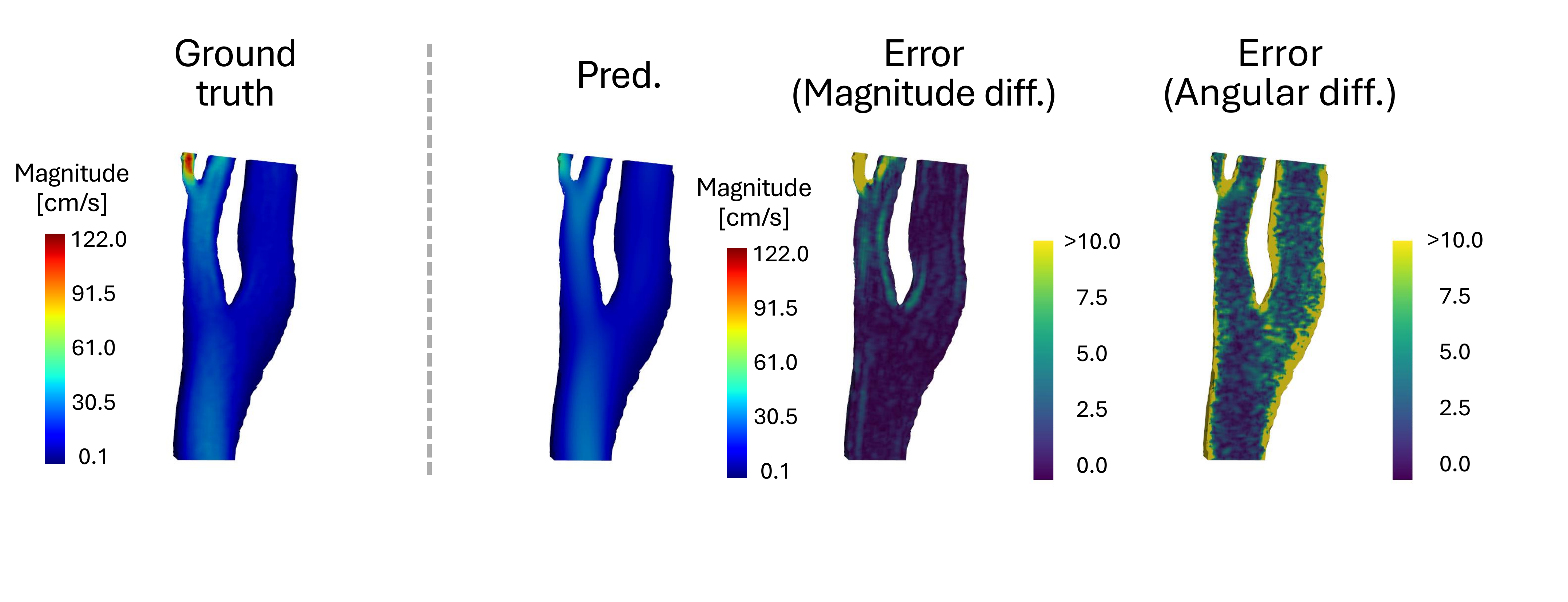}
    \caption{Siren (timeframe=10)}
    \label{fig:sub6_Siren}
  \end{subfigure}
  
  \begin{subfigure}[b]{0.55\textwidth}
    \centering
    \includegraphics[width=0.9\textwidth]{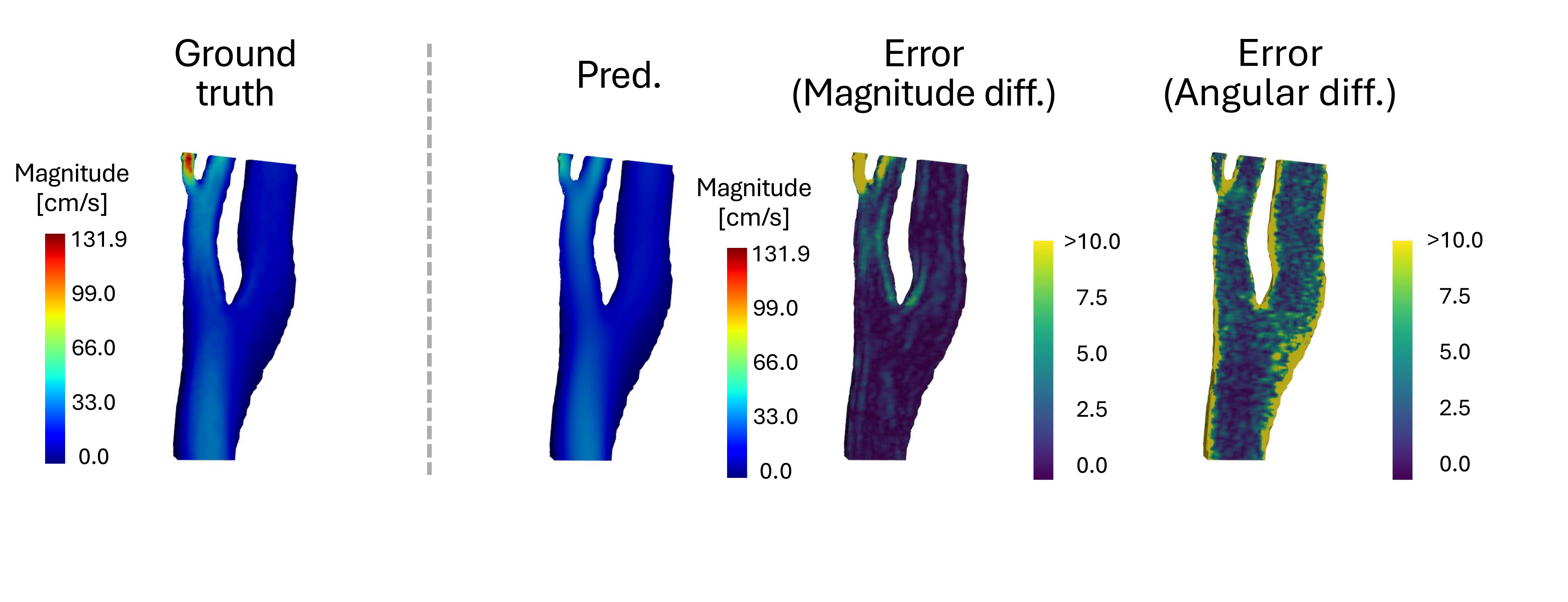}
    \caption{Siren (timeframe=11)}
    \label{fig:sub7_Siren}
  \end{subfigure}
  
  \begin{subfigure}[b]{0.55\textwidth}
    \centering
    \includegraphics[width=0.9\textwidth]{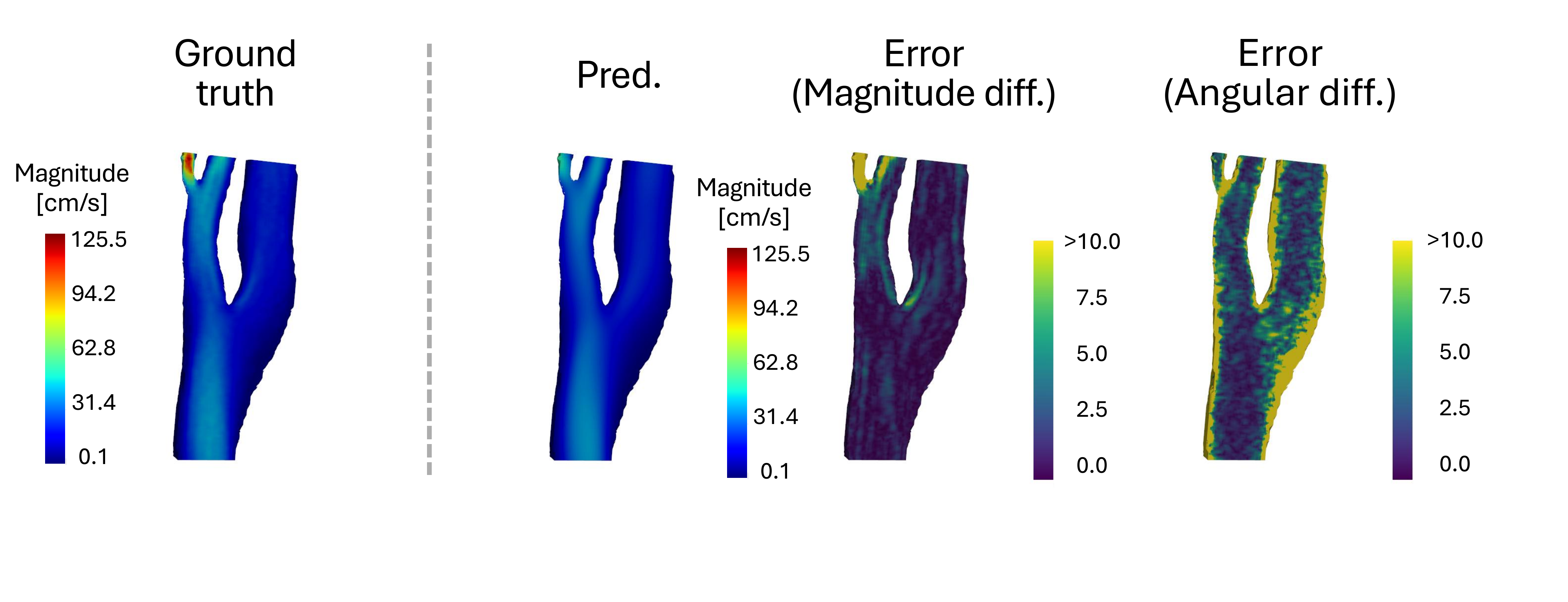}
    \caption{Siren (timeframe=12)}
    \label{fig:sub8_Siren}
  \end{subfigure}
  
  \begin{subfigure}[b]{0.55\textwidth}
    \centering
    \includegraphics[width=0.9\textwidth]{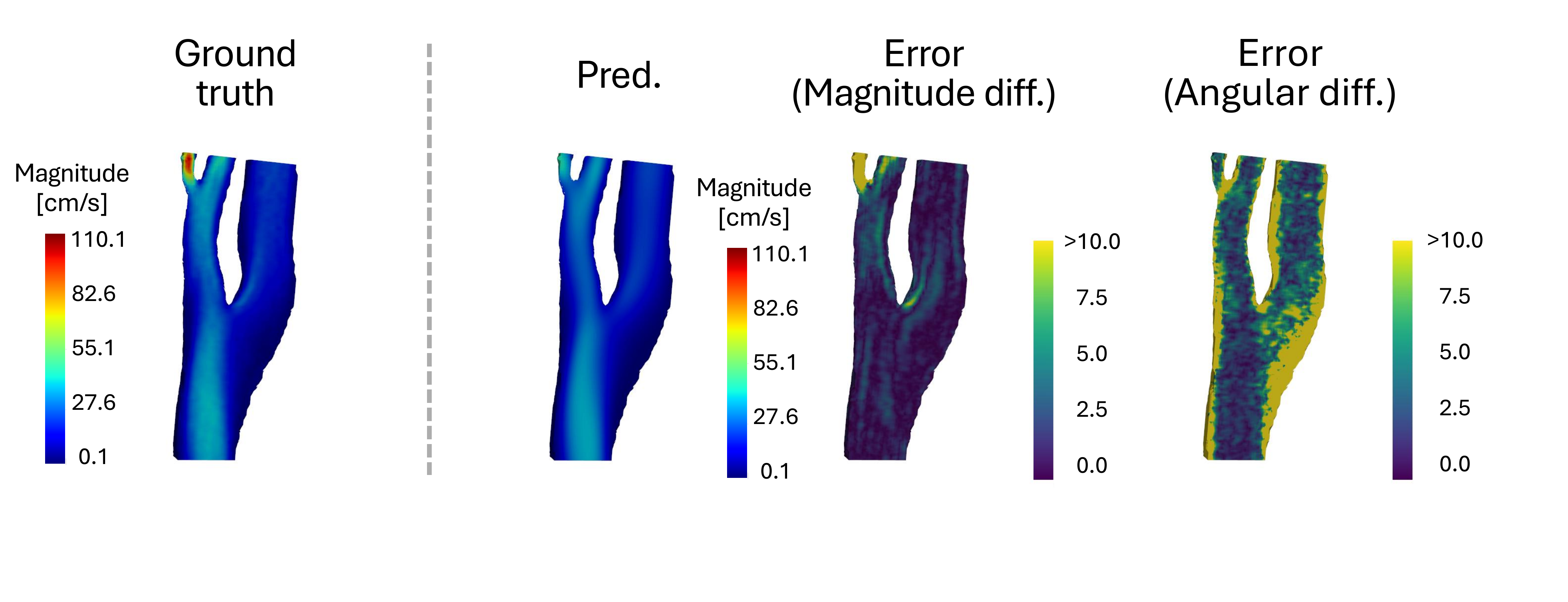}
    \caption{Siren (timeframe=13)}
    \label{fig:sub9_Siren}
  \end{subfigure}
  
  \begin{subfigure}[b]{0.55\textwidth}
    \centering
    \includegraphics[width=0.9\textwidth]{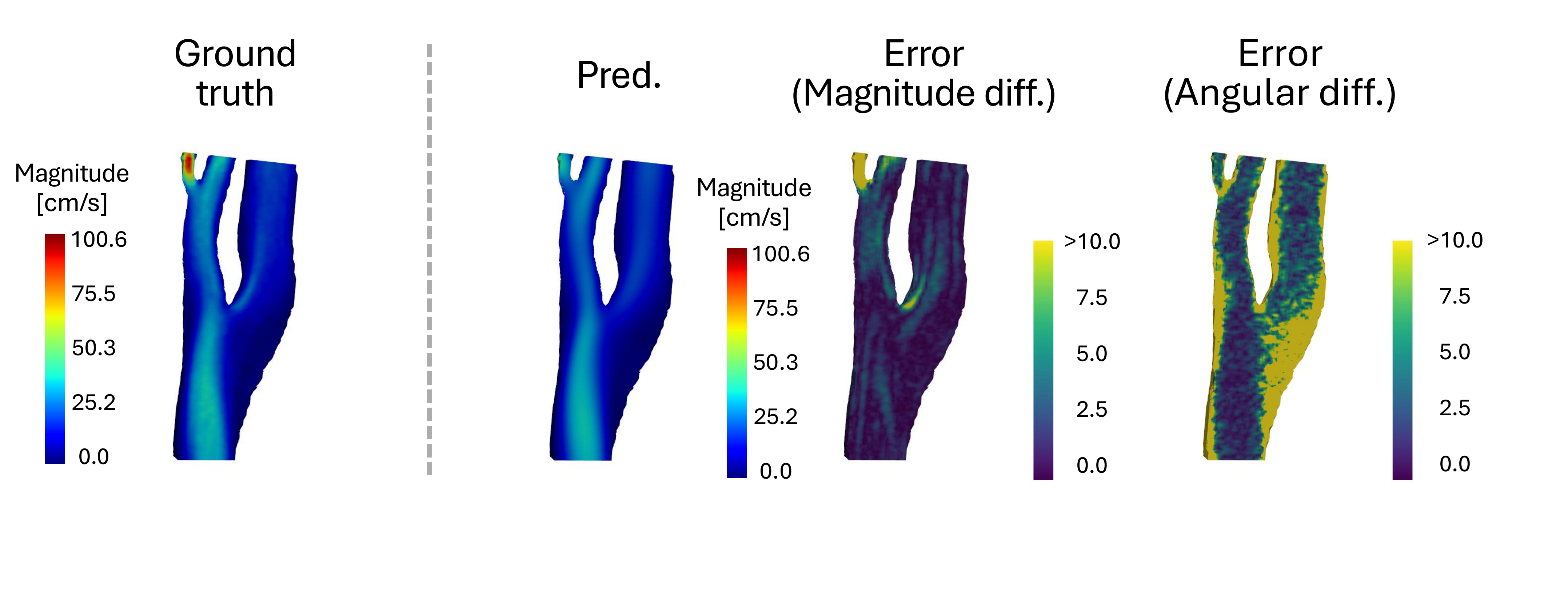}
    \caption{Siren (timeframe=14)}
    \label{fig:sub10_Siren}
  \end{subfigure}
    \caption{Visualizations of the Siren model on $\times 64$ spatially averaged 4D flow MRI dataset over five time frames during the peak systolic phase (10 -- 14). The figure visualizes ground truth velocity field and prediction error maps (magnitude and direction). The color scale for the ground truth and prediction is fixed to the ground truth's minimum and maximum values. For error maps, color scale is fixed at [0,10] for both error maps (yellow = high error).}
     \label{fig:timeframe_X64_Siren}
\end{figure*}

For a qualitative assessment on the 4D flow MRI data, we compare PINGS-X against the PINN and Siren baselines at two resolutions ($\times$8 and $\times$64 spatial averaging) across five peak systolic time frames. The visual results for PINGS-X are presented in Fig.~\ref{fig:timeframe_X8} and Fig.~\ref{fig:timeframe_X64}, with the corresponding baseline results in Fig.~\ref{fig:timeframe_X8_PINN} through Fig.~\ref{fig:timeframe_X64_Siren}.

A visual inspection suggests that while all models perform reasonably at the $\times$8 resolution, the reconstructions from the baseline models degrade more noticeably at the challenging $\times$64 resolution. In contrast, PINGS-X maintains a higher level of detail, producing visibly lower magnitude and angular errors, which indicates a greater robustness to sparse input data.




  
  

\end{document}